\documentclass{article}

\usepackage[margin=0.8in]{geometry}
\usepackage{graphicx}
\usepackage{subfigure}
\usepackage{amsmath,amssymb,color}
\usepackage{cite}

\newcommand{\trace}{ {\rm tr } }
\newcommand{\Pcal}{\mathcal{P}}
\newcommand{\Dcal}{\mathcal{D}}
\newcommand{\skipp}[1]{}

\title{A Path Following Algorithm for the Graph Matching Problem}
\author{Mikhail Zaslavskiy,
        Francis Bach,
        and~Jean-Philippe~Vert
\thanks{Mikhail Zaslavskiy is with the Centre for Computational Biology and the Centre for Mathematical Morphology; Jean-Philippe Vert is with the Centre for Computational Biology, Mines ParisTech, 35 rue Saint-Honor\'e, 77305 Fontainebleau, France. They are also with the Institut Curie, Paris, F-75248 France, and with INSERM U900, Paris, F-75248 France.}
\thanks{E-mail: mikhail.zaslavskiy@ensmp.fr, jean-philippe.vert@ensmp.fr}
\thanks{Francis Bach is with the INRIA-Willow Project Team, Laboratoire d'Informatique de l'Ecole Normale Sup\'erieure 
(CNRS/ENS/INRIA UMR 8548), 45 rue d'Ulm, 75230 Paris, France}
\thanks{E-mail: francis.bach@mines.org}}
\begin{document}
\maketitle
\begin{abstract}
We propose a convex-concave programming approach for the labeled weighted graph matching problem. The convex-concave programming formulation is obtained by rewriting the weighted graph matching problem as a least-square problem on the set of permutation matrices and relaxing it to two different optimization problems: a quadratic convex and a quadratic concave optimization problem on the set of doubly stochastic matrices. The concave relaxation has the same global minimum as the initial graph matching problem, but the search for its global minimum is also a hard combinatorial problem. We therefore construct an approximation of the concave problem solution by following a solution path of a convex-concave problem obtained by linear interpolation of the convex and concave formulations, starting from the convex relaxation. This method allows to easily integrate the information on graph label similarities into the optimization problem, and therefore to perform labeled weighted graph matching. The algorithm is compared with some of the best performing graph matching methods on four datasets: simulated graphs, QAPLib, retina vessel images and handwritten chinese characters. In all cases, the results are competitive with the state-of-the-art. 
\end{abstract}

\begin{center} {\bf Keywords}: Graph algorithms, graph matching, convex programming, gradient methods,  machine learning, classification, image processing \end{center}

\section{Introduction}
The graph matching problem is among the most important challenges of graph processing, and plays a central role in various fields of pattern recognition. Roughly speaking, the problem consists in finding a correspondence between vertices of two given graphs which is optimal in some sense. Usually, the optimality refers to the alignment of graph structures and, when available, of vertices labels, although other criteria are possible as well. A non-exhaustive list of graph matching applications includes document processing tasks like optical character recognition \cite{raymond_graphm_chinese_chars, filatov_graphm_string}, image analysis (2D and 3D) \cite{wang_hancock_matching_kpca, luo_hancock_alignment_svd, carcassoni_hancock_spectral_point_pattern_matching, Schellewald_convex_probabilistic_subgraph_matching}, or bioinformatics \cite{Singh_PPI_matching,wang_graphm_protein_structure, taylor_graphm_protein_structure}. 

During the last decades, many different algorithms for graph matching have been  proposed. Because of the combinatorial nature of this problem, it is very hard to solve it exactly for large graphs, however some methods based on incomplete enumeration may be applied to search for an exact optimal solution in the case of small or sparse graphs. Some examples of such algorithms may be found in \cite{schmidt_backtracking,ullmann_subisomorphism,foggia_graphm_large}.

Another group of methods includes approximate algorithms which are supposed to be more scalable. The price to pay for the scalability is that the solution found is usually only an approximation of the optimal matching. Approximate methods may be divided into two   groups of algorithms. The first group is composed of methods which use spectral representations of adjacency matrices, or equivalently embed the vertices into a Euclidean space where linear or nonlinear matching algorithms can be deployed. This approach was pioneered by Umeyama \cite{umeyama_eigendecomposition_graphm}, while further different methods based on spectral representations were proposed in \cite{shapiro_brady_feature_based_correspondance,carcassoni_hancock_spectral_point_pattern_matching,luo_hancock_alignment_svd,wang_hancock_matching_kpca,caelli_eigenspace_projection_cluster_graphm}. The second   group of approximate algorithms is composed of algorithms which work directly with graph adjacency matrices, and typically involve a relaxation of the discrete optimization problem. The most effective algorithms were proposed in \cite{almohamad_linear_programming_graphm,Gold_graduated_graph_matching,convexgraphmatching,Schellewald_convex_probabilistic_subgraph_matching}.

An interesting instance of the graph matching problem is the matching of labeled graphs. In that case, graph vertices have associated labels, which may be  numbers, categorical variables, etc... The important point is that there is also a similarity measure between these labels. Therefore, when we search for the optimal correspondence between vertices, we search a correspondence which matches not only the structures of the graphs but also vertices with similar labels. Some widely used approaches for this application only use the information about similarities between graph labels.  In vision, one such algorithm is the shape context algorithm proposed in \cite{belongie_shape_matching_shape_context}, which involves an efficient algorithm of node label construction. Another example is the BLAST-based algorithms in bioinformatics such as the Inparanoid algorithm \cite{brein_inparanoid}, where correspondence between different protein networks is established on the basis of BLAST scores between pairs of proteins.  The main advantages of all these  methods are their speed and simplicity. However, the main drawback of these methods is that they do not take into account information about the graph structure. Some graph matching methods try to combine information on graph structures and vertex similarities, examples of such method are presented in \cite{convexgraphmatching,Singh_PPI_matching}.

In this article we propose an approximate method for labeled weighted graph matching, based on a convex-concave programming approach which can be applied for matching of  graphs of large sizes. Our method is based on a formulation of the labeled weighted graph matching problem as a quadratic assignment problem (QAP) over the set of permutation matrices, where the quadratic term encodes the structural compatibility and the linear term encodes local compatibilities. We propose two relaxations of this problem, resulting in one quadratic convex and one quadratic concave minimization problem on the set of doubly stochastic matrices. While the concave relaxation has the same global minimum as the initial QAP, solving it is also a hard combinatorial problem. We find a local minimum of this problem by following a solution path of a family of convex-concave minimization problems, obtained by linearly interpolating between the convex and concave relaxations. Starting from the convex formulation with a unique local (and global) minimum, the solution path leads to a local optimum of the concave relaxation. We refer to this procedure as the PATH algorithm\footnote{The PATH algorithm as well as other referenced approximate methods are implemented in the software GraphM available at {\tt http://cbio.ensmp.fr/graphm}}. We perform an extensive comparison of this PATH algorithm with several state-of-the-art matching methods on small simulated graphs and various QAP benchmarks, and show that it consistently provides state-of-the-art performances while scaling to graphs of up to a few thousands vertices on a modern desktop computer. We further illustrate the use of the algorithm on two applications in image processing, namely the matching of retina images based on vessel organization, and the matching of handwritten chinese characters.

The rest of the paper is organized as follows: Section \ref{sec:pb_description} presents the mathematical formulation of the graph matching problem. In Section \ref{sec:convex-concave_theory}, we present our new approach. Then, in Section \ref{sec:num_exp}, we present the comparison of our method with Umeyama's algorithm \cite{umeyama_eigendecomposition_graphm} and the linear programming approach \cite{almohamad_linear_programming_graphm} on the example of artificially simulated graphs. In Section \ref{sec:qap}, we test our algorithm on the QAP benchmark library, and we compare obtained results with the results published in \cite{convexgraphmatching} for the QBP algorithm and graduated assignment algorithms. Finally, in Section \ref{sec:vision} we present two examples of applications to real-world image processing tasks.

\section{Problem description}
\label{sec:pb_description}
A graph $G=(V,E)$ of size $N$ is defined by a finite set of vertices $V=\{1,\ldots,N\}$ and a set of edges $E \subset V\times V$. We consider only undirected graphs with no self-loop, i.e., such that $(i,j)\in E \implies (j,i)\in E$ and $(i,i)\notin E$ for any vertices $i,j\in V$. Each such graph can be equivalently represented by a symmetric adjacency matrix $A$ of size $|V|\times |V|$, where $A_{ij}$ is equal to one if there is an edge between vertex $i$ and vertex $j$, and zero otherwise. An interesting generalization is a weighted graph which is defined by association of nonnegative real values $w_{ij}$ (weights) to all edges of graph $G$. Such graphs are represented by real valued adjacency matrices $A$ with $A_{ij}=w_{ij}$. This generalization  is   important because in many applications the graphs of interest have associated weights for all their edges, and taking into account these weights may be crucial  in construction of efficient methods. In the following, when we talk about ``adjacency matrix'' we mean a real-valued ``weighted'' adjacency matrix.

Given two graphs $G$ and $H$ with the same number of vertices $N$, the problem of matching $G$ and $H$ consists in finding a correspondence between vertices of $G$ and vertices of $H$ which aligns $G$ and $H$ in some optimal way. We will consider in Section \ref{sec:dummy_nodes} an extension of the problem to graphs of different sizes. For graphs with the same size $N$, the correspondence between vertices is a permutation of $N$ vertices, which can  be defined by a permutation matrix $P$, i.e., a $\{0,1\}$-valued $N\times N$ matrix with exactly one entry $1$ in each column and each row. The matrix $P$ entirely defines the mapping between vertices of $G$ and vertices of $H$, $P_{ij}$ being equal to $1$ if the $i$-th vertex of $G$ is matched to the $j$-th vertex of $H$, and $0$ otherwise. After applying the permutation defined by $P$ to the vertices of $H$ we obtain a new graph isomorphic to $H$ which we denote by $P(H)$. The adjacency matrix of the permuted graph, $A_{P(H)}$, is simply obtained from $A_H$ by the equality $A_{P(H)}=PA_HP^T$. 

In order to assess whether a permutation $P$ defines a good matching between the vertices of $G$ and those of $H$, a quality criterion must be defined. Although other choices are possible, we focus in this paper on measuring the discrepancy between the graphs after matching, by the number of edges (in the case of weighted graphs, it will be the total weight of edges) which are present in one graph and not in the other. In terms of adjacency matrices, this number can be computed as:
\begin{equation}
F_0(P)=||A_G-A_{P(H)}||^2_F=||A_G-P A_H P^T||^2_F\,, \label{eq:adj_distance}
\end{equation}
where $||.||_F$ is the Frobenius matrix norm  defined by
$\|A\|_F^2 = \trace A^T A =(\sum_i\sum_j A^2_{ij})
$.
A popular alternative to the Frobenius norm formulation (\ref{eq:adj_distance}) is the $1$-norm formulation obtained by replacing the Frobenius norm by the $1$-norm $\|A\|_1 =\sum_i\sum_j |A_{ij}|$,  which is equal to the square of the Frobenius norm $\|A\|_F^2$ when comparing $\{0,1\}$-valued matrices, but may differ in the case of general matrices. 

Therefore, the problem of graph matching can be reformulated as the problem of minimizing $F_0(P)$ over the set of permutation matrices. This problem has a combinatorial nature and there is no known polynomial algorithm to solve it \cite{garey_NP}. It is therefore very hard to solve it in the case of large graphs, and numerous approximate methods have been developed.

An interesting generalization of the graph matching problem is the problem of \emph{labeled} graph matching. Here, each graph has associated labels to all its vertices and the objective is to find an alignment that fits well graph labels and graph structures at the same time. If we let $C_{ij}$ denote the cost of fitness between the $i$-th vertex of $G$ and the $j$-th vertex of $H$, then the matching problem based only on label comparison can be formulated as follows:
\begin{equation}
\min_{P \in \mathcal P} \ \mbox{tr}(C^TP) = \sum_{i=1}^{N}
\sum_{j=1}^{N}C_{ij} P_{ij}\,,
\label{eq:context_matching}
\end{equation}
where $\mathcal P$ denotes the set of permutation matrices.
A natural way of unifying (\ref{eq:context_matching}) and (\ref{eq:adj_distance}) to match both the graph structure and the vertices' labels is then to minimize a convex combination \cite{convexgraphmatching}:
\begin{equation}
\min_{P \in \mathcal P }\ (1-\alpha)F_0(P)+\alpha\mbox{tr}(C^TP),
\label{eq:F_0_alpha}
\end{equation}
that makes explicit, through the parameter $\alpha\in[0,1]$, the trade-off between cost of individual matchings and faithfulness to the graph structure. A small $\alpha$ value emphasizes the matching of structures, while a large $\alpha$ value gives more importance to the matching of labels.

\subsection{Permutation matrices}

Before describing how we propose to solve (\ref{eq:adj_distance}) and (\ref{eq:F_0_alpha}), we first introduce some notations and bring to notice some important characteristics of these optimization problems. They are defined on the set of permutation matrices, which we   denoted by $\mathcal P$. The set $\mathcal P$ is a set of extreme points of the set of doubly stochastic matrices, that is, matrices with nonnegative entries and  with row sums and column sums equal to one: ${\mathcal D}=\{A: A1_N=1_N,~ A^T1_N=1_N,~ A\geq 0\}$, where $1_N$ denotes the $N$-dimensional vector of all ones \cite{lewis}. This shows that when a linear function is minimized over the set of doubly stochastic matrices $\mathcal D$, a solution can always be found in the set of permutation matrices. Consequently, minimizing a linear function over $\mathcal P$ is in fact equivalent to a linear program and can thus be solved in polynomial time by, e.g., interior point methods \cite{Boyd2003Convex}. In fact, one of the most efficient methods to solve this problem is the Hungarian algorithm, which uses a specific primal-dual strategy to solve this problem in $O(N^3)$ \cite{hungarian-2}. Note that the Hungarian algorithm allows to avoid the generic $O(N^7)$ complexity associated with a linear program with $N^2$ variables.

\begin{figure}[htb]
\centering
\includegraphics[width=6cm]{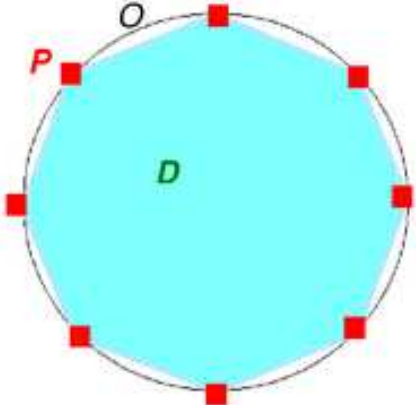}

\vspace*{-.5cm}

\caption{Relation between three matrix sets. $\mathcal O$---set of
orthogonal matrices, $\mathcal D$ --- set of doubly stochastic
matrices, $\mathcal P=D \cap O $---set of permutation matrices.}
\label{fig:qcvqcc_relaxation}
\end{figure}

At the same time $\mathcal P$ may be considered as a subset of orthonormal matrices ${\mathcal O}=\{A: A^TA=I\}$ of $\mathcal D$ and in fact $\mathcal P=\mathcal D\cap \mathcal O$. An (idealized) illustration of these sets is presented in Figure~\ref{fig:qcvqcc_relaxation}: the discrete set $\mathcal P$ of permutation matrices is the intersection of the convex set $\mathcal D$ of doubly stochastic matrices and the manifold $\mathcal O$ of orthogonal matrices.

\subsection{Approximate methods: existing works}
\label{sec:related_methods}

A good review of graph matching algorithms may be found in \cite{conte_foggia_30_years_of_graph_matching}. Here, we only present a brief description of some approximate methods which illustrate well ideas behind two subgroups of these algorithms. 
As mentioned in the introduction, one popular approach to find approximate solutions to the graph matching problem is based on the spectral decomposition of the adjacency matrices of the graphs to be matched. In this approach, the singular value decompositions of the graph adjacency matrices are used:
\begin{equation*}
A_G=U_G\Lambda_GU_G^T,~A_H=U_H\Lambda_HU_H^T,
\end{equation*}
where the columns of the orthogonal matrices $U_G$ and $U_H$ consist of eigenvectors of $A_G$ and $A_H$ respectively, and $\Lambda_G$ and $\Lambda_H$ are diagonal matrices of eigenvalues.

If we consider the rows of eigenvector matrices $U_G$ and $U_H$ as graph node coordinates in eigenspaces, then we can match the vertices with similar coordinates through a variety of methods \cite{umeyama_eigendecomposition_graphm,carcassoni_hancock_spectral_point_pattern_matching,caelli_eigenspace_projection_cluster_graphm}. However, these methods suffer from the fact that the spectral embedding of graph vertices is not uniquely defined. First, the unit norm eigenvectors are at most defined up to a sign flip and we have to choose their signs synchronously. Although it is possible to use some normalization convention, such as choosing the sign of each eigenvector in such a way that the biggest component is always positive, this usually does not guarantee a perfect sign synchronization, in particular in the presence of noise. Second, if the adjacency matrix has multiple eigenvalues, then the choice of eigenvectors becomes arbitrary within the corresponding eigen-subspace, as they are defined only up to rotations \cite{golub}. 

One of the first spectral approximate algorithms was presented by Umeyama \cite{umeyama_eigendecomposition_graphm}. To avoid the ambiguity of eigenvector selection, Umeyama proposed to consider the absolute values of eigenvectors. According to this approach, the correspondence between graph nodes is established by matching the rows of $|U_G|$ and $|U_H|$ (which are defined as matrices of absolute values). The criterion of optimal matching is the total distance between matched rows, leading to the optimization problem:
$$\min_{P\in \mathcal{P}} \ \| \ \vert U_G\vert-P \vert U_H\vert\ \|_F \,,
$$
or equivalently:
\begin{equation}\label{eq:U_optimization}
\max_{P\in \mathcal{P}} \ \trace (|U_H||U_G|^TP)\,.
\end{equation}
The optimization problem (\ref{eq:U_optimization}) is a linear program on the set of permutation matrices which can be solved by the Hungarian algorithm in $O(N^3)$ \cite{mcginnis_hungarian,kuhn_hungarian}. 

The second group of approximate methods consists of algorithms which work directly with the objective function in (\ref{eq:adj_distance}), and typically involve various relaxations to optimizations problems that can be efficiently solved. An example of such an approach is the linear programming method proposed by Almohamad and Duffuaa in \cite{almohamad_linear_programming_graphm}. They considered the $1$-norm as the matching criterion for a permutation matrix $P\in\mathcal{P}$:
\begin{equation*}
F_0'(P)=||A_G-P A_H P^T||_{1}=||A_GP-P A_H||_{1}.
\end{equation*}
The linear program relaxation is obtained by optimizing $F_0'(P)$ on the set of doubly stochastic matrices $\mathcal D$ instead of $\Pcal$:
\begin{equation}\label{eq:adj_distance_l1}
\min_{P \in \mathcal D} \ F_0'(P)\, ,
\end{equation}
where the 1-norm of a matrix is defined as the sum of the absolute values of all the elements of a matrix.
A priori the solution of (\ref{eq:adj_distance_l1}) is an arbitrary doubly stochastic matrix $X\in\Dcal$, so the final step is a projection of $X$ on the set of permutation matrices
(we let denote $\Pi_{\mathcal P} X$ the projection of $X$ onto $\mathcal{P}$) :
$$
P^*=\Pi_{\mathcal P} X = \arg\min_{P\in \mathcal P} \ ||P-X||^2_F\,,
$$
or equivalently:
\begin{equation}\label{eq:projsurP}
P^*=\arg\max_{P\in \mathcal P}{X^T P}\,.
\end{equation}
The projection (\ref{eq:projsurP}) can be performed with the Hungarian algorithm, with a complexity cubic in the dimension of the problem. The main disadvantage of this method is that the dimensionality (i.e., number of variables and number of constraints) of the linear program (\ref{eq:projsurP}) is $O(N^2)$, and therefore it is quite hard to process graphs of size more than one hundred nodes. 

Other convex relaxations of (\ref{eq:adj_distance}) can be found in \cite{convexgraphmatching} and \cite{Gold_graduated_graph_matching}. In the next section we describe our new algorithm which is based on the technique of convex-concave relaxations of the initial problems (\ref{eq:adj_distance}) and (\ref{eq:F_0_alpha}).
 
\section{Convex-concave relaxation}
\label{sec:convex-concave_theory}
Let us start the description of our algorithm for unlabeled weighted graphs. The generalization to labeled weighted graphs is presented in Section \ref{sec:addlabels}. The graph matching criterion we consider for unlabeled graphs is the square of the Frobenius norm of the difference between adjacency matrices (\ref{eq:adj_distance}). Since permutation matrices are also orthogonal matrices (i.e., $PP^T=I$ and $P^TP=I$), we can rewrite $F_0(P)$ on $\mathcal P$ as follows:
\begin{equation*}
\begin{split}
F_0(P)&=\|A_G-PA_HP^T\|_F^2=\|(A_G-PA_HP^T)P\|_F^2\\
&=\|A_G P-P A_H\|_F^2\,.
\end{split}
\end{equation*}
The graph matching problem is then the problem of minimizing $F_0(P)$ over $\Pcal$, which we call \textbf{GM}:
\begin{equation}
\mbox{\textbf{GM}: } \min_{P \in \mathcal P} \  F_0(P)\,.
\label{eq:gm}
\end{equation} 

\subsection{Convex relaxation}
A first relaxation of \textbf{GM} is obtained by expanding the convex quadratic function $F_0(P)$ on the set of doubly stochastic matrices $\mathcal D$: 
\begin{equation}
\mbox{\textbf{QCV}: } \min_{P \in \mathcal D} \  F_0(P)\,.
\label{eq:qcv}
\end{equation} 
The \textbf{QCV} problem is a convex quadratic program that can be solved in polynomial time, e.g., by the Frank-Wolfe algorithm \cite{frank_wolfe} (see Section \ref{sec:FW} for more details). However, the optimal value is usually not an extreme points of $\mathcal D$, and therefore not a permutation matrix. If we want to use only \textbf{QCV} for the graph matching problem, we therefore have to project its solution on the set of permutation matrices, and to make, e.g., the following approximation:
\begin{equation}
\arg \min_{\mathcal P} \ F_0(P) \thickapprox \Pi_{\mathcal P}\arg\min_{\mathcal D} \ F_0(P)\,.
\label{eq:qcv_proj}
\end{equation}
Although the projection $\Pi_{\mathcal P}$ can be made efficiently in $O(N^3)$ by the Hungarian algorithm, the difficulty with this approach is that if $\arg\min_{\mathcal D} \ F_0(P)$ is far from $\mathcal P$ then the quality of the approximation (\ref{eq:qcv_proj}) may be poor: in other words, the work performed to optimize $F_0(P)$ is partly lost by the projection step which is independent of the cost function. The PATH algorithm that we present later can be thought of as a improved projection step that takes into account the cost function in the projection.

\subsection{Concave relaxation}

We now present a second relaxation of \textbf{GM}, which results in a concave minimization problem. For that purpose, let us introduce the diagonal degree matrix $D$ of an adjacency matrix $A$, which is the diagonal matrix with entries $D_{ii} = d(i) = \sum_{i=1}^N A_{ij}$ for $i=1,\ldots,N$, as well as the Laplacian matrix $L=D-A$. $A$ having only nonnegative entries, it is well-known that the Laplacian matrix is positive semidefinite~\cite{chung}. We can now rewrite $F_0(P)$ as follows:
\begin{equation}\label{eq:toto1}
\begin{split}
F_0(P)=&||A_G P-P A_H||_F^2 \\
=&||(D_G P-P D_H)-(L_GP-PL_H)||^2_F\\
=&||D_G P -P D_H||_F^2\\
&-2\trace[(D_G P-P D_H)^T(L_GP-PL_H)]\\
&+||L_GP-PL_H||^2_F\,.
 \end{split}
\end{equation}
Let us now consider more precisely the second term in this last expression:
\begin{equation}\label{eq:toto2}
\begin{split}
&\trace[(D_G P-P D_H)^T(L_GP-PL_H)]\\
&= \underbrace{\trace PP^TD_G^TL_G}_{\sum d_G^2(i)}+\underbrace{\trace L_HD_H^TP^TP}_{\sum d_H^2(i)}-\underbrace{\trace P^TD_G^TPL_H}_{\sum d_G(i)d_{P(H)}(i)}\\
&\qquad \qquad \qquad \qquad \qquad \qquad \qquad   -\underbrace{\trace D_H^TP^TL_GP}_{\sum d_{P(H)}(i)d_G(i)}\\
&=\sum (d_G(i)-d_{P(H)}(i))^2 =\|D_G-D_{P(H)}\|^2_F\\
&=\|D_GP-PD_H\|^2_F\,.
\end{split}
\end{equation}
Plugging (\ref{eq:toto2}) into (\ref{eq:toto1}) we obtain
\begin{equation}\label{eq:last_expr_F1}
\begin{split}
&F_0(P)= \|D_G P -P D_H\|_F^2-2\|D_G P -P D_H\|_F^2\\
&~+\|L_GP-PL_H\|^2_F\\
&= -\|D_G P -P D_H\|_F^2+\|L_GP-PL_H\|^2_F\\
&= -\sum_{i,j}P_{ij}(D_G(j)-D_H(i))^2 + \trace(\underbrace{PP^T}_{I}L_G^TL_G)\\
&\qquad +\trace(L_H^T\underbrace{P^TP}_{I}L_H)-2\underbrace{\trace(P^TL_G^TPL_H)}_{\mbox{vec}(P)^T(L^T_H\otimes L^T_G)\mbox{vec}(P)}\\
&= -\trace(\Delta P) + \trace(L_G^2)+\trace(L_H^2)\\
&\quad \qquad \qquad -2\mbox{vec}(P)^T(L^T_H\otimes L^T_G)\mbox{vec}(P)\,,
\end{split}
\end{equation}
where we introduced the matrix $\Delta_{i,j}=(D_H(j,j)-D_G(i,i))^2$ and we used $\otimes$ to denote the Kronecker product of two matrices (definition of the Kronecker product and some important properties may be found in the appendix \ref{sec:kron}).

Let us denote $F_1(P)$ the part of  (\ref{eq:last_expr_F1}) which depends on $P$:
\begin{equation*}
F_1(P)=-\trace(\Delta P)-2\mbox{vec}(P)^T(L^T_H\otimes L^T_G)\mbox{vec}(P).
\end{equation*}
From (\ref{eq:last_expr_F1}) we see that the permutation matrix which minimizes $F_{1}$ over $\Pcal$ is the solution of the graph matching problem. Now, minimizing $F_{1}(P)$ over $\Dcal$ gives us a relaxation of (\ref{eq:gm}) on the set of doubly stochastic matrices. Since graph Laplacian matrices are positive semi-definite, the matrix $L_H\otimes L_G$ is also positive semidefinite as a Kronecker product of two symmetric positive semi-definite matrices~\cite{golub}.  Therefore the function $F_1(P)$ is concave on $\Dcal$, and we obtain a concave relaxation of the graph matching problem:
\begin{equation}
\mbox{\textbf{QCC}: } \min_{P \in \mathcal D} \ F_1(P).
\label{eq:qcc}
\end{equation} 
Interestingly, the global minimum of a concave function is necessarily located at a boundary of the convex set where it is minimized\cite{rockafeller}, so the minimium of $F_{1}(P)$ on $\Dcal$ is in fact in $\Pcal$.

At this point, we have obtained two relaxations of \textbf{GM} as nonlinear optimization problems on~$\Dcal$: the first one is the convex minimization problem \textbf{QCV} (\ref{eq:qcv}), which can be solved efficiently but leads to a solution in $\Dcal$ that must then be projected onto $\Pcal$, and the other is the concave minimization problem \textbf{QCC} (\ref{eq:qcc}) which does not have an efficient (polynomial) optimization algorithm but has the same solution as the initial problem \textbf{GM}. We note that these convex and concave relaxation of the graph matching problem can be more generally derived for any quadratic assignment problems \cite{anstreicher01new}.
\subsection{PATH algorithm}
We propose to   approximately solve \textbf{QCC} by tracking a path of local minima over $\Dcal$ of a series of functions that linearly interpolate between $F_{0}(P)$ and $F_{1}(P)$, namely:
$$
F_\lambda(P)=(1-\lambda)F_0(P)+\lambda F_1(P)\,,
$$
for $0\leq\lambda\leq 1$. For all $\lambda\in[0,1]$, $F_{\lambda}$ is a quadratic function (which is in general neither convex nor concave for $\lambda$ away from zero or one). We recover the convex function $F_{0}$ for $\lambda=0$, and the concave function $F_{1}$ for $\lambda=1$. Our method searches sequentially local minima of $F_{\lambda}$, where $\lambda$ moves from $0$ to $1$. More precisely, we start at $\lambda=0$, and find the unique local minimum of $F_{0}$ (which is in this case its unique global minimum) by any classical QP solver. Then, iteratively, we find a local minimum of $F_{\lambda+d\lambda}$ given a local minimum of $F_{\lambda}$ by performing a local optimization of $F_{\lambda+d\lambda}$ starting from the local minimum of $F_{\lambda}$, using for example the Frank-Wolfe algorithm. Repeating this iterative process for $d\lambda$ small enough we obtain a path of solutions $P^*(\lambda)$, where $P^*(0) = \arg\min_{P\in\Dcal} \ F_{0}(P)$ and $P^*(\lambda)$ is a local minimum of $F_{\lambda}$ for all $\lambda\in[0,1]$. Noting that any local minimum of the concave function $F_{1}$ on $\Dcal$ is in $\Pcal$, we finally output $P^*(1)\in\Pcal$ as an approximate solution of \textbf{GM}.

Our approach is similar to graduated non-convexity \cite{blake_vr}: this approach  is often used to approximate the global minimum of a non-convex objective function. This function consists of two part, the convex component, and non-convex component, and the graduated non-convexity framework proposes to track the linear combination of the convex and non-convex parts (from the convex relaxation to the true objective function) to approximate the minimum of the non-convex function. The PATH algorithm may indeed be considered as an example of such an approach. However, the main difference is the construction of the objective function. Unlike \cite{blake_vr}, we construct two relaxations of the initial optimization problem, which lead to the same value  on the set of interest ($\mathcal P$), the goal being to  choose convex/concave relaxations which approximate in the best way the objective function on the set of permutation matrices.

\begin{figure}[htbp]
\centering
\includegraphics[width=8cm]{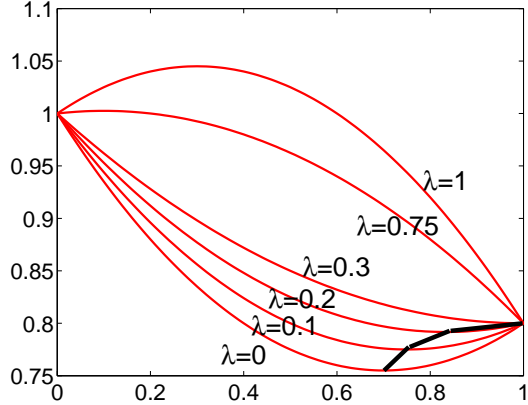}
\caption{Illustration for path optimization approach.
$F_0$ ($\lambda=0$) --- initial convex function, $F_1$ ($\lambda=1$) ---
initial concave function, bold black line --- path of function
minima $P^*(\lambda)$ ($\lambda=0\dots 0.1\dots 0.2\dots 0.3\dots 0.75\dots 1$)}
\label{fig:qcvqcc_path}
\end{figure}

The pseudo-code for the PATH algorithm is presented in Figure \ref{fig:algo_schema}. The rationale behind it is that among the local minima of $F_{1}(P)$ on $\Dcal$, we expect the one connected to the global minimum of $F_{0}$ through a path of local minima to be a good approximation of the global minima. Such a situation is for example shown in Figure \ref{fig:qcvqcc_path}, where in $1$ dimension the global minimum of a concave quadratic function on an interval (among two candidate points) can be found by following the path of local minima connected to the unique global minimum of a convex function. 

\begin{figure}[htb]
\centering
\begin{enumerate}
\item {\bf Initialization}:
\begin{enumerate} 
\item $\lambda:=0$
\item $P^*(0)=\arg\min F_{0}$ --- convex optimization problem, global minimum is found by Frank-Wolfe algorithm.
\end{enumerate}
\item {\bf Cycle over $\lambda$}: \\
while $\lambda<1$ \\
\begin{enumerate}
\item  $\lambda_{new}:=\lambda+d\lambda$
\item if $\vert F_{\lambda_{new}}(P^*(\lambda))-F_{\lambda}(P^*(\lambda))\vert <\epsilon_{\lambda}$ then\\
	\qquad $\lambda = \lambda_{new}$\\
	else $P^*(\lambda_{new})=\arg\min F_{\lambda_{new}}$ is found\\
	\qquad \quad by Frank-Wolfe starting from $P^*(\lambda)$\\
	\qquad $\lambda=\lambda_{new}$\\
\end{enumerate}
\item {\bf Output}: $P^{out}:= P^*(1)$
\end{enumerate}
\caption{Schema of the  PATH algorithm}
\label{fig:algo_schema}
\end{figure}

More precisely, and although we do not have any formal result about the optimality of the PATH optimization method (beyond the lack of global optimality, see Appendix \ref{app:toy}), we can mention a few interesting properties of this method:
\begin{itemize}
\item We know from (\ref{eq:last_expr_F1}) that for $P\in\Pcal$, $F_1(P) = F_0(P) - \kappa$, where $\kappa = \trace(L_G^2)+\trace(L_H^2)$ is a constant independent of $P$. As a result, it holds for all $\lambda\in[0,1]$ that, for $P\in\Pcal$: 
$$
F_{\lambda}(P) =  F_0(P) -\lambda\kappa\,.
$$
This shows that if for some $\lambda$ the global minimum of $F_\lambda(P)$ over $\Dcal$ lies in ${\mathcal P}$, then this minimum is also the global minimum of $F_0(P)$ over $\Pcal$ and therefore the optimal solution of the initial problem.  Hence, if for example the global minimum of $F_{\lambda}$ is found on $\Pcal$ by the PATH algorithm (for instance, if $F_\lambda$ is still convex), then the PATH algorithm leads to the global optimum of $F_1$. This situation can be seen in the toy example in Figure \ref{fig:qcvqcc_path} where, for $\lambda=0.3$, $F_{\lambda}$  has its unique minimum at the boundary of the domain.

\item The sub-optimality of the PATH algorithm comes from the fact that, when $\lambda$ increases, the number of local minima of $F_{\lambda}$ may increase and the sequence of local minima tracked by PATH may not be global minima. However we can expect the local minima followed by the PATH algorithm to be interesting approximations for the following reason. First observe that if $P_{1}$ and $P_{2}$ are two local minima of $F_{\lambda}$ for some $\lambda\in[0,1]$, then the restriction of $F_{\lambda}$ to $(P_{1},P_{2})$ being a quadratic function it has to be concave and $P_{1}$ and $P_{2}$ must be on the boundary of $\Dcal$. Now, let $\lambda_{1}$ be the smallest $\lambda$ such that $F_{\lambda}$ has several local minima on $\Dcal$. If $P_{1}$ denotes the local minima followed by the PATH algorithm, and $P_{2}$ denotes the ``new'' local minimum of $F_{\lambda_{1}}$, then necessarily the restriction of $F_{\lambda_{1}}$ to $(P_{1},P_{2})$ must be concave and have a vanishing derivative in $P_{2}$ (otherwise, by continuity of $F_{\lambda}$ in $\lambda$, there would be a local minimum of $F_{\lambda}$ near $P_{2}$ for $\lambda$ slightly smaller than $\lambda_{1}$). Consequently we necessarily have $F_{\lambda_{1}}(P_{1}) < F_{\lambda_{1}}(P_{2})$. This situation is illustrated in Figure \ref{fig:qcvqcc_path} where, when the second local minimum appears for $\lambda=0.75$, it is worse than the one tracked by the PATH algorithm. More generally, when ``new'' local minima appear, they are strictly worse than the one tracked by the PATH algorithm. Of course, they may become better than the PATH solution when $\lambda$ continues to increase.
\end{itemize}

Of course, in spite of these justifications, the PATH algorithm only gives an approximation of the global minimum in the general case. In  Appendix \ref{app:toy}, we provide two simple examples when the PATH algorithm respectively succeeds and fails to find the global minimum of the graph matching problem.

\subsection{Numerical continuation method interpretation}
Our path following algorithm may be considered as a particular case of numerical continuation methods (sometimes called path following methods) \cite{Allgower_Georg_numerical_continuation}. These allow to estimate curves given in the following implicit form:
\begin{equation}
T(u)=0\mbox{ where $T$ is a mapping } T:R^{K+1}\rightarrow R^K\,.
\label{eq:analysis_T}
\end{equation} 
In fact, our PATH algorithm corresponds to a particular implementation of the so-called Generic Predictor Corrector Approach \cite{Allgower_Georg_numerical_continuation} widely used in numerical continuation methods.

In our case, we have a set of problems $\min_{P\in \Dcal} \ {(1-\lambda)F_0(P)+\lambda F_1(P)}$ parametrized by $\lambda \in [0,1]$. In other words for each $\lambda$ we have to solve the following system of Karush-Kuhn-Tucker (KKT) equations:
\begin{eqnarray*}
(1-\lambda)\nabla_P F_0(P)+\lambda \nabla_P F_1(P)+B^T\nu+\mu_S&=&0 \,, \\
B\mbox{P}-1_{2N}&=&0 \,, \\
P_S&=&0 \,,
\end{eqnarray*}
where $S$ is a set of active constraints, i.e., of pairs of indices $(i,j)$ that satisfy $P_{ij}=0$, $B\mbox{P}-1_{2N}=0$ codes the conditions $\sum_j P_{ij}=1~\forall i$ and $\sum_i P_{ij}=1~\forall j$, $\nu$ and $\mu_S$ are dual variables. We have to solve this system for all possible sets of active constraints $S$ on the open set of matrices $P$ that satisfy $P_{i,j}>0$ for $(i,j)\notin S$, in order to define the set of stationary points of the functions $F_{\lambda}$. Now if we let $T(P,\nu,\mu,\lambda)$ denote the left-hand part of the KKT equation system then we have exactly (\ref{eq:analysis_T}) with $K=N^2+2N+\#S$. 
From the implicit function theorem \cite{milnor}, we know that for each set of constraints $S$,
\begin{equation*}
\begin{split}
W_S=&\{(P,\nu,\mu_S,\lambda):T(P,\nu,\mu_S,\lambda)=0\mbox{ and } \\
&T'(P,\nu,\mu_S,\lambda) \mbox{ has the maximal possible rank}\}
\end{split}
\end{equation*}
is a smooth 1-dimensional curve or the empty set and can be parametrized by $\lambda$. In term of the objective function $F_\lambda(P)$, the condition on $T'(P,\nu,\mu_S,\lambda)$ may be interpreted as a prohibition for the projection of $F_\lambda(P)$ on any feasible direction to be a constant. Therefore the whole set of stationary points of $F_\lambda(P)$ when $\lambda$ is varying from 0 to 1 may be represented as a union $W(\lambda)=\cup_{S}W_S(\lambda)$ where each $W_S(\lambda)$ is homotopic to a 1-dimensional segment. The set $W(\lambda)$ may have quite complicated form. Some of $W_S(\lambda)$ may intersect each other, in this case we observe a bifurcation point, some of $W_S(\lambda)$ may connect each other, in this case we have  a transformation point of one path into another, some of $W_S(\lambda)$ may appear only for $\lambda>0$ and/or disappear before $\lambda$ reaches 1. At the beginning the PATH algorithm starts from $W_{\emptyset}(0)$, then it follows $W_{\emptyset}(\lambda)$ until  the border of $\Dcal$ (or a bifurcation point). If such an event occurs before $\lambda=1$ then PATH moves to another segment of solutions corresponding to different constraints $S$, and keeps moving along segments and sometimes jumping between segments until $\lambda=1$. As we said in the previous section one of the interesting properties of PATH algorithm is the fact that if $W_S^*(\lambda)$ appears only when $\lambda=\lambda_1$ and $W_S^*(\lambda_1)$ is a local minimum then the value of the objective function $F_{\lambda_{1}}$ in $W_S^*(\lambda_1)$ is greater than in the point traced by the PATH algorithm.

\subsection{Some implementation details}
\label{sec:FW}

In this section we provide a few details relevant for the efficient implementation of the PATH algorithms.

\paragraph{Frank-Wolfe}

Among the different optimization techniques for the optimization of $F_{\lambda}(P)$ starting from the current local minimum tracked by the PATH algorithm, we use in our experiments the Frank-Wolfe algorithm which is particularly suited to optimization over doubly stochastic matrices \cite{bertsekas_nonlinear_optimization}. The idea of the this algorithm is to sequentially minimize linear approximations of $F_0(P)$. Each step includes three operations:
\begin{enumerate}
\item estimation of the gradient $\nabla F_{\lambda}(P_n)$, 
\item resolution of the linear program $P_n^*=\arg\min_{P\in D}\langle\nabla F_{\lambda}(P_n),P\rangle$,  
\item line search: finding the minimum of $F_{\lambda}(P)$ on the segment $[P_n~P_n^*]$.
\end{enumerate}
An important property of this method is that the second operation can be done efficiently by the Hungarian algorithm, in $O(N^3)$. 

\paragraph{Efficient gradient computations}
Another essential point is that we do not need to store matrices of size $N^2\times N^2$ for the computation of $\nabla F_{1}(P)$, because the tensor product in $\nabla F_1(P)=-\mbox{vec}(\Delta^T) -2(L_H^T\otimes L^T_G)\mbox{vec}(P)$ can be expressed in terms of $N\times N$ matrix multiplication:
\begin{equation*}
\begin{split}
\nabla F_1(P)&=-\mbox{vec}(\Delta^T) -2(L^T_H\otimes L^T_G)\mbox{vec}(P)\\
&=-\mbox{vec}(\Delta^T) -2\mbox{vec}(L^T_GP L_H).
\end{split}
\end{equation*}
The same thing may be done for the gradient of the convex component
\begin{equation*}
\begin{split}
\nabla F_0(P)&=\nabla [\mbox{vec}(P)^TQ\mbox{vec}(P)]\\
\mbox{ where }& Q=(I\otimes A_G -A_H^T\otimes I)^T(I\otimes A_G -A_H^T\otimes I)\\
\nabla F_0(P)&=2Q\mbox{vec}(P)\\
&=2\mbox{vec}(A_G^2 P-A^T_GPA^T_H-A_GPA_H+PA_H^2)
\end{split}
\end{equation*}

\paragraph{Initialization}
The proposed algorithm can be accelerated by the application of Newton algorithm as the first step of \textbf{QCV} (minimization of $F_0(P)$). First, let us rewrite the \textbf{QCV} problem as follows:
\begin{eqnarray}
\nonumber
&&\min_{P \in \mathcal D} \  \|A_GP-PA_H\|_F^2 \Leftrightarrow\\
\label{eq:syseq_newton}
&&\min_{P \in \mathcal D} \ \mbox{vec}(P)^TQ\mbox{vec}(P)\Leftrightarrow
\left\{
\begin{matrix} 
\min_P \mbox{vec}(P)^TQ\mbox{vec}(P)\\
\mbox{ such that }\\ 
B\mbox{vec}(P)=1_{2N}\\
\mbox{vec}(P)\geq 0_{N^2}\\
\end{matrix}\right.
\end{eqnarray}
where $B$ is the matrix which codes the conditions $\displaystyle \sum_j P_{i,j}=1$ and $\displaystyle \sum_i P_{i,j}=1$. The Lagrangian has the following form
\begin{equation*}
\begin{split}
{\mathcal L}(P,\nu,\lambda)=&\mbox{vec}(P)^TQ\mbox{vec}(P)+\nu^T(B\mbox{vec}(P)\\
&-1_{2N})+\mu^T\mbox{vec}(P),
\end{split}
\end{equation*} 
where $\nu$ and $\mu$ are Lagrange multipliers. Now we would like to use Newton method for constrained optimization \cite{bertsekas_nonlinear_optimization} to solve (\ref{eq:syseq_newton}). Let $\mu_a$ denote the set of variables associated to the set of active constraints $\mbox{vec}(P)=0$ at the current points, then the Newton step consist in solving the following system of equations:
\begin{equation}
\left[\begin{array}[c]{lll}
2Q & B^T&I_a \\
B & 0&0\\
I_a&0&0
\end{array}\right]
\left [
\begin{array}[c]{l}
\mbox{vec}(P) \\
\nu\\
\mu_a
\end{array}\right ]=
\left [
\begin{array}[c]{l}
0 \\
1\\
0
\end{array}\right ]
\begin{array}[c]{l}
N^2~\mbox{elements,} \\
2N ~\mbox{elements,}\\
\mbox{\footnotesize\# of act. ineq. cons.}
\end{array}
\label{eq:KKT_newton_full}
\end{equation}
More precisely we have to solve (\ref{eq:KKT_newton_full}) for $P$. The problem is that in general situations this problem is computationally demanding because it involves the inversion of  matrices of size $O(N^2)\times O(N^2)$. In some particular cases, however, the Newton step becomes feasible. Typically, if none of the constraints  $\mbox{vec}(P) \geq 0$ are active, then (\ref{eq:KKT_newton_full}) takes the following form\footnote{It is true if we start our algorithm, for example, from the constant matrix $P_0=\frac1N 1_N1_N^T$.}:
\begin{equation}
\left[\begin{array}[c]{ll}
2Q & B^T \\
B & 0
\end{array}\right]
\left [
\begin{array}[c]{l}
\mbox{vec}(P) \\
\nu
\end{array}\right ]=
\left [
\begin{array}[c]{l}
0 \\
1
\end{array}\right ]~~
\begin{array}[c]{l}
N^2~\mbox{elements}\,, \\
2N ~\mbox{elements}\,.
\end{array}
\label{eq:KKT_newton}
\end{equation}
The solution is then obtained as follows:
\begin{equation}
\mbox{vec}(P)_{KKT}=\frac{1}{2}Q^{-1}B^T(BQ^{-1}B^T)^{-1}1_{2N}.
\label{eq:P_KKT}
\end{equation}
Because of the particular form of matrices $Q$ and $B$, the expression  (\ref{eq:P_KKT}) may be computed very simply with the help of Kronecker product properties in $O(N^3)$ instead of $O(N^6)$. More precisely, the first step is the calculation of $M=BQ^{-1}B^T$ where $Q=(I\otimes A_G-A^T_H\otimes I)^2$. The matrix $Q^{-1}$ may be represented as follows:
\begin{equation}
Q^{-1}=(U_H\otimes U_G)(I\otimes \Lambda_G-\Lambda_H \otimes I)^{-2}(U_H\otimes U_G)^T.
\label{eq:Q_spectral}
\end{equation}
Therefore the $(i,j)$-th element of $M$ is the following product:
\begin{equation}
\begin{split}
B_iQ^{-1}B_j^T=\mbox{vec}(U_H^T\widetilde{B_i}^TU_G)^T)&(\Lambda_G-\Lambda_H)^{-2}\\
&\times \mbox{vec}(U_G^T\widetilde{B_j}^TU_H)\,,
\end{split}
\end{equation}
where $B_i$ is the $i$-th row of $B$ and $\widetilde{B_i}$ is $B_i$  reshaped into a $N\times N$ matrix. The second step is an inversion of the $2N\times2N$ matrix $M$, and a sum over columns $M^s=M^{-1}1_{2N}$. The last step is a multiplication of $Q^{-1}$ by $B^TM^s$, which can be done with the same tricks as the first step. The result is the value of matrix $P_{KKT}$. We then have two possible scenarios:
\begin{enumerate}
\item If $P_{KKT} \in {\mathcal D}$, then we have found the solution of (\ref{eq:syseq_newton}).
\item Otherwise we take the point of intersection of the line $(P_0,P_{KKT})$ and the border $\partial {\mathcal D}$ as the next point and we continue with Frank-Wolfe algorithm. Unfortunately we can do the Newton step only once, then some of $P\geq0$ constraints become active and efficient calculations are not feasible anymore. But even in this case the Newton step is generally very useful because it decreases a lot the value of the objective function. 
\end{enumerate}

\paragraph{$d\lambda$-adaptation strategy}

In practice, we found it useful to have the parameter $d\lambda$ in the algorithm of Figure \ref{fig:algo_schema} vary between iterations. Intuitively, $d\lambda$ should depend on the form of the objective function as follows: if $F_\alpha^\lambda(P)$ is smooth and if increasing the parameter $\lambda$ does not change a lot the form of the function, then we can afford large steps, in contrast, we should do a lot of small steps in the situation where the objective function is very sensitive to changes in the parameter $\lambda$. The adaptive scheme we propose is the following. First, we fix a constant $d\lambda_{min}=10^{-5}$, which represents the lower limit for $d\lambda$. When the PATH algorithm starts, $d\lambda$ is set to $d\lambda_{min}$. If we see after an update $\lambda_{new}=\lambda+d\lambda$ that  $\vert F_{\lambda_{new}}(P^*(\lambda))-F_{\lambda}(P^*(\lambda))\vert \leq \epsilon_{\lambda}$ then we double $d\lambda$ and keep multiplying $d\lambda$ by $2$ as long as $\vert F_{\lambda_{new}}(P^*(\lambda))-F_{\lambda}(P^*(\lambda))\vert \leq \epsilon_{\lambda}$. On the contrary, if $d\lambda$ is too large in the sense that $\vert F_{\lambda_{new}}(P^*(\lambda))-F_{\lambda}(P^*(\lambda))\vert >\epsilon_{\lambda}$, then we divide $d\lambda$ by $2$ until the criterion $\vert F_{\lambda_{new}}(P^*(\lambda))-F_{\lambda}(P^*(\lambda))\vert \leq \epsilon_{\lambda}$ is met, or $d\lambda = d\lambda_{min}$. Once the update on $d\lambda$ is done, we run the optimization (Frank-Wolfe) for the new value $\lambda+d\lambda$. The idea behind this simple adaptation schema is to choose  $d\lambda$ which keeps $\vert F_{\lambda_{new}}(P^*(\lambda))-F_{\lambda}(P^*(\lambda))\vert$ just below $\epsilon_{\lambda}$.

\paragraph{Stopping criterion}
\label{sec:stop_criterion}

The choice of the update criterion $\vert F_{\lambda_{new}}(P^*(\lambda))-F_{\lambda}(P^*(\lambda))\vert$ is not unique.  Here we check whether the function value has been changed a lot at the given point. But in fact it may be more interesting to trace the minimum of the objective function. To compare the new minimum with the current one, we need to check the distance between these minima and the difference between function values. It means that we use the following condition as the stopping criterion
\begin{equation*}
\begin{split}
\vert F_{\lambda_{new}}(P^*(\lambda_{new}))-F_{\lambda}(P^*(\lambda))\vert<\epsilon_\lambda^F \ \ \mbox{and} \ \\ 
\ \ \ \vert\vert P^*(\lambda_{new})-P^*(\lambda)\vert \vert<\epsilon_\lambda^P
\end{split}
\end{equation*}

Although this approach takes a little bit more computations (we need to run Frank-Wolfe on each update of $d\lambda$), it is quite efficient if we use the adaptation schema for $d\lambda$. 

To fix the values $\epsilon^F_{\lambda}$ and $\epsilon^P_{\lambda}$ we use a parameter $M$ which defines a ratio between these parameters and the parameters of the stopping criterion used in the Frank-Wolfe algorithm: $\epsilon^F_{FW}$ (limit value of function decrement) and $\epsilon^P_{FW}$ (limit value of argument changing): $\epsilon^F_\lambda=M\epsilon^F_{FW}$ and $\epsilon^P_\lambda=M\epsilon^P_{FW}$. The parameter $M$ represents an authorized level of stopping criterion relaxation when we increment $\lambda$. In practice, it means that when we start to increment $\lambda$ we may move away from the local minima and the extent of this move is defined by the parameter $M$. The larger the value of $M$, the further we can move away and the larger $d\lambda$ may be used.  
In other words, the parameter $M$ controls the width of the tube around the path of optimal solutions.

\subsection{Algorithm complexity}
\label{sec:complexity}
Here we present the complexity of the algorithms discussed in the paper.
\begin{itemize}
\item Umeyama's algorithm has three components: matrix multiplication, calculation of eigenvectors and application of the  Hungarian algorithm for (\ref{eq:U_optimization}). Complexity of each component is equal to $O(N^3)$. Thus Umeyama's algorithm has complexity $O(N^3)$.  
\item LP approach (\ref{eq:adj_distance_l1}) has complexity  $O(N^7)$ (worst case) because it may be rewritten as an linear optimization problem with $3N^2$ variables~\cite{Boyd2003Convex}.
\end{itemize}

In the PATH algorithm, there are three principal parameters which have a big impact on the algorithm complexity. These parameters are $\epsilon^{F}_{FW}$, $\epsilon^{P}_{FW}$, $M$ and $N$. The first parameter $\epsilon_{FW}$  defines the precision of the Frank-Wolfe algorithm, in some cases its speed may be sublinear \cite{bertsekas_nonlinear_optimization}, however it should work much better when the optimization polytope has a ``smooth'' border~\cite{bertsekas_nonlinear_optimization}. \skipp{It seems that the polytope of permutation matrices is more or less smooth, but it is hard to prove it exactly.}

The influence of the ratio parameter $M$  is more complicated. In practice, in order to ensure that the objective function takes values between $0$ and $1$, we usually use the normalized version of the objective function:
\begin{equation*}
F_{norm}(P)=\frac{||A_GP-PA_H||_F^2}{||A_G||_F^2+||A_H||_F^2}
\end{equation*}
In this case if we use  the simple stopping criterion based on the value of the objective function then the number of iteration over $\lambda$ (number of Frank-Wolfe algorithm runs) is at least equal to $\frac{C}{M\epsilon_{FW}^F}$ where $C=\min\limits_\Pcal F_{norm}-\min\limits_\Dcal F_{norm}$. 

The most important thing  is how the algorithm complexity depends on the graph size $N$. In general the number of iterations of the Frank-Wolfe algorithm scales as $O\left(\frac{\kappa}{\epsilon^F_{FW}}\right)$ where $\kappa$ is the conditional number of the Hessian matrix describing the objective function near a local minima \cite{bertsekas_nonlinear_optimization}. It means that in terms of numbers of iterations, the parameter $N$ is not crucial. $N$ defines the dimensionality of the minimization problem, while $\kappa$ may be close to zero or one depending on the graph structures, not explicitly on their size. On the other hand, $N$ has a big influence on the cost of one iteration. Indeed, in each iteration step we need to calculate the gradient and to minimize a linear function over the polytope of doubly stochastic matrices. The gradient estimation and the minimization may be done in $O(N^3)$. In Section \ref{sec:synres} we present the empirical results on how algorithm complexity and optimization precision depend on $M$ (Figure \ref{fig:path_m_d}b) and $N$ (Figure \ref{fig:timing}).

\subsection{Vertex pairwise similarities}
\label{sec:addlabels}
If we match two labeled graphs, then we may increase the performance of our method by using information on pairwise similarities between their nodes. In fact one method of image matching   uses only this type of information, namely shape context matching \cite{belongie_shape_matching_shape_context}. To integrate the information on vertex similarities we use the approach proposed in (\ref{eq:F_0_alpha}), but in our case we use $F_\lambda(P)$ instead of $F_0(P)$
\begin{equation}
\min_{P} \ F_{\lambda}^{\alpha}(P)=\min_{P} \ (1-\alpha)F_\lambda(P)+\alpha\mbox{tr}(C^TP),.
\label{eq:F_lambda_alpha}
\end{equation}
The advantage of the last formulation is that $F_{\lambda}^{\alpha}(P)$ is just $F_\lambda(P)$ with an additional linear term. Therefore  we can use the same algorithm for the minimization of $F_{\lambda}^{\alpha}(P)$ as the one we presented for the minimization of $F_{\lambda}(P)$.

\subsection{Matching  graphs of different sizes}
\label{sec:dummy_nodes}
Often in practice we have to match graphs of different sizes $N_G$ and $N_H$ (suppose, for example, that $N_G>N_H$). In this case we have to match all vertices of graph $H$ to a subset of vertices of graph $G$. In the usual case when $N_G=N_H$, the error (\ref{eq:adj_distance}) corresponds to the number of mismatched edges (edges which exist in one graph and do not exist in the other one). When we match graphs of different sizes the situation is a bit more complicated. Let $V_G^+\subset V_G$ denote the set of vertices of graph $G$ that are selected for matching to vertices of graph $H$, let $V_G^-=V_G\setminus V_G^+$ denote all the rest. Therefore all edges of the graph $G$ are divided into four parts $E_G=E_G^{++}\cup E_G^{+-}\cup E_G^{-+}\cup E_G^{--}$, where $E_G^{++}$ are edges between vertices from $V_G^+$, $E_G^{--}$ are edges between vertices from $V_G^-$, $E_G^{+-}$ and  $E_G^{+-}$ are edges from $V_G^+$ to $V_G^-$ and from $V_G^-$ to $V_G^+$ respectively. For undirected graphs the sets $E_G^{+-}$ and  $E_G^{+-}$ are the same (but, for directed graphs we do not consider, they would be different). The edges from $E_G^{--}$, $E_G^{+-}$ and $E_G^{-+}$ are always mismatched and a question is whether we have to take them into account in the objective function or not. According to the answer we have three types of matching error (four for directed graphs) with interesting interpretations.
\begin{enumerate}
\item We count only the number of mismatched edges between $H$ and the chosen subgraph  $G^+\subset G$. It corresponds to the case when  the matrix $P$ from (\ref{eq:adj_distance}) is a matrix of size $N_G\times N_H$ and $N_G-N_H$ rows of $P$  contain only zeros. 
\item We count the number of mismatched edges between $H$ and the chosen subgraph  $G^+\subset G$. And we also count all edges from $E_G^{--}$, $E_G^{+-}$ and $E_G^{-+}$. In this case $P$ from (\ref{eq:adj_distance}) is a matrix of size $N_G\times N_G$. And we transform matrix $A_H$ into a matrix of size of size $N_G\times N_G$ by adding $N_G-N_H$ zero rows and zero columns. It means that we add dummy isolated vertices to the smallest graph, and then we match graphs of the same size.
\item We count the number of mismatched edges between $H$ and chosen subgraph  $G^+\subset G$. And we also count all edges from $E_G^{+-}$ (or $E_G^{-+}$). It means that we count matching error between $H$ and $G^+$ and we count also the number of edges which connect $G^+$ and $G^-$. In other words we are looking for subgraph $G^+$ which is similar to $H$ and which is maximally isolated in the graph $G$.    
\end{enumerate}

Each type of error may be useful according to context and interpretation, but a priori, it seems that the best choice is the second one where we add dummy nodes to the smallest graph. The main reason is the following. Suppose that graph $H$ is quite sparse, and suppose that graph $G$ has two candidate subgraphs $G^+_s$ (also quite sparse) and $G^+_{d}$ (dense). The upper bound for the matching error between $H$ and $G_{s}^+$ is $\#V_H + \#V_{G_s^+}$, the lower bound for the matching error between $H$ and $G_{d}^+$ is $\#V_{G_d^+}-\#V_H$. So if $\#V_H+\#V_{G_s^+}<\#V_{G_d^+}-\#V_H$ then we will always choose the graph $G_s^+$ with the first strategy, even if it is not similar at all to the graph $H$. The main explanation of this effect lies in the fact that the algorithm tries to minimize the number of mismatched edges, and not to maximize the number of well matched edges. In contrast, when we use dummy nodes, we do not have this problem because if we take a very sparse subgraph $G^{+}$ it increases the number of edges in $G^{-}$(the common number of edges in $G^+$ and $G^-$ is constant and is equal to the number of edges in $G$) and finally it decreases the quality of matching.     


\section{Simulations}
\label{sec:num_exp} 

\subsection{Synthetic examples}

In this section we compare the proposed algorithm  with some classical methods on artificially generated graphs. Our choice of random graph types is based on \cite{newman_random_graphs} where authors discuss different types of random graphs which are the most frequently observed in various real world applications (world wide web, collaborations networks, social networks, etc...). Each type of random graphs is defined by the distribution function of node degree ${\rm Prob}(\mbox{node degree}=k)=VD(k)$. The vector of node degrees of each graph is supposed to be an i.i.d sample from $VD(k)$. In our experiments we have used the following types of random graphs:
\begin{center}
\begin{tabular}{|l||l|}
\hline
Binomial law&$VD(k)=C_N^kp^k(1-p)^{1-k}$\\
\hline
Geometric law &$VD(k)=(1-e^{-\mu})e^{\mu k}$\\
\hline
Power law & $VD(k)=C_{\tau}k^{-\tau}$\\
\hline
\end{tabular}
\end{center}
The schema of graph generation is the following
\begin{enumerate}
\item generate a sample $d=(d_1,\dots,d_N)$ from $VD(k)$
\item if $\sum_i d_i$ is odd then goto step 1 
\item while $\sum_i d_i >0$ 
\begin{enumerate}
\item choose randomly two non-zero elements from $d$: $d_{n1}$ and $d_{n2}$
\item add edge $(n1,n2)$ to the graph
\item $d_{n1}\leftarrow d_{n1}-1~~~d_{n2}\leftarrow d_{n2}-1$
\end{enumerate}  
\end{enumerate}
If we are interested in isomorphic graph matching then we compare just the initial graph and its randomly permuted copy. To test the matching of non-isomorphic graphs, we add randomly $\sigma N_E$ edges to the initial graph and to its permitted copy, where $N_E$ is the number of edges in the original graph, and $\sigma$ is the noise level. 

\subsection{Results}
\label{sec:synres}
\begin{figure*}[htbp]
 \centering
 \subfigure[bin]{\includegraphics[width=5cm]{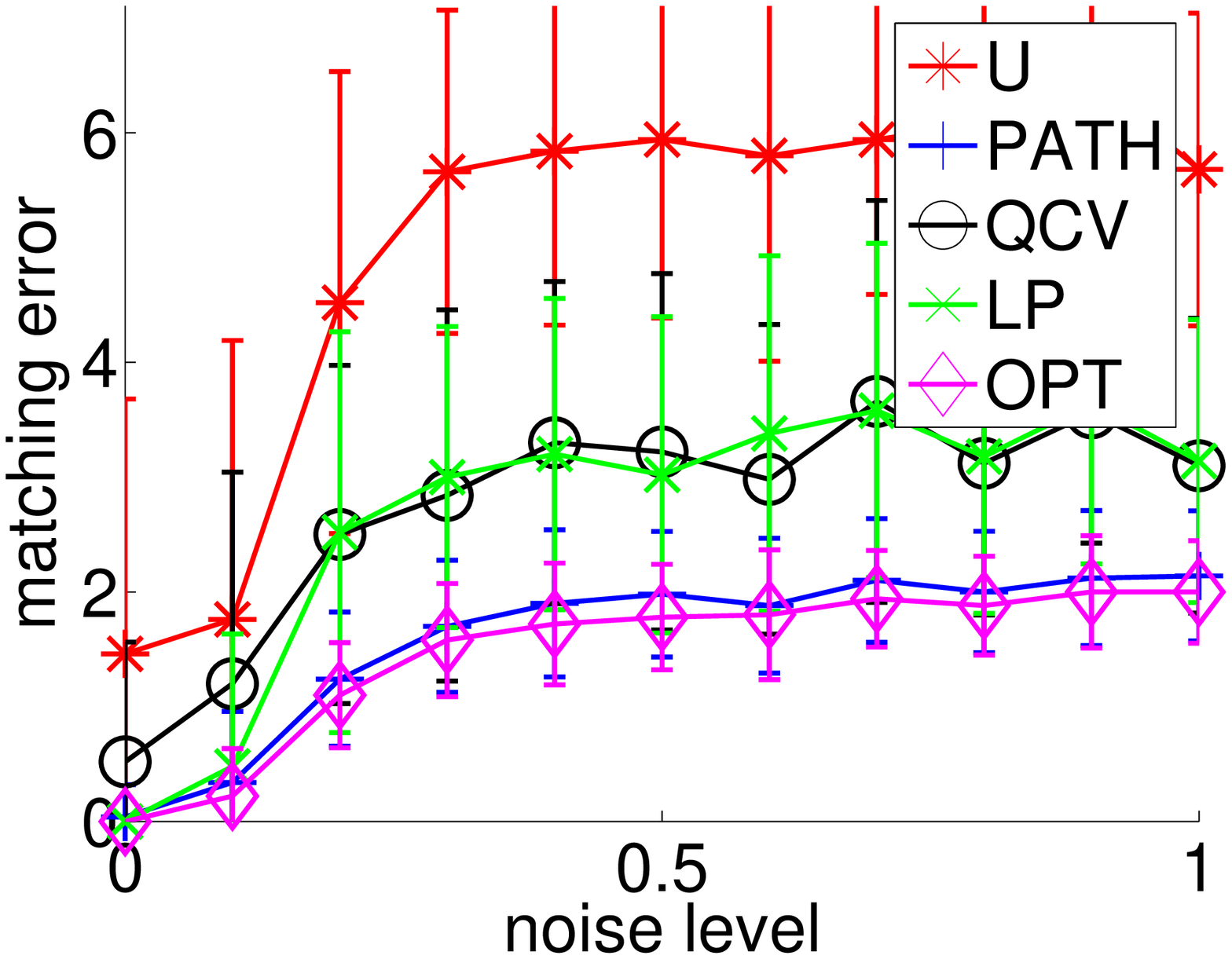}}
 \subfigure[exp]{\includegraphics[width=5cm]{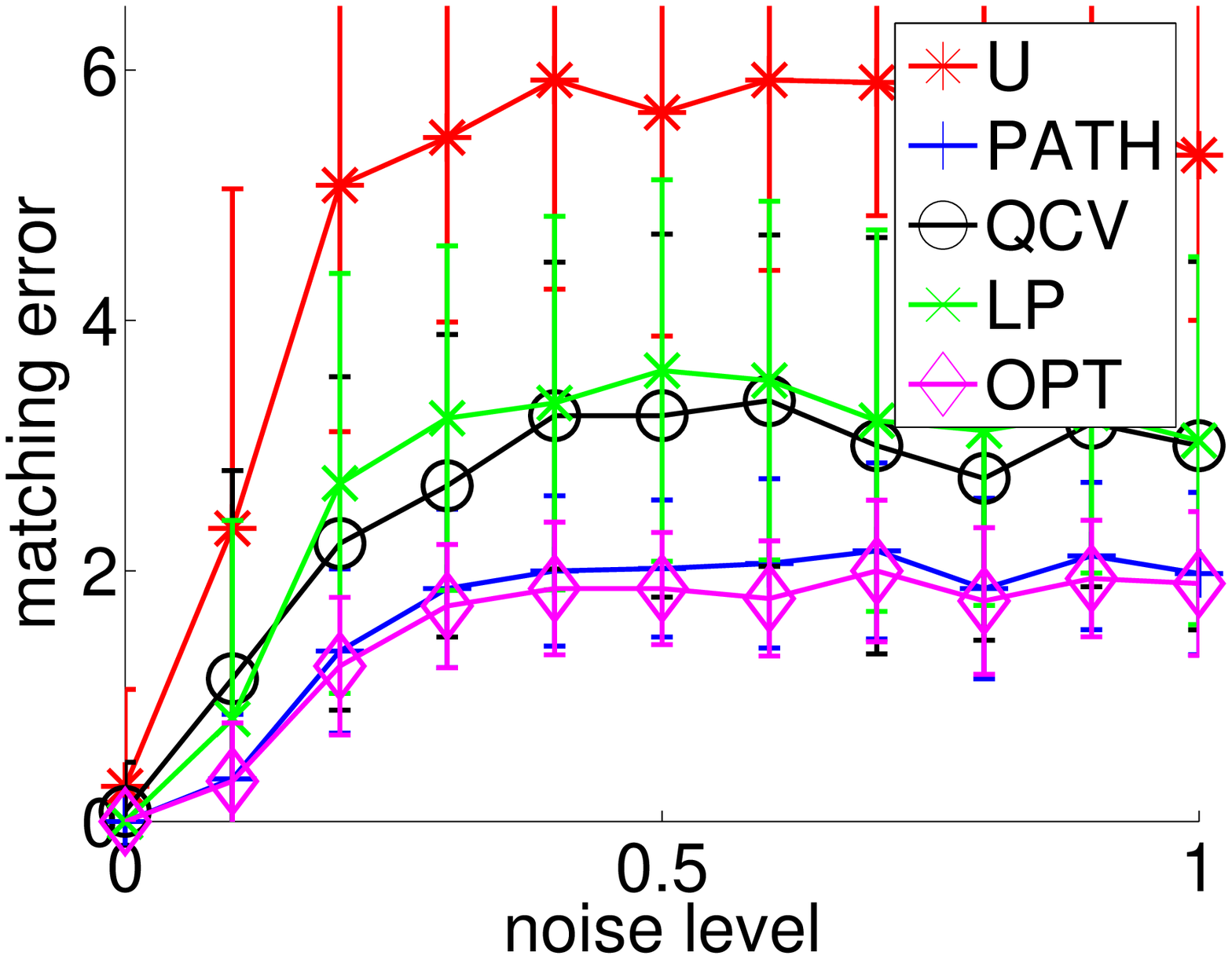}}
 \subfigure[pow]{\includegraphics[width=5cm]{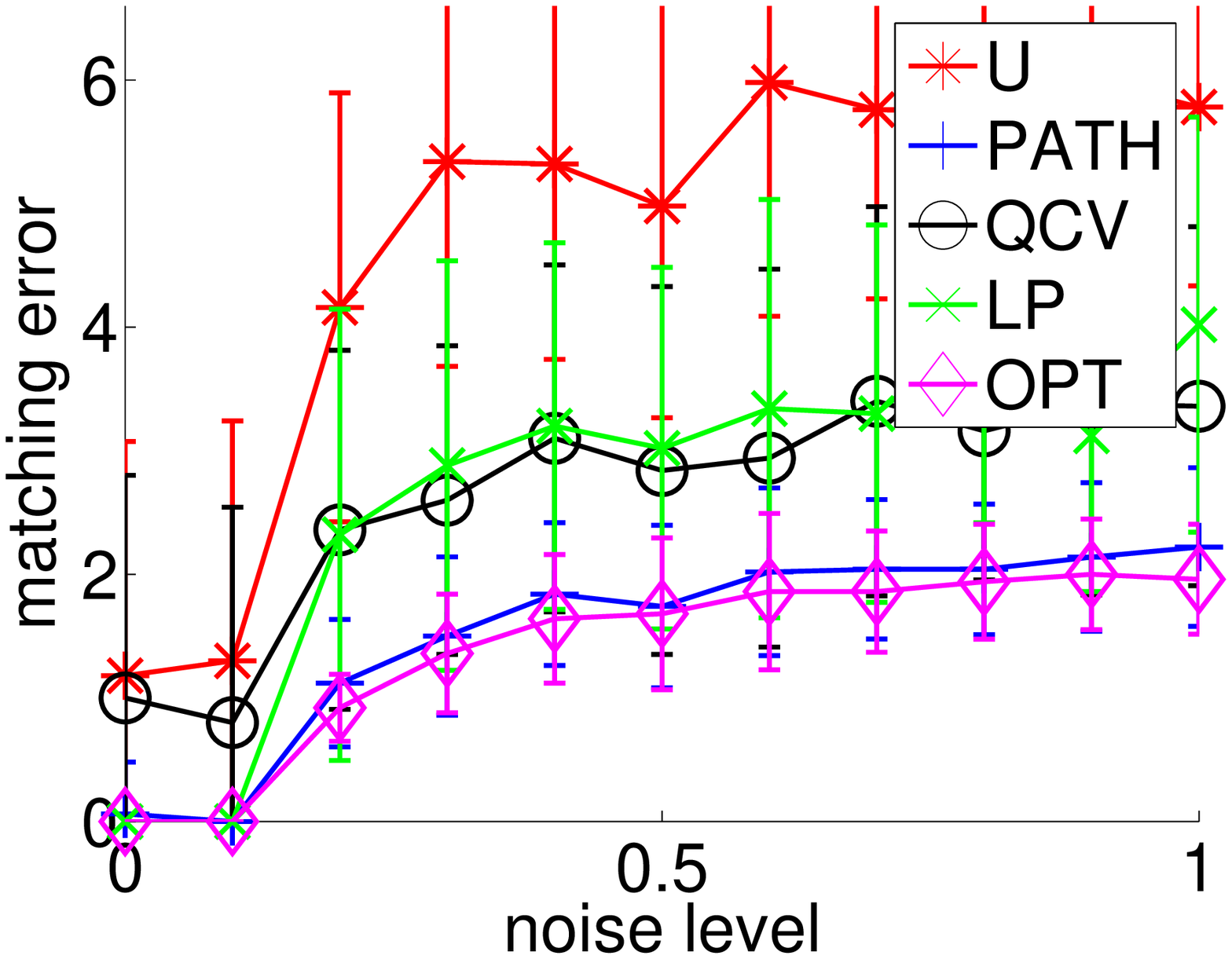}}
 \caption{Matching error (mean value over sample of size 100) as a function of noise. Graph size N=8. U --- Umeyama's algorithm, LP --- linear programming algorithm, QCV --- convex optimization, PATH --- path minimization algorithm,OPT --- an exhaustive search (the global minimum). The range of   error bars is the standard deviation of matching errors}
 \label{fig:num_exp_n_8}
\end{figure*}

\begin{figure*}[htbp]
 \centering
 \subfigure[bin]{\includegraphics[width=5cm]{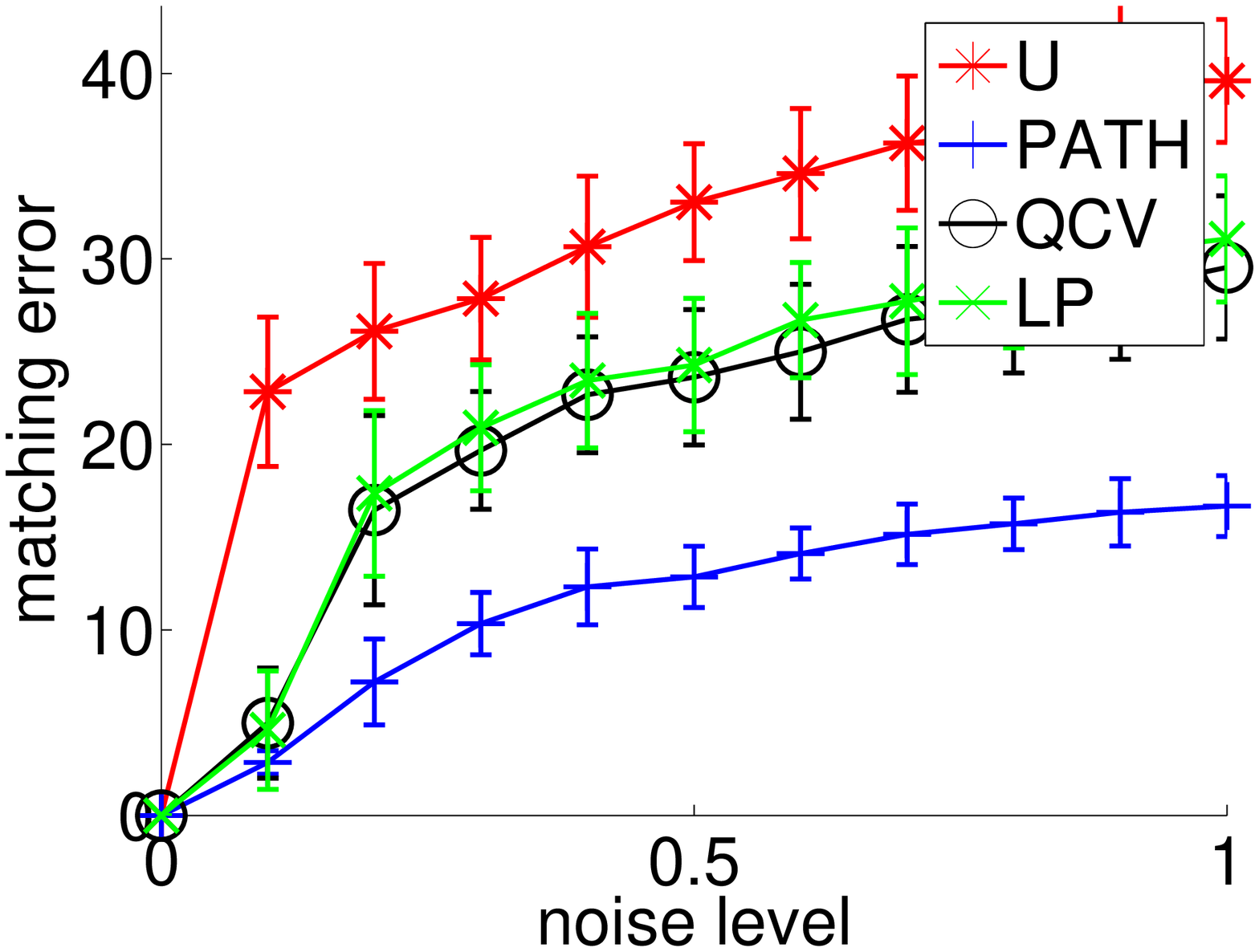}}
 \subfigure[exp]{\includegraphics[width=5cm]{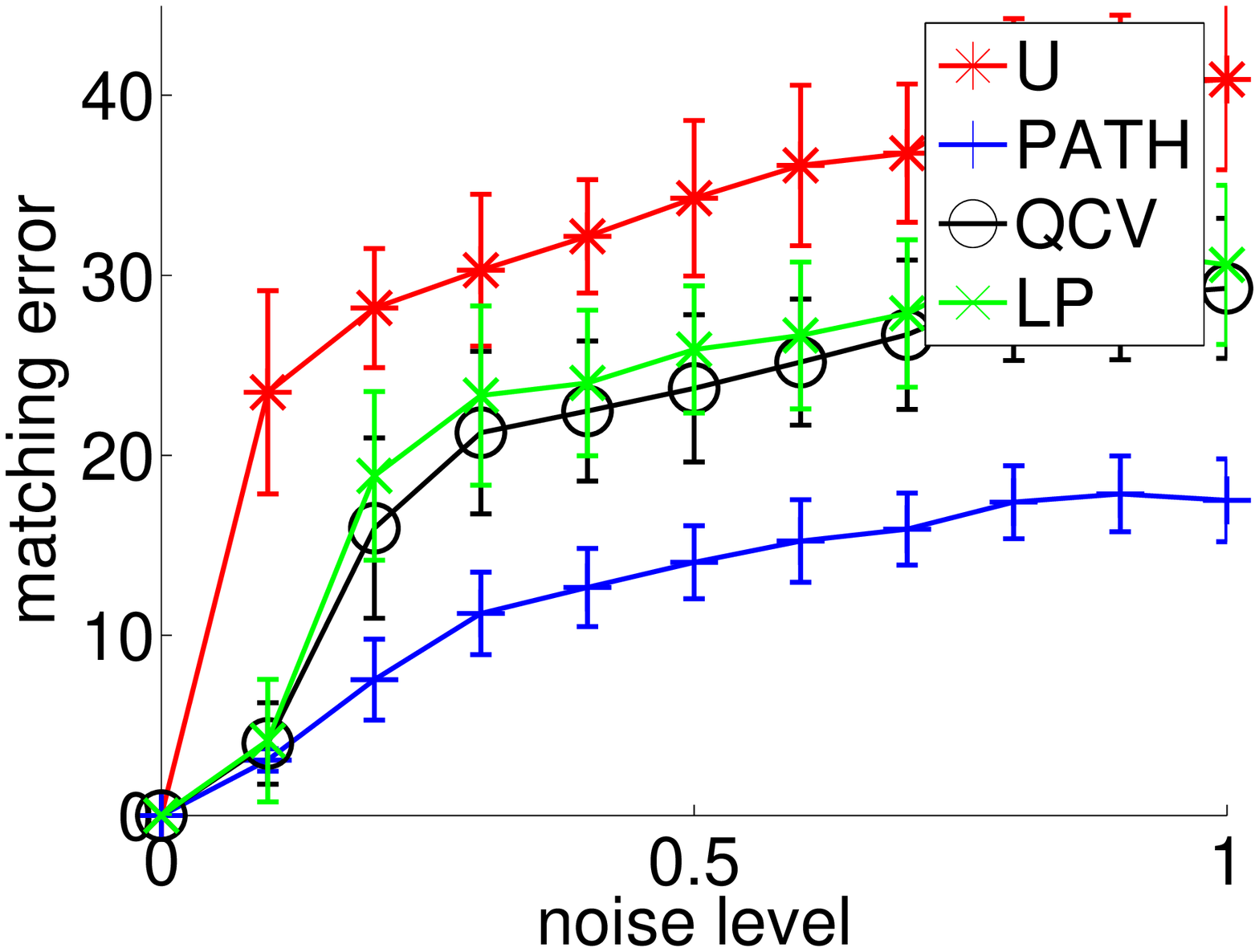}}
 \subfigure[pow]{\includegraphics[width=5cm]{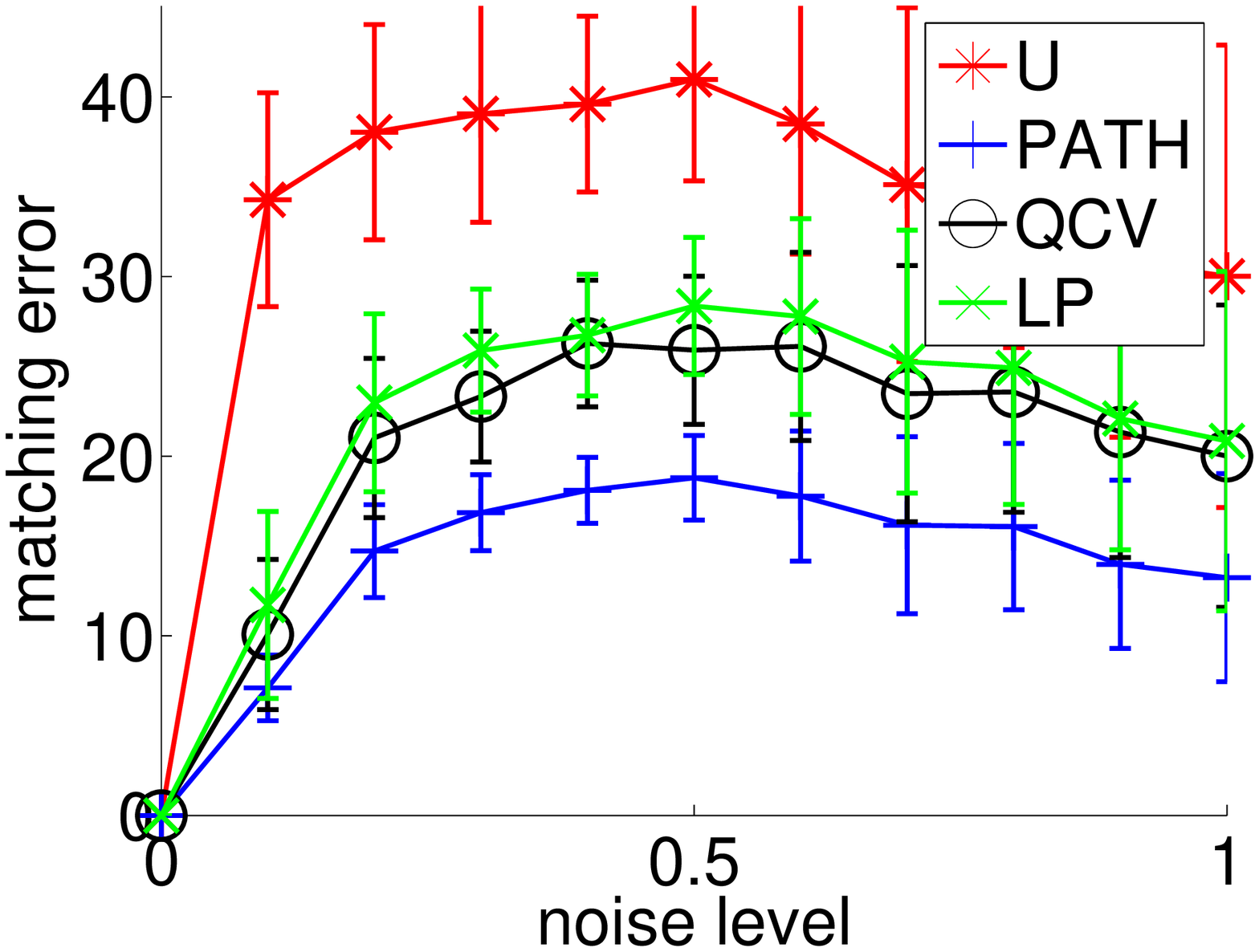}}
 \caption{Matching error (mean value over sample of size 100) as a function of noise. Graph size N=20. U --- Umeyama's algorithm, LP --- linear programming algorithm, QCV --- convex optimization, PATH --- path minimization algorithm.}
 \label{fig:num_exp_n_20}
\end{figure*}

\begin{figure*}[htbp]
 \centering
 \subfigure[bin]{\includegraphics[width=5cm]{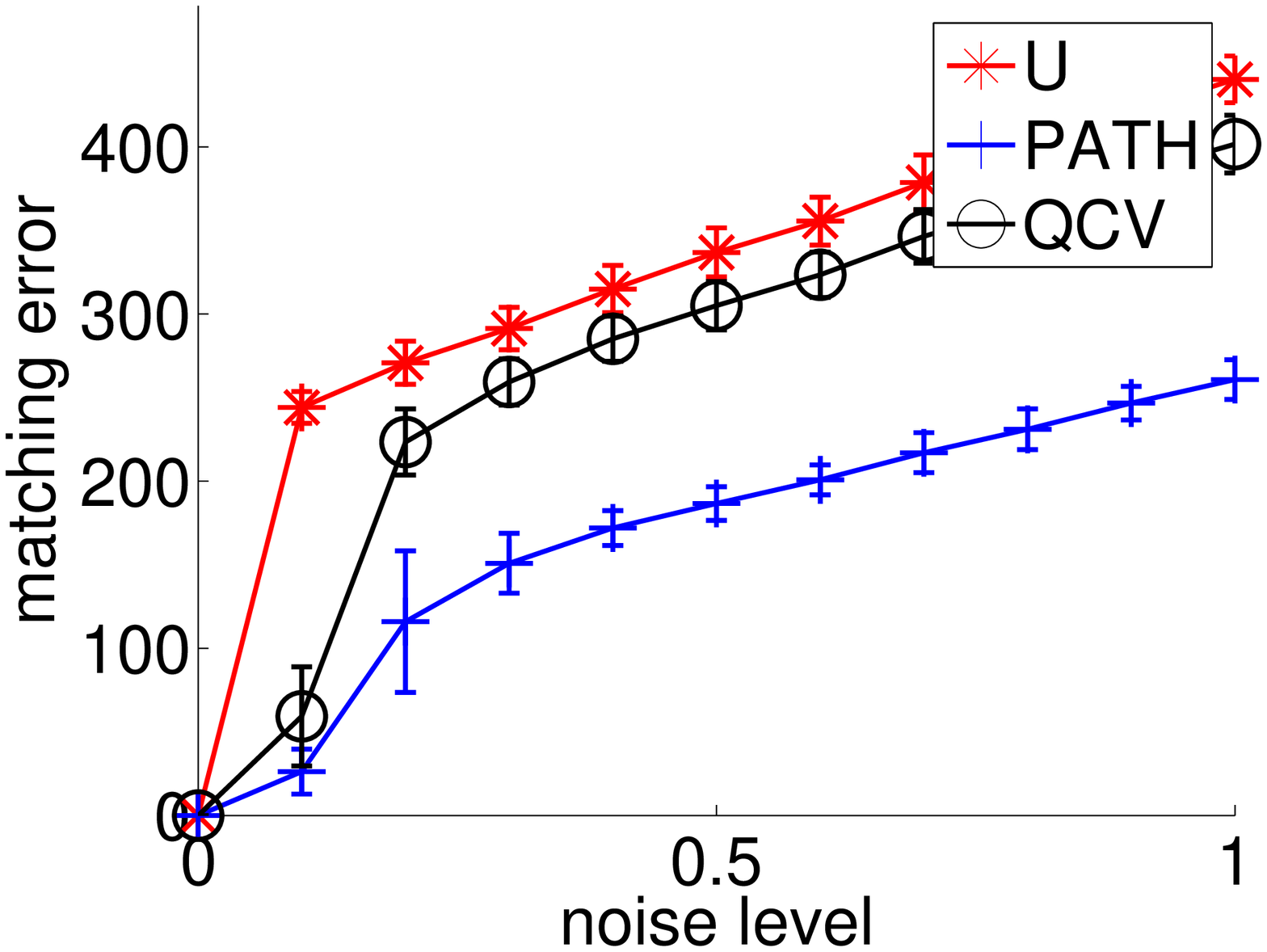}}
 \subfigure[exp]{\includegraphics[width=5cm]{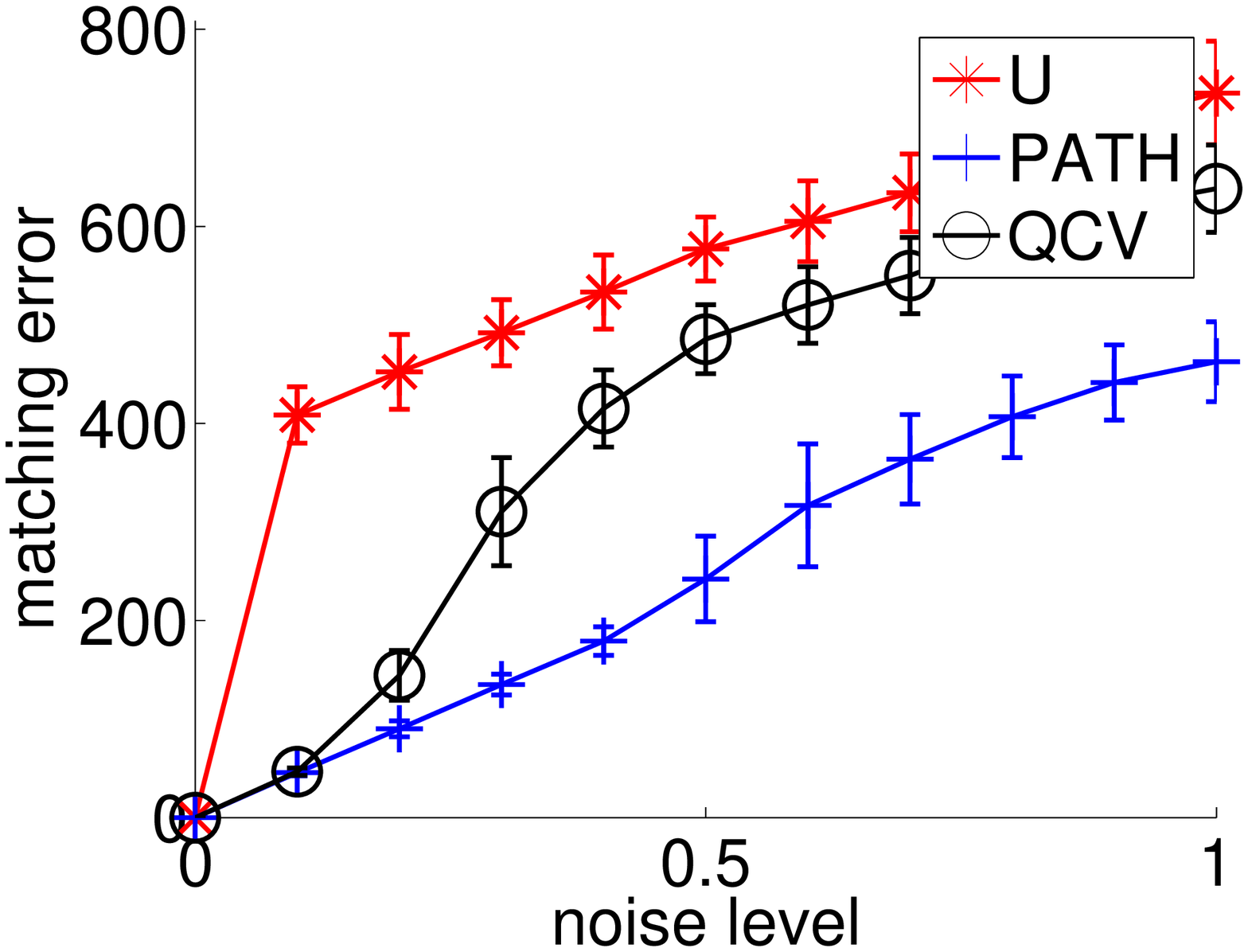}}
 \subfigure[pow]{\includegraphics[width=5cm]{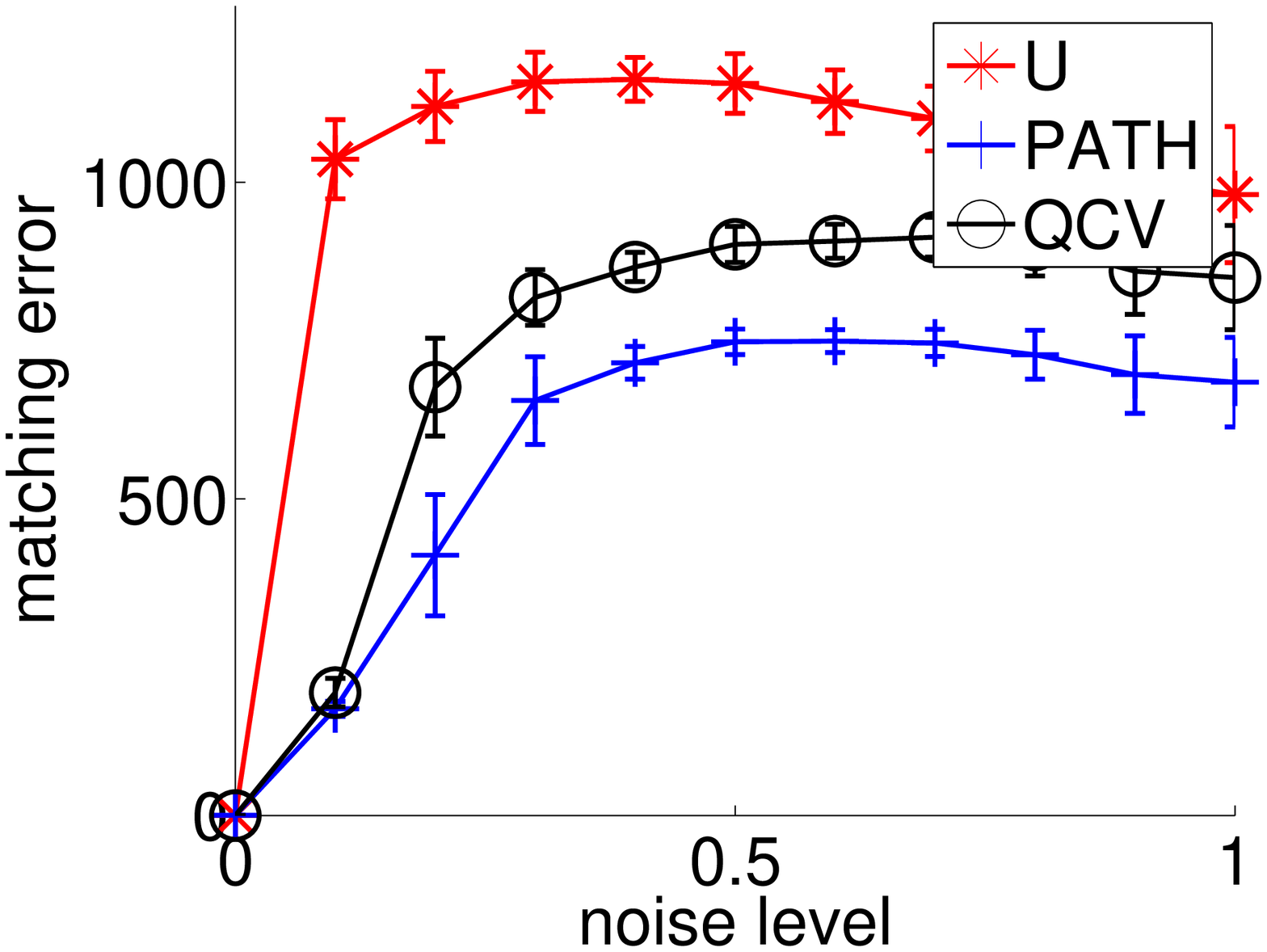}}
 \caption{Matching error (mean value over sample of size 100) as a function of noise. Graph size N=100. U --- Umeyama's algorithm, QCV --- convex optimization,  PATH --- path minimization algorithm. }
 \label{fig:num_exp_n_100}
\end{figure*}

The first series  of experiments are experiments on small size graphs (N=8), here we are interested in comparison of the PATH algorithm (see Figure \ref{fig:algo_schema}), the \textbf{QCV} approach (\ref{eq:qcv}), Umeyama spectral algorithm (\ref{eq:U_optimization}), the linear programming approach (\ref{eq:adj_distance_l1}) and exhaustive search which is feasible for the small size graphs. The algorithms were tested on the three types of random graphs (binomial, exponential and power). The results are presented in Figure \ref{fig:num_exp_n_8}.  
The same experiment was repeated for middle-sized graphs ($N=20$, Figure \ref{fig:num_exp_n_20}) and for large graphs ($N=100$, Figure \ref{fig:num_exp_n_100}). 

In all cases, the PATH algorithm works much better than all other approximate algorithms. There are some important things to note here. First, the choice of norm in (\ref{eq:adj_distance}) is not very important --- results of QCV and LP are about the same. Second, following the solution paths is very useful compared to just minimizing the convex relaxation and projecting the solution on the set of permutation matrices (PATH algorithms works much better than QCV). Another noteworthy observation is that the performance of PATH is very close to the optimal solution when the later can be evaluated.  

\begin{figure*}[htbp]
 \centering
 \subfigure[]{\includegraphics[width=6cm]{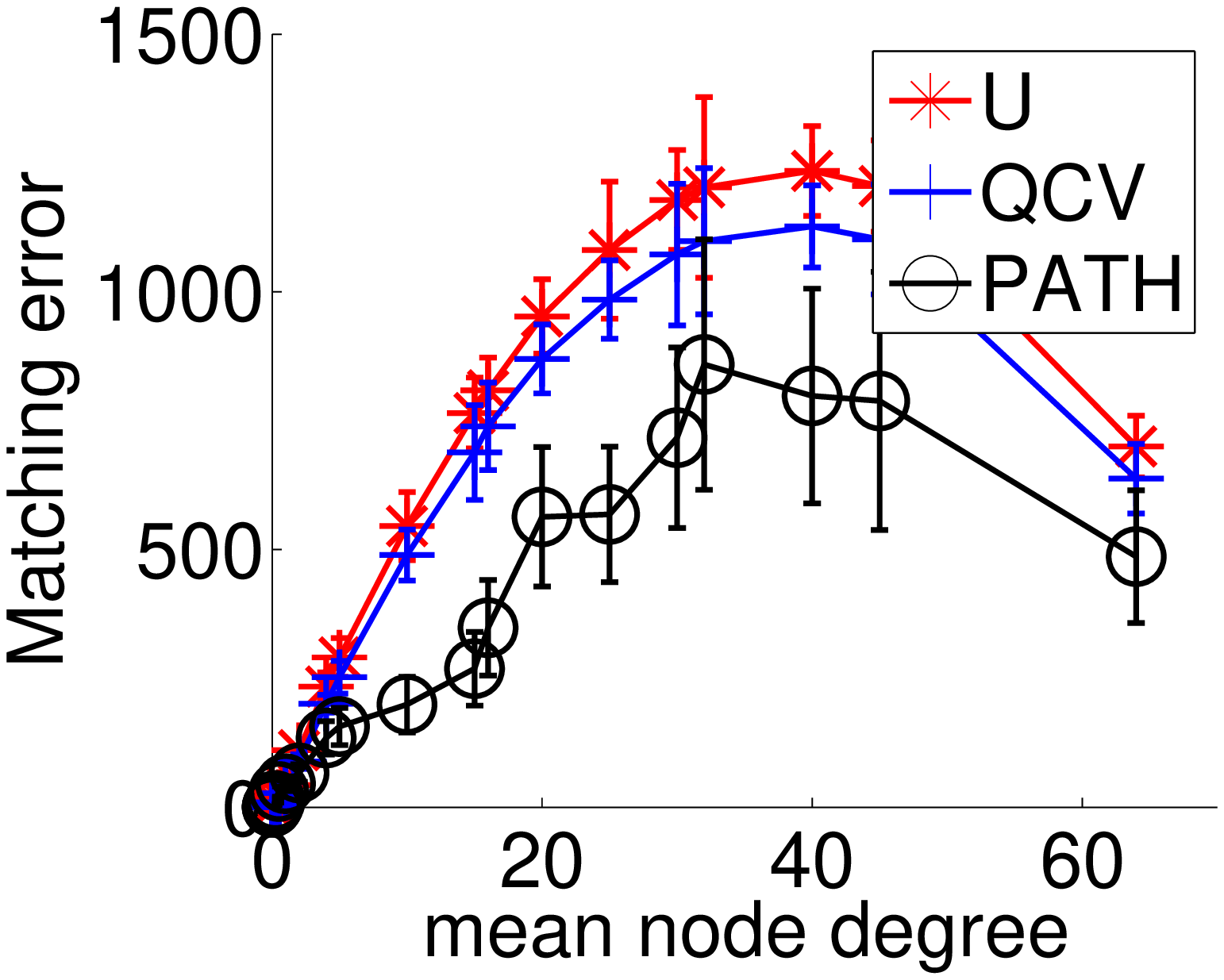}}
 \subfigure[]{\includegraphics[width=6cm]{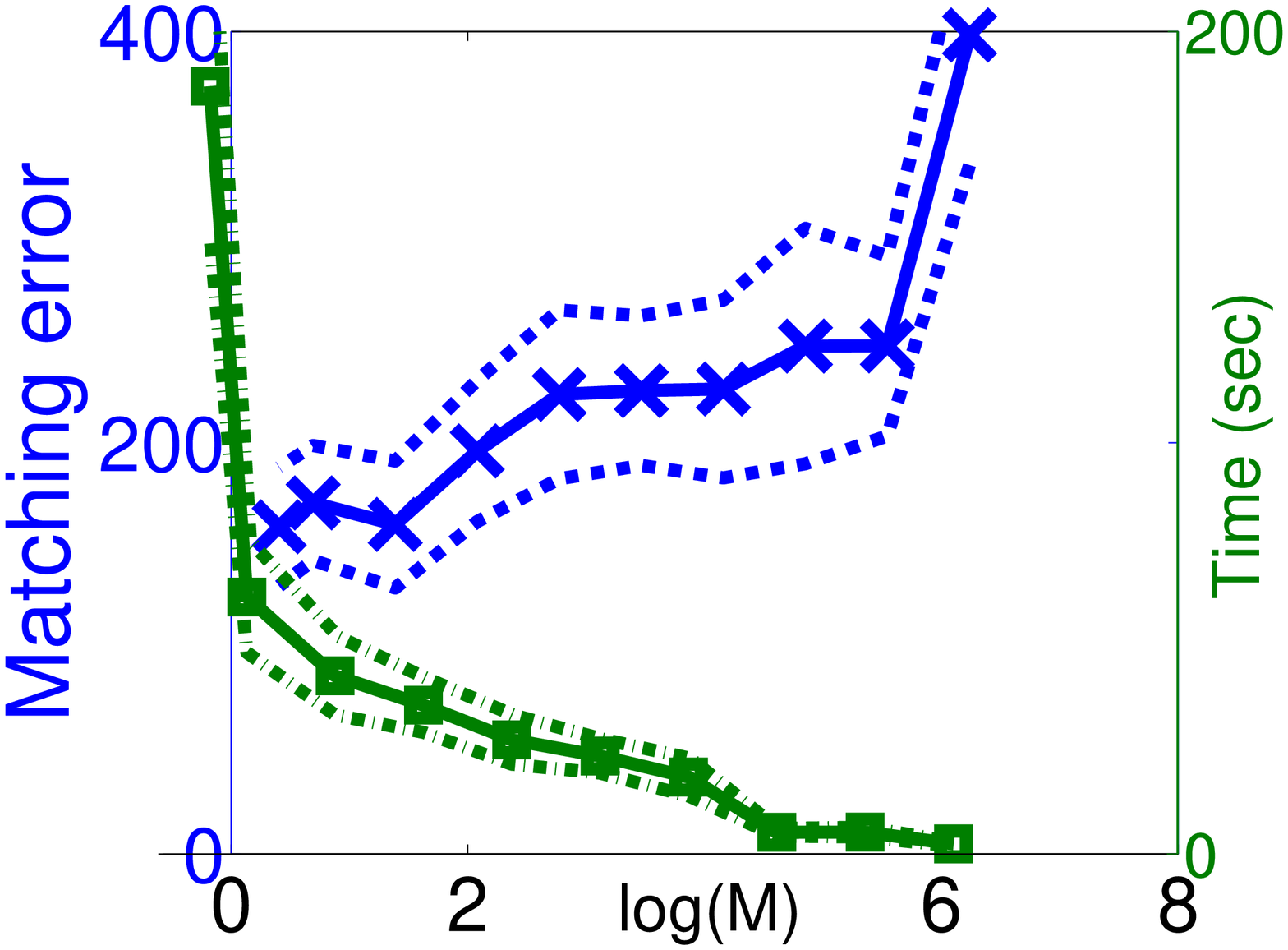}}
 \caption{(a) Algorithm performance as a function of graph density. (b) Precision and speed of the PATH algorithm as a function of $M$, the relaxation constant used in the PATH algorithm (see section \ref{sec:stop_criterion}). In both cases, graph size N=100, noise level $\sigma$=0.3, sample size is equal to 30. Error bars represent standard deviation of the matching error (not averaged)}
 \label{fig:path_m_d}
\end{figure*}

We note that sometimes the matching error decreases as the noise level increases (e.g., in Figures \ref{fig:num_exp_n_100}c,\ref{fig:num_exp_n_20}c), which can be explained as follows. The matching error is upper bounded by the minimum of the total number of zeros in the adjacency matrices $A_G$ and $A_H$, so in general this upper bound deacreases when the edge density increases. When the noise level increases, it makes graphs denser, and consequently the upper bound of matching error decreases. The general behavior of graph matching algorithms as functions of the graph density is presented in Figure \ref{fig:path_m_d}a). Here again the matching error decreases when the graph density becomes very large.

The parameter $M$ (see section \ref{sec:stop_criterion}) defines how precisely the PATH algorithm tries to follow the path of local minimas. The larger $M$, the faster the PATH algorithm. At the extreme, when $M$ is close to 1/$\epsilon_{FW}$, we jump directly from the convex function ($\lambda=0$) to the concave one ($\lambda=1$). Figure \ref{fig:path_m_d}b) shows in more details how algorithm speed and precision depend on $M$. 

\begin{figure*}[htbp]
 \centering
 \subfigure[bin]{\includegraphics[width=5cm]{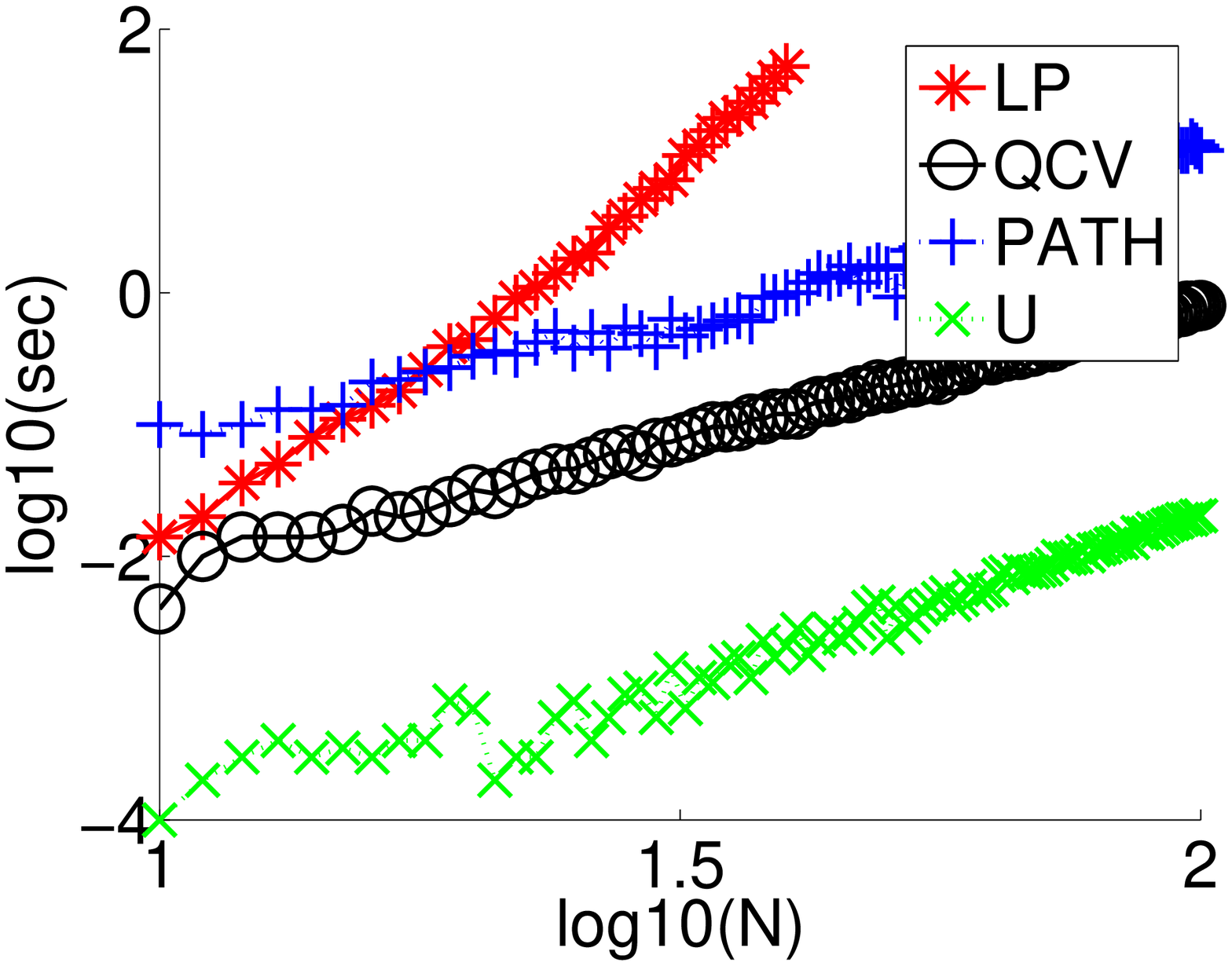}}
 \subfigure[exp]{\includegraphics[width=5cm]{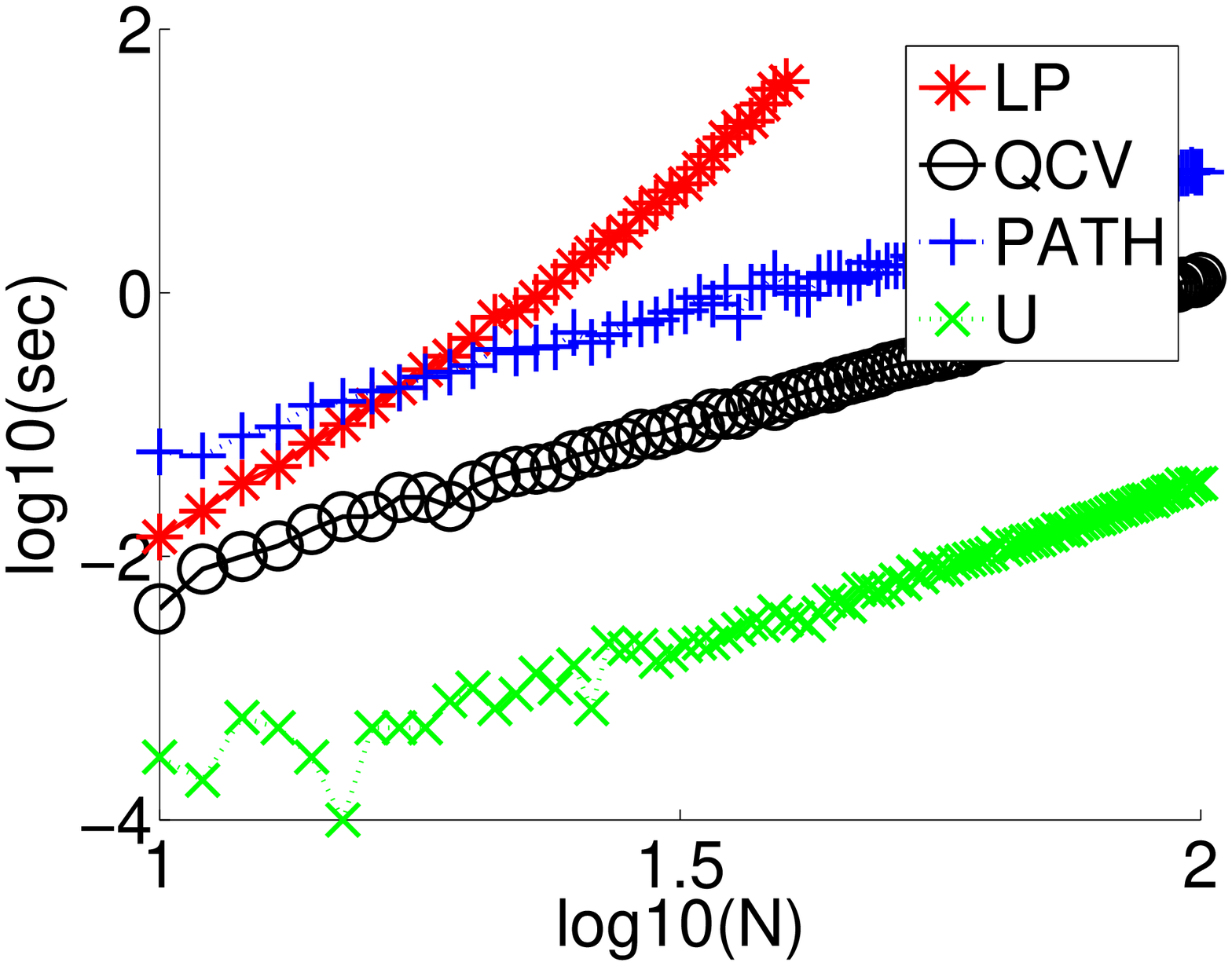}}
 \subfigure[pow]{\includegraphics[width=5cm]{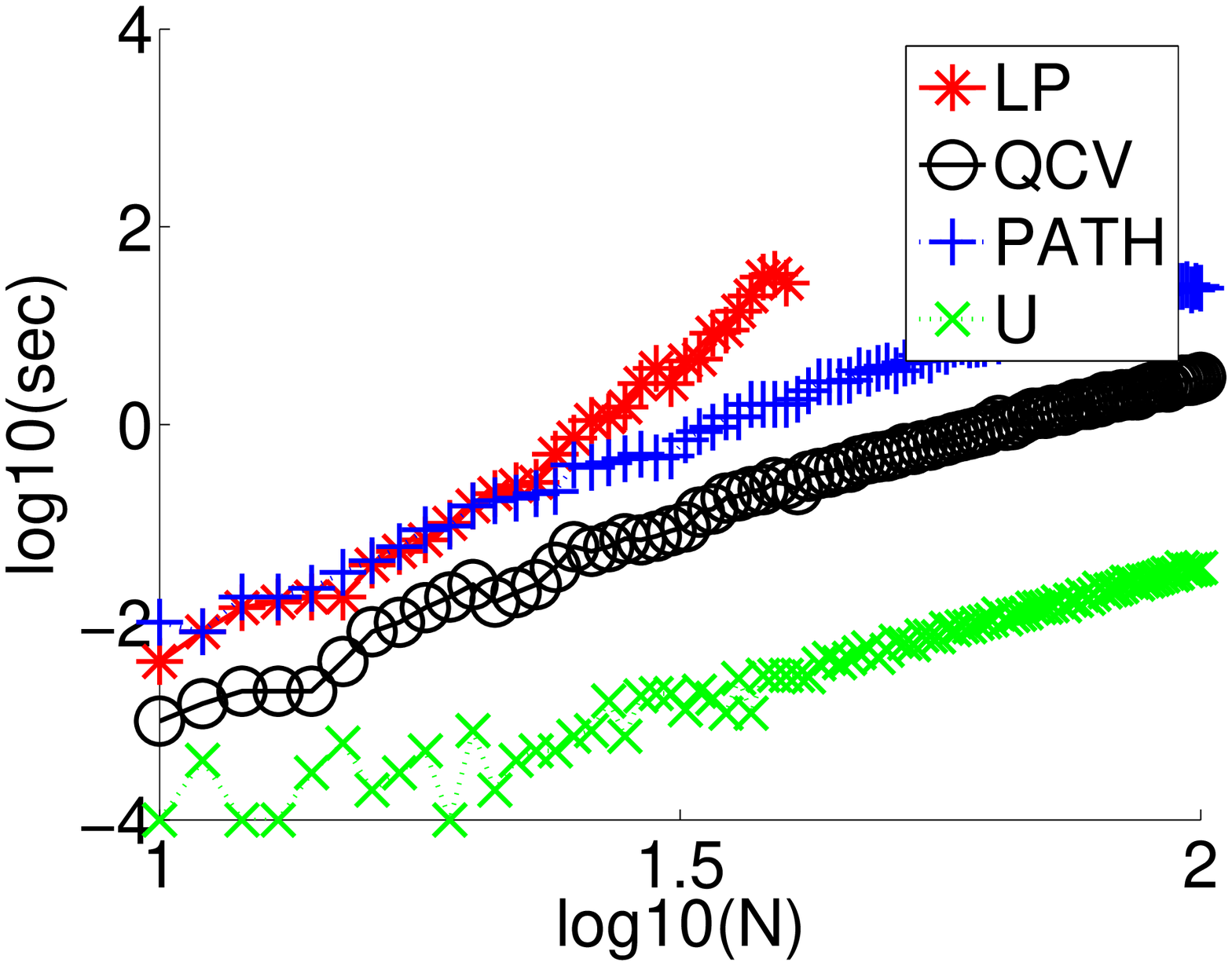}}
 \caption{Timing of U,LP,QCV and PATH algorithms as a function of graph size, for the different random graph models. LP slope $\approx$ 6.7, U, QCV and PATH slope $\approx$ 3.4}
 \label{fig:timing}
\end{figure*}

Another important aspect to compare the different algorithms is their run-time complexity as a function of $N$. Figure \ref{fig:timing} shows the time needed to obtain the matching between two graphs as a function of the number of vertices $N$ (for $N$ between $10$ and $100$), for the different methods. These curves are coherent with theoretical values of algorithm complexities summarized in Section \ref{sec:complexity}. In particular we observe that Umeyama's algorithm is the fastest method, but that QCV and PATH have the same complexity in $N$. The LP method is competitive with QCV and PATH for small graphs, but has a worse complexity in $N$.

\section{QAP benchmark library}
\label{sec:qap} 
The problem of graph matching may be considered as a particular case of  the quadratic assignment problem (QAP). The minimization of the loss function (\ref{eq:adj_distance}) is equivalent to the maximization of the following function:
\begin{equation*}
\max_{P} \ \trace(P^TA_G^TPA_H)\label{eq:qap}\,.
\end{equation*}
Therefore it is interesting to compare our method with other approximate methods proposed for QAP. \cite{convexgraphmatching} proposed the QPB algorithm for that purpose and tested it on matrices from the QAP benchmark library \cite{qap_benchmark_lib}, QPB results were compared to the results of graduated assignment algorithm GRAD\cite{Gold_graduated_graph_matching} and Umeyama's algorithm. Results of PATH application to the same matrices are presented in Table \ref{tab:qap}, scores for QPB and graduated assignment algorithm are taken directly from the publication \cite{convexgraphmatching}. We observe that on 14 out of 16 benchmark, PATH is the best optimization method among the methods tested.
\begin{table*}
\caption{Experiment results for QAPLIB benchmark datasets.}
\label{tab:qap}
\centering
\begin{tabular}{|l|l|l|l|l|l|}
\hline
QAP&MIN&PATH&QPB&GRAD&U\\
\hline \hline
chr12c& 11156&  \bf 18048&  20306&  19014&  40370\\
\hline
chr15a& 9896&   \bf 19086&  26132&  30370&  60986\\
\hline
chr15c& 9504&   \bf 16206&  29862&  23686&  76318\\
\hline
chr20b& 2298&\bf   5560&   6674&   6290&   10022\\
\hline
chr22b& 6194&\bf   8500&   9942&   9658&   13118\\
\hline
esc16b& 292&  300&\bf 296 & 298&306 \\
\hline
rou12&  235528&\bf 256320& 278834& 273438& 295752\\
\hline
rou15&  354210& 391270&\bf 381016& 457908& 480352\\
\hline
rou20&  725522&\bf 778284& 804676& 840120& 905246\\
\hline
tai10a& 135028&\bf 152534& 165364& 168096& 189852\\
\hline
tai15a& 388214&\bf 419224& 455778& 451164& 483596\\
\hline
tai17a& 491812&\bf 530978& 550852& 589814& 620964\\
\hline
tai20a& 703482&\bf 753712&799790&  871480& 915144\\
\hline
tai30a& 1818146&\bf 1903872&1996442&2077958&2213846\\
\hline
tai35a& 2422002&\bf 2555110&2720986&2803456&2925390\\
\hline
tai40a& 3139370&\bf 3281830&3529402&3668044&3727478\\
\hline
\end{tabular}
\end{table*}
\section{Image processing}
\label{sec:vision} 
In this section, we present two applications in image processing. The first one (Section \ref{sec:vision_1}) illustrates how taking into account information on graph structure may increase image alignment quality. The second one (Section \ref{sec:vision_2}) shows that the structure of contour graphs may be very important in classification tasks.  In both examples we compare the performance of our method with the shape context approach \cite{belongie_shape_matching_shape_context}, a state-of-the-art method for image matching.
\subsection{Alignment of vessel images}
\label{sec:vision_1} 
\begin{figure*}[htbp]
\centering
\includegraphics[width=6cm]{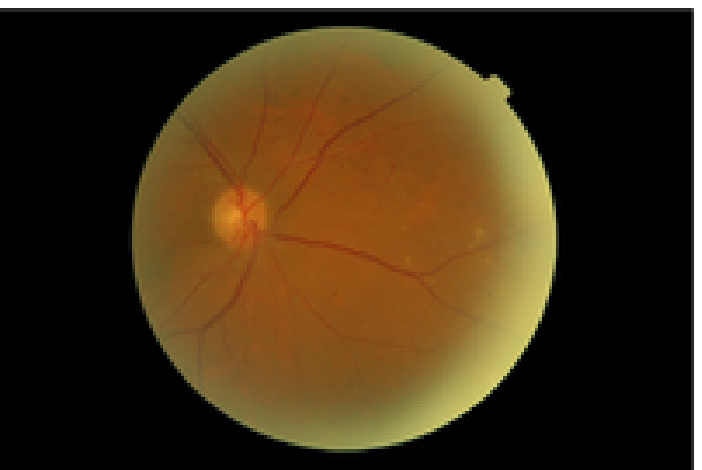}
\includegraphics[width=6cm]{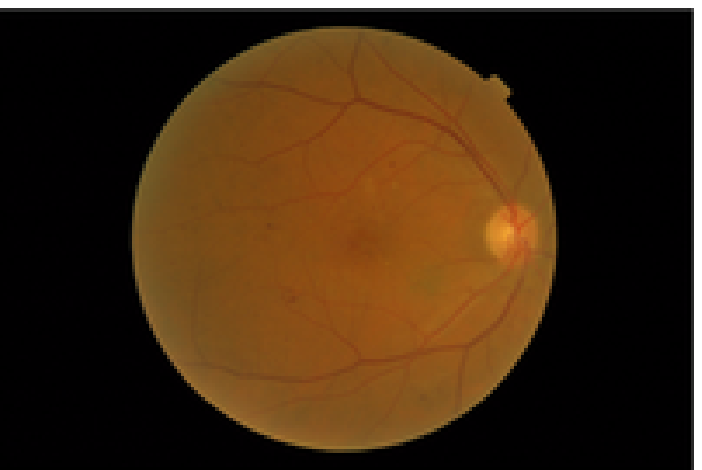}\\
\includegraphics[width=6cm]{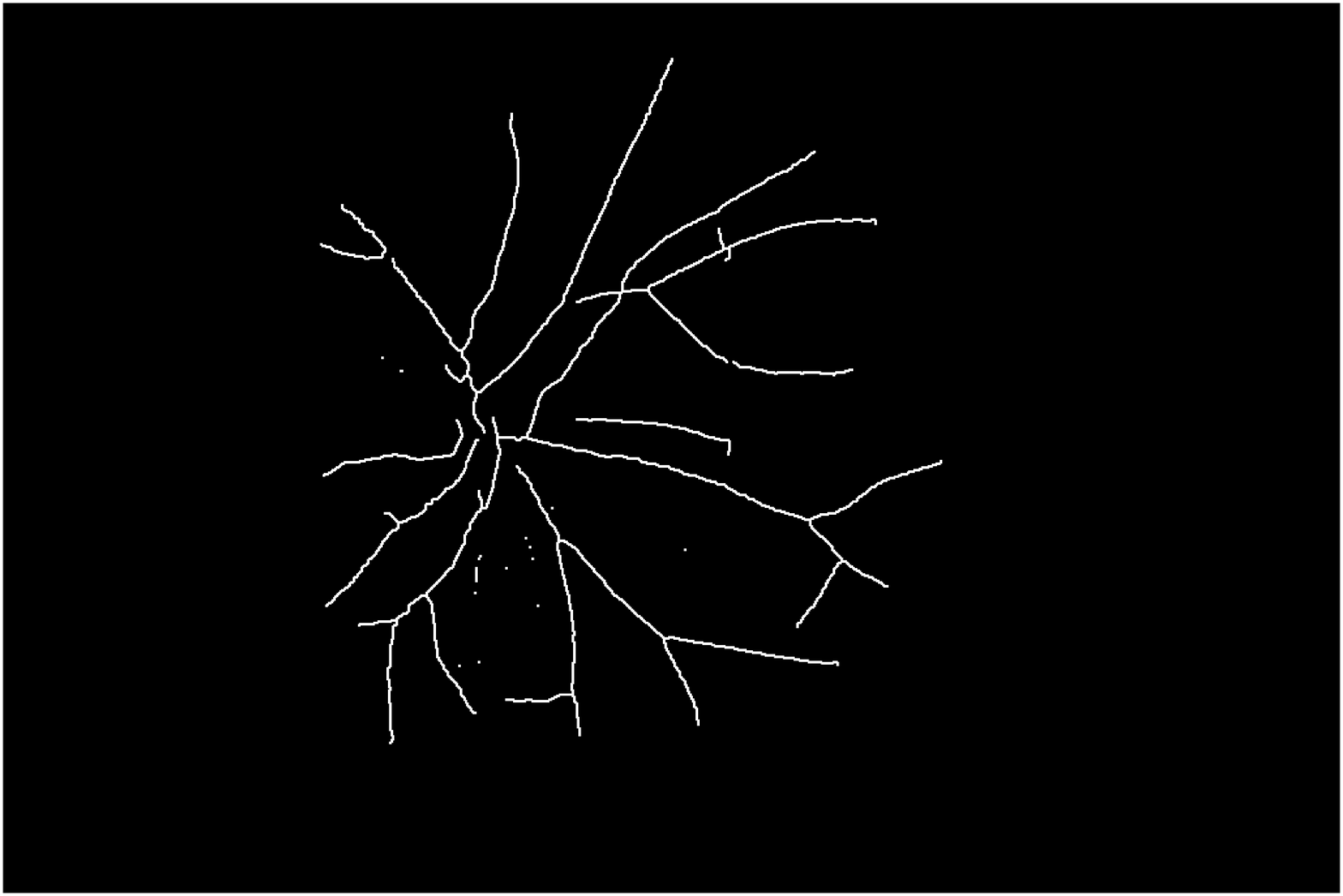}
\includegraphics[width=6cm]{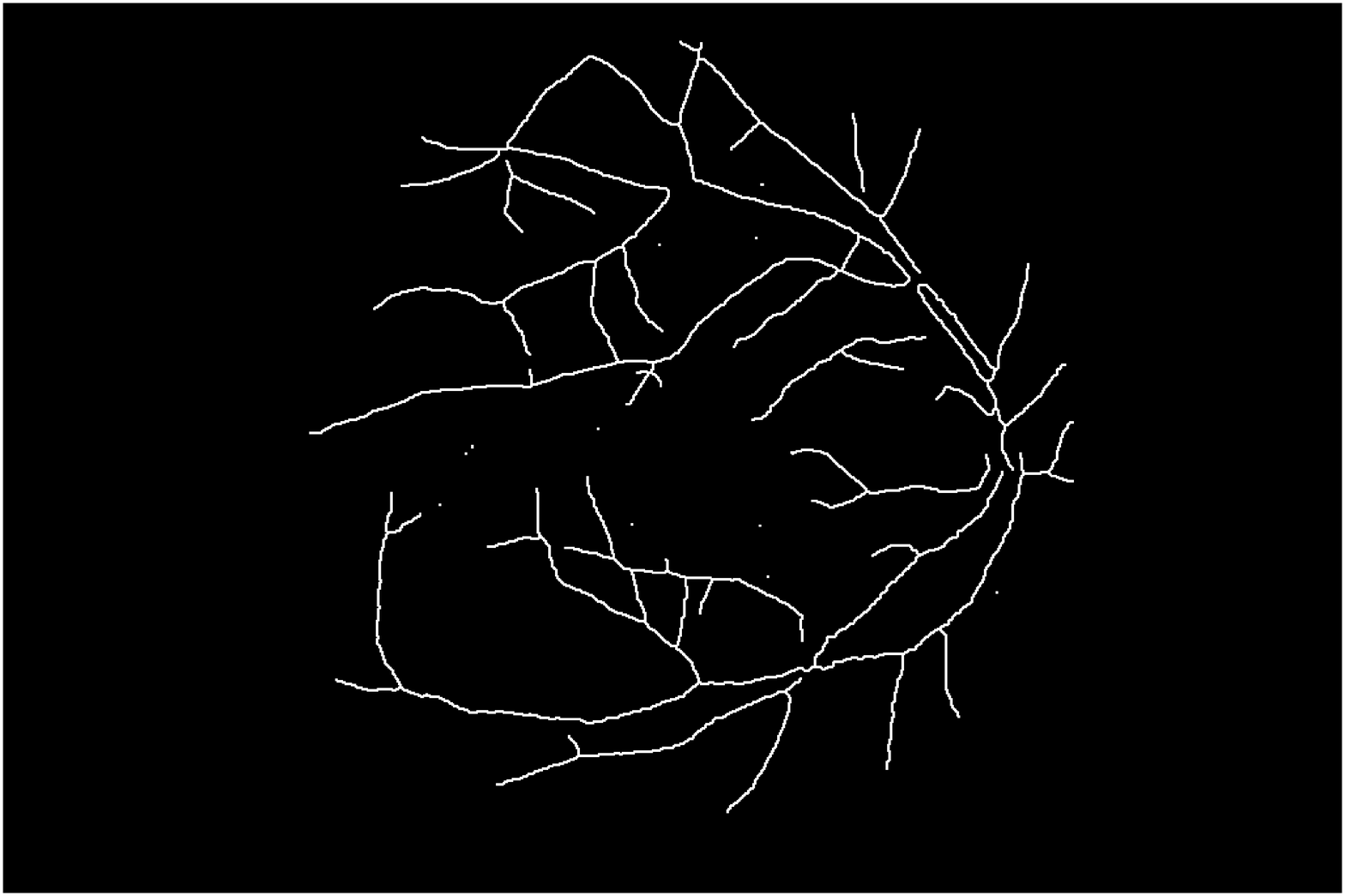}
\caption{Eye photos (top) and vessel contour extraction (bottom).}
\label{fig:im_photo_contour}
\end{figure*}

The first example is dedicated to the problem of image alignment. We consider two photos of vessels in human eyes. The original photos and the images of extracted vessel contours (obtained from the method of \cite{thomas}) are presented in Figure \ref{fig:im_photo_contour}. To align the vessel images, the shape context algorithm uses the context radial histograms of contour points (see \cite{belongie_shape_matching_shape_context}). In other words, according to the shape context algorithm one aligns points which have similar context histograms. The PATH algorithm uses also information about the graph structure. When we use the PATH algorithm we have to tune the parameter $\alpha$ (\ref{eq:F_lambda_alpha}), we tested several possible values and we took the one which produced the best result. To construct graph we use all points of vessel contours as   graph nodes and we connect all nodes within a circle of radius $r$ (in our case we use $r=50$). Finally, to each edge $(i,j)$ we associate the weight $w_{i,j}=\exp(-|x_i-y_j|)$. 

\begin{figure*}[htbp]
\centering
\includegraphics[width=6cm]{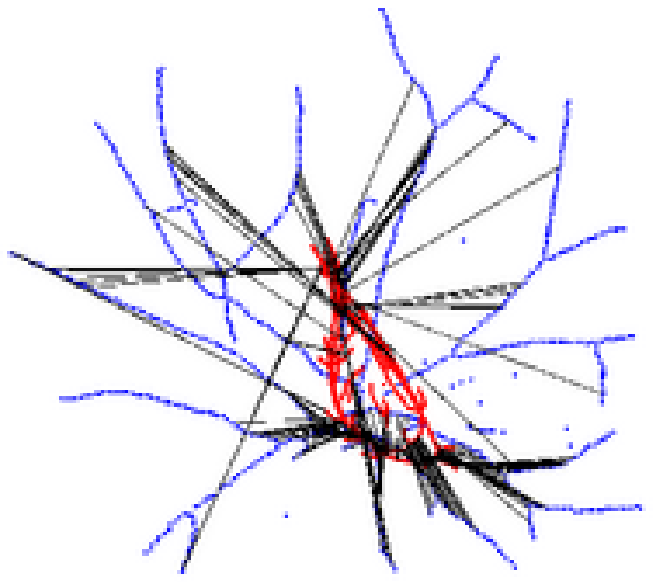}
\includegraphics[width=6cm]{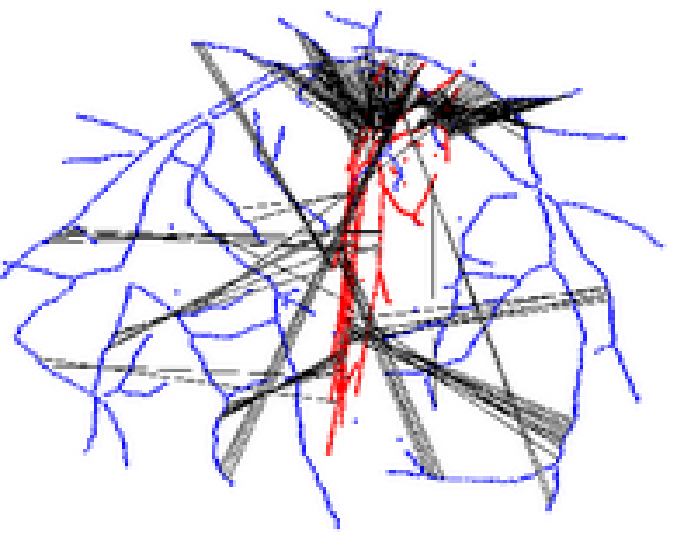}\\
\includegraphics[width=6cm]{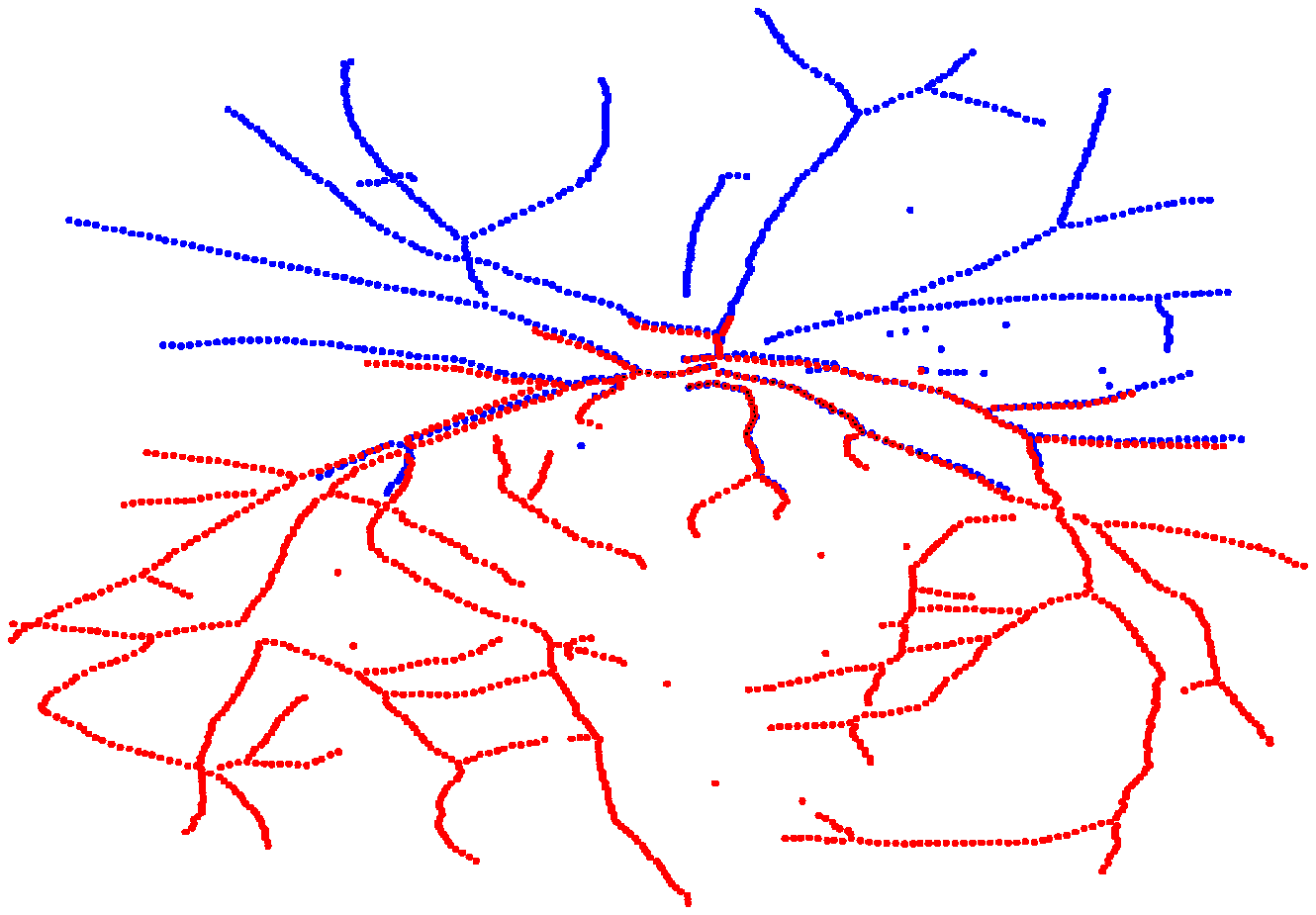}
\includegraphics[width=6cm]{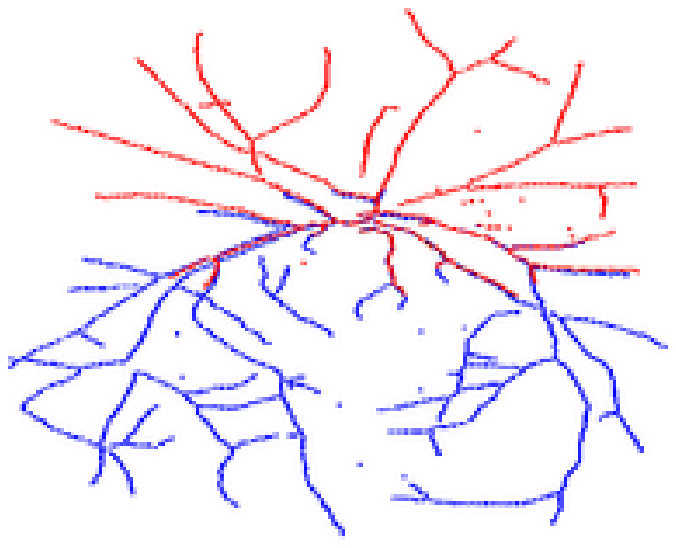}
\caption{Comparison of alignment based on shape context (top) and alignment based on the PATH optimization algorithm (bottom).  For each algorithm we present two alignments: image '1' on image '2' and the inverse. Each alignment is a spline-based transformation (see text).}
\label{fig:im_results}
\end{figure*}

A graph matching algorithm produces an alignment of image contours, then to align two images we have to expand this alignment to the rest of image. For this purpose, we use a smooth spline-based transformation \cite{bookstein}. In other words, we estimate parameters of the spline transformation from the known alignment of contour points and then we apply this transformation to the whole image. Results of image matching based on shape context algorithm and on PATH algorithm are presented in Figure \ref{fig:im_results}, where black lines designate connections between associated points.   We observe that the context shape method creates many unwanted matching, while PATH produces a matching that visually corresponds to a correct alignment of the structure of vessels.
The main reason why graph matching works better than shape context matching is the fact that shape context does not take into account the relational positions of matched points and may lead to totally incoherent graph structures. In contrast, graph matching tries to match pairs of nearest points in one image to pairs of nearest points in another one. 

Among graph matching methods, different results are obtained with different optimization algorithms. Table \ref{tab:vessel_error} shows the matching errors produced by different algorithms on this vessel alignment problem. The PATH algorithm has the smallest matching error, with the alignment shown on Figure \ref{fig:im_results}. QCV comes next, with an alignment that is also visually correct. On the other hand, the Umeyama algorithm has a much larger matching error, and visually fails to find a correct alignment, similar to the shape context method.
\begin{table}[htbp]
\caption{Alignment of vessel images, algorithm performance}
\label{tab:vessel_error}
\centering
\begin{tabular}{|l|c|c|c|c|c|}
\hline
Method&Shape context&Umeyama&QCV&PATH\\
\hline
\hline
matching error&870.61&764.83&654.42&625.75\\
\hline
\end{tabular}
\end{table}  

\subsection{Recognition of handwritten chinese characters}
\label{sec:vision_2} 
\begin{center}
\begin{figure*}
\centering
\begin{tabular}{|c|c|c|}
\hline
character 1& character 2&character 3\\
\hline
\hline
\includegraphics[width=2cm]{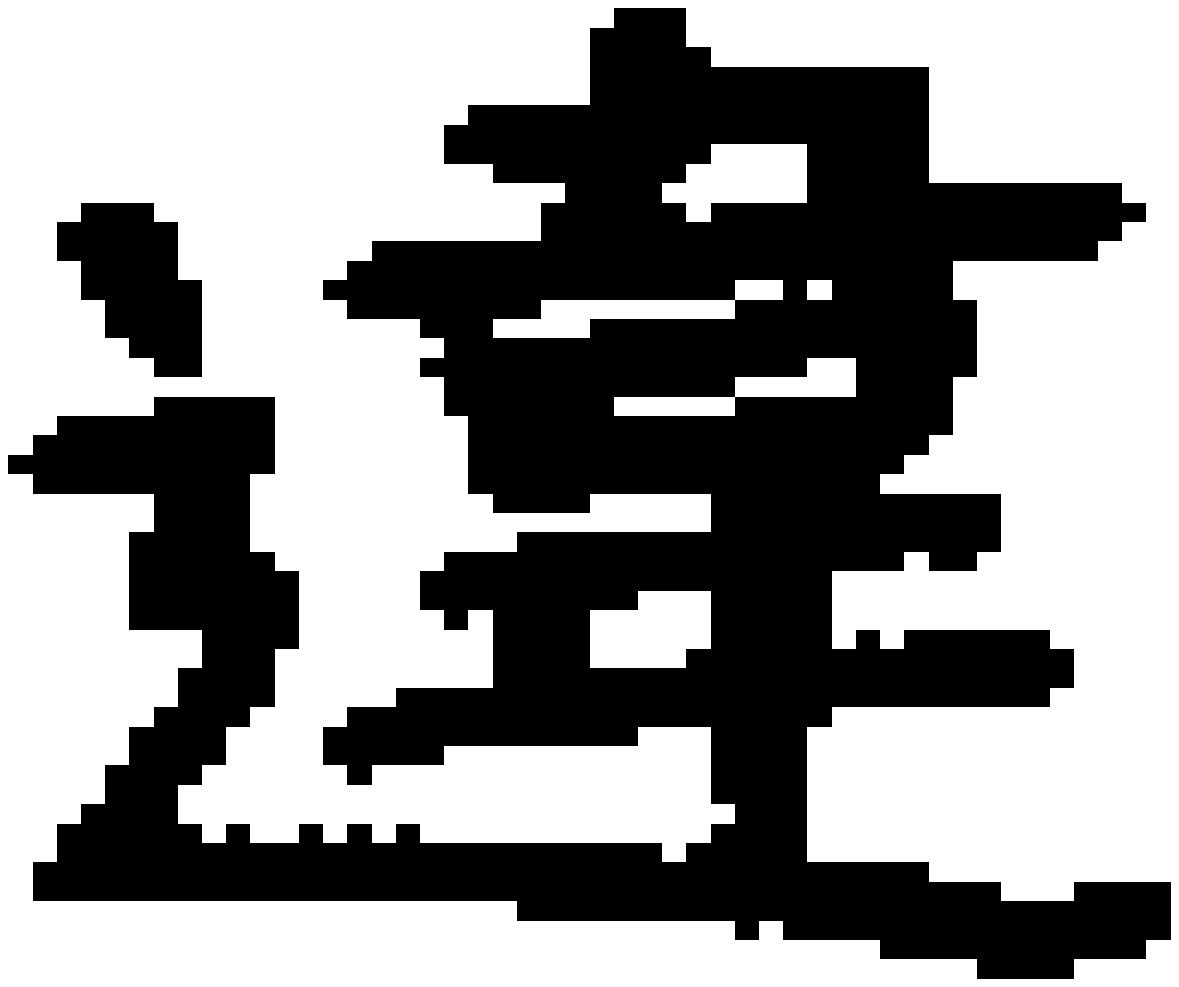}
\includegraphics[width=2cm]{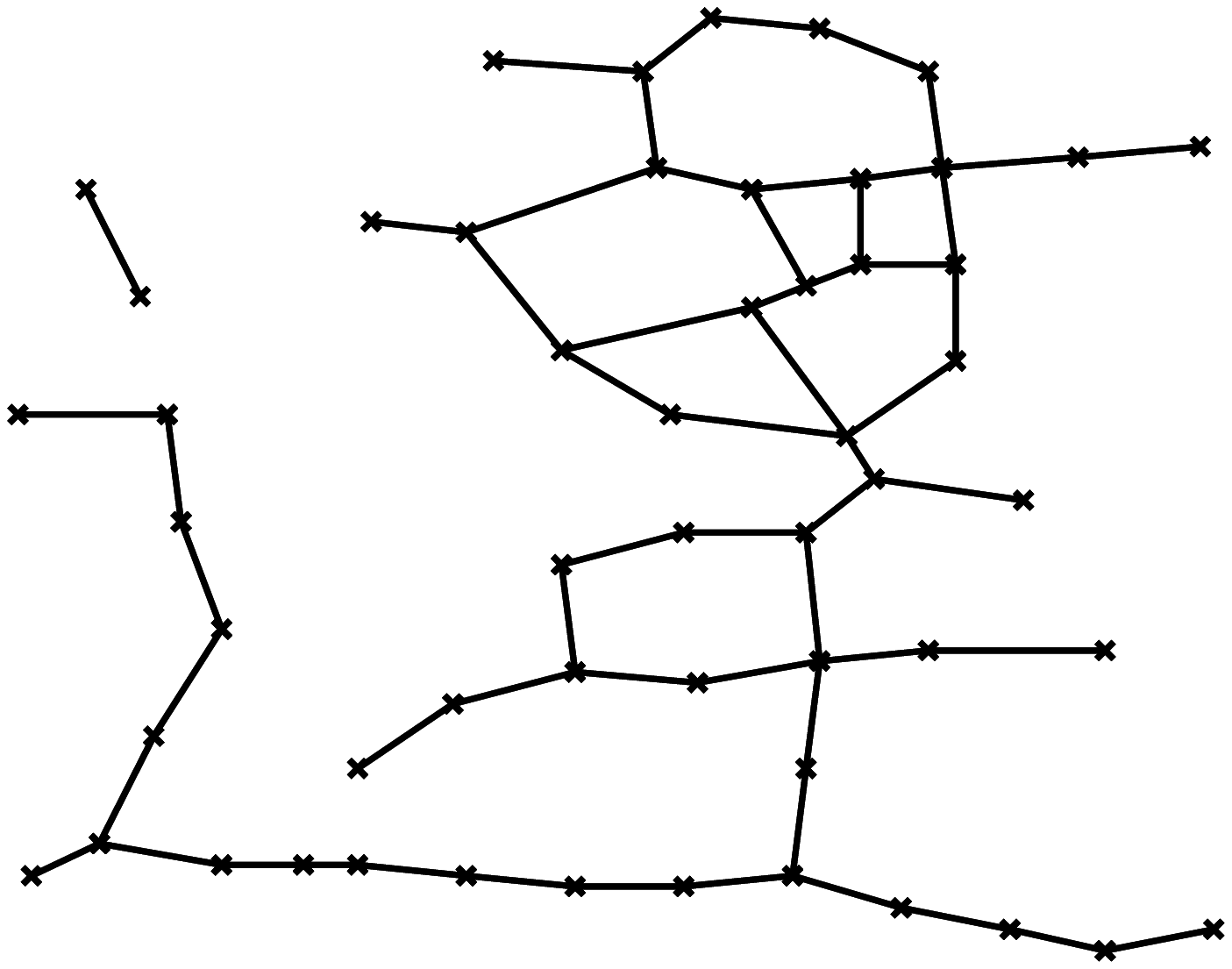}&
\includegraphics[width=2cm]{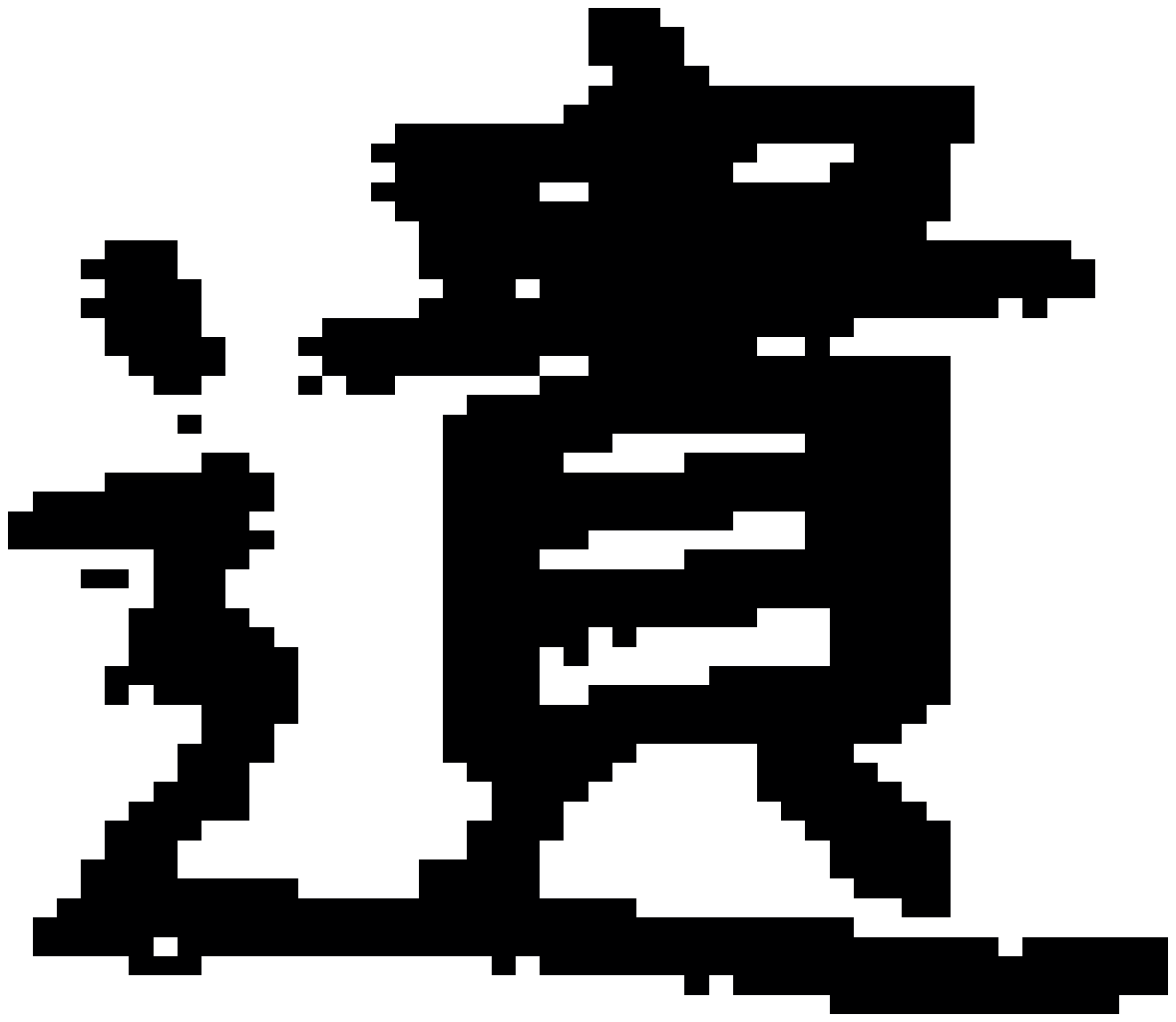}
\includegraphics[width=2cm]{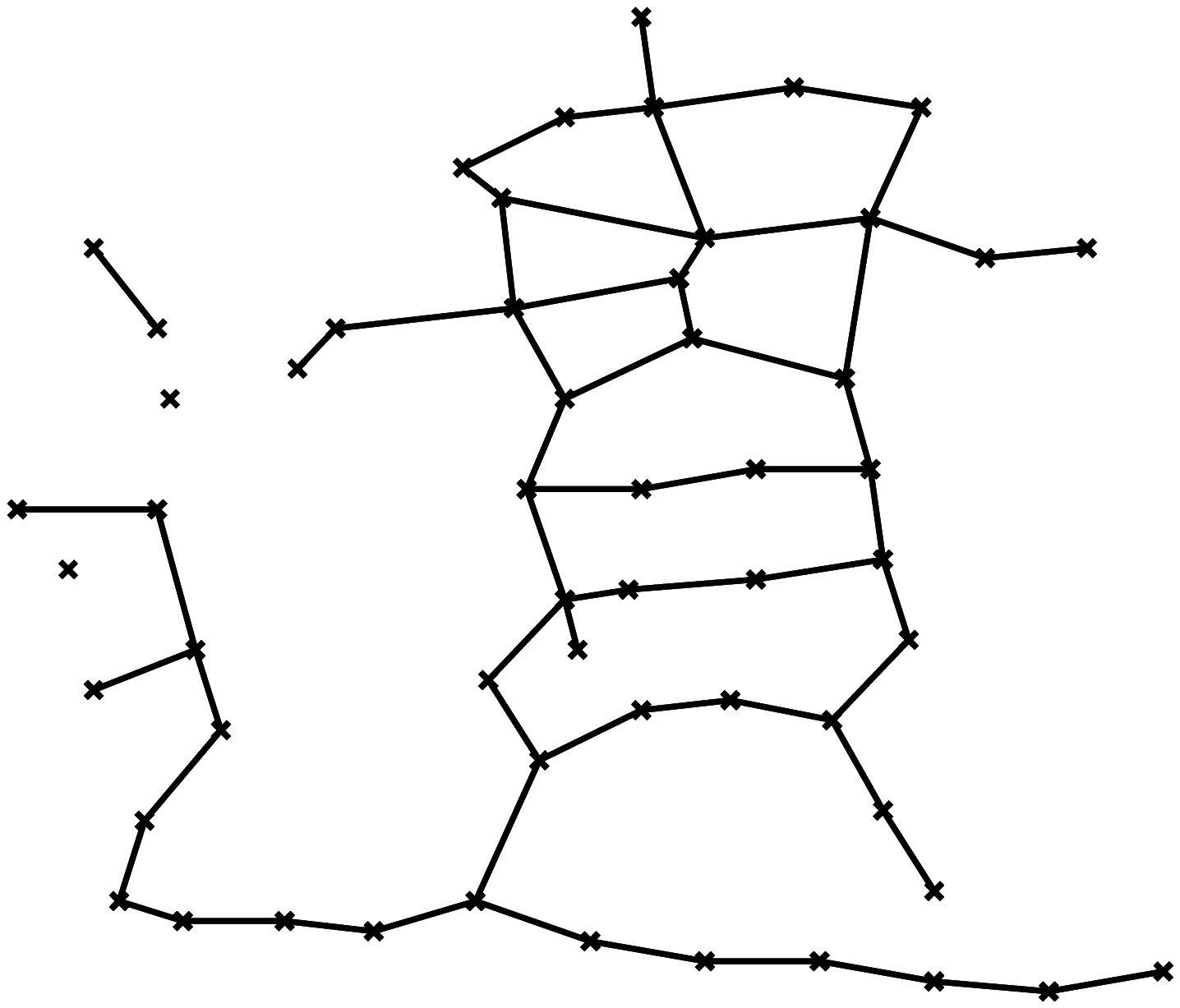}&
\includegraphics[width=2cm]{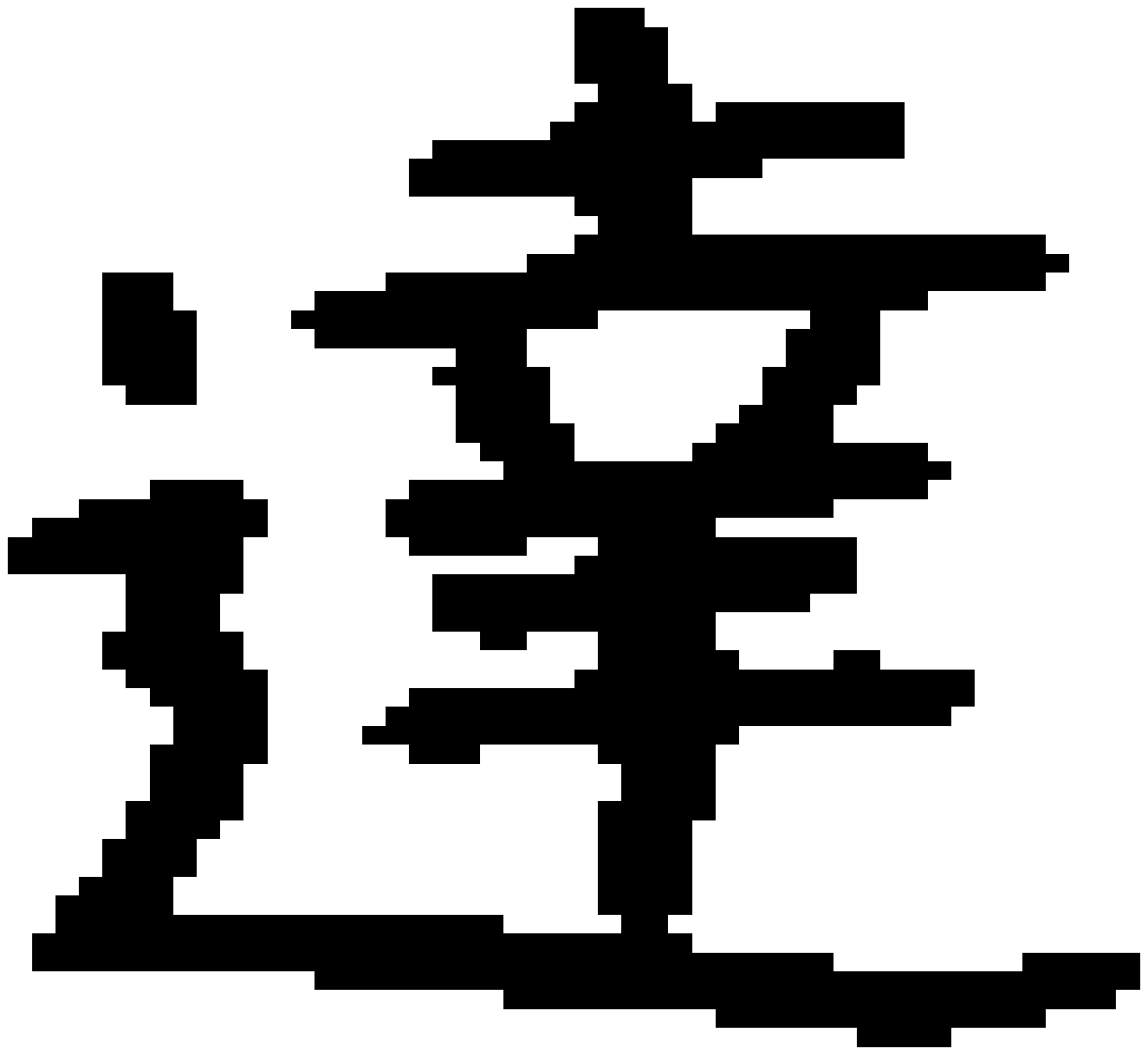}
\includegraphics[width=2cm]{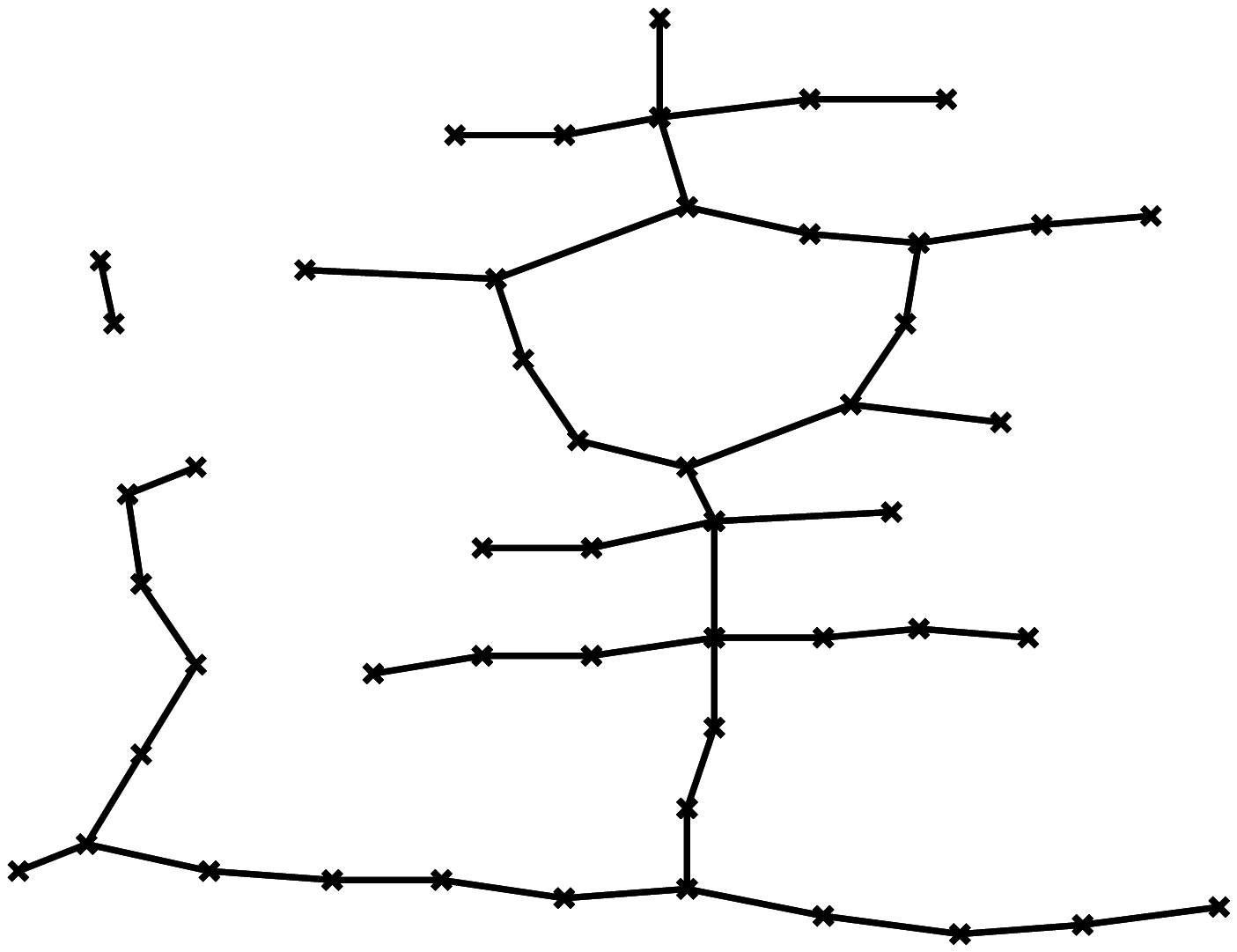}\\
\includegraphics[width=2cm]{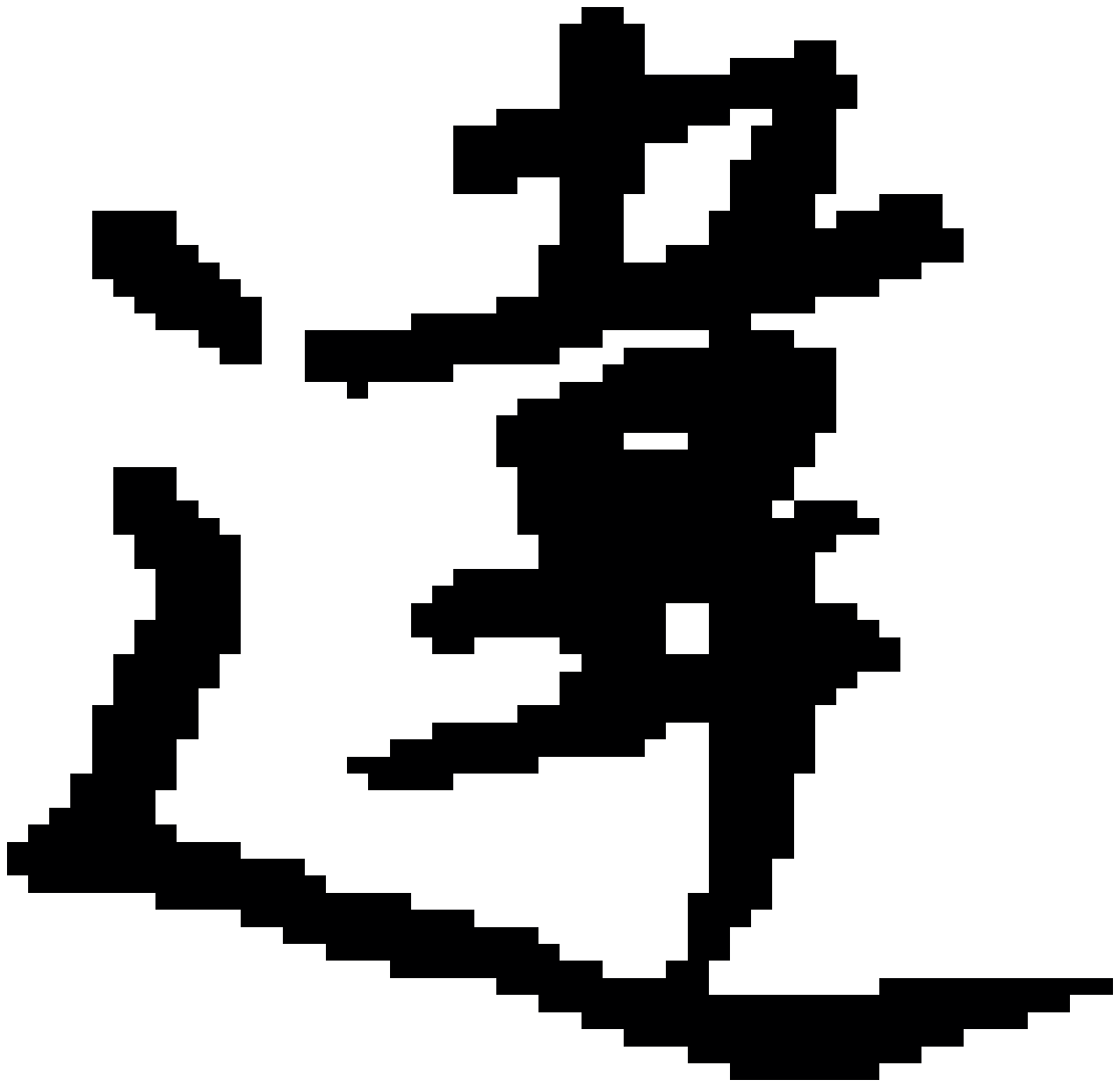}
\includegraphics[width=2cm]{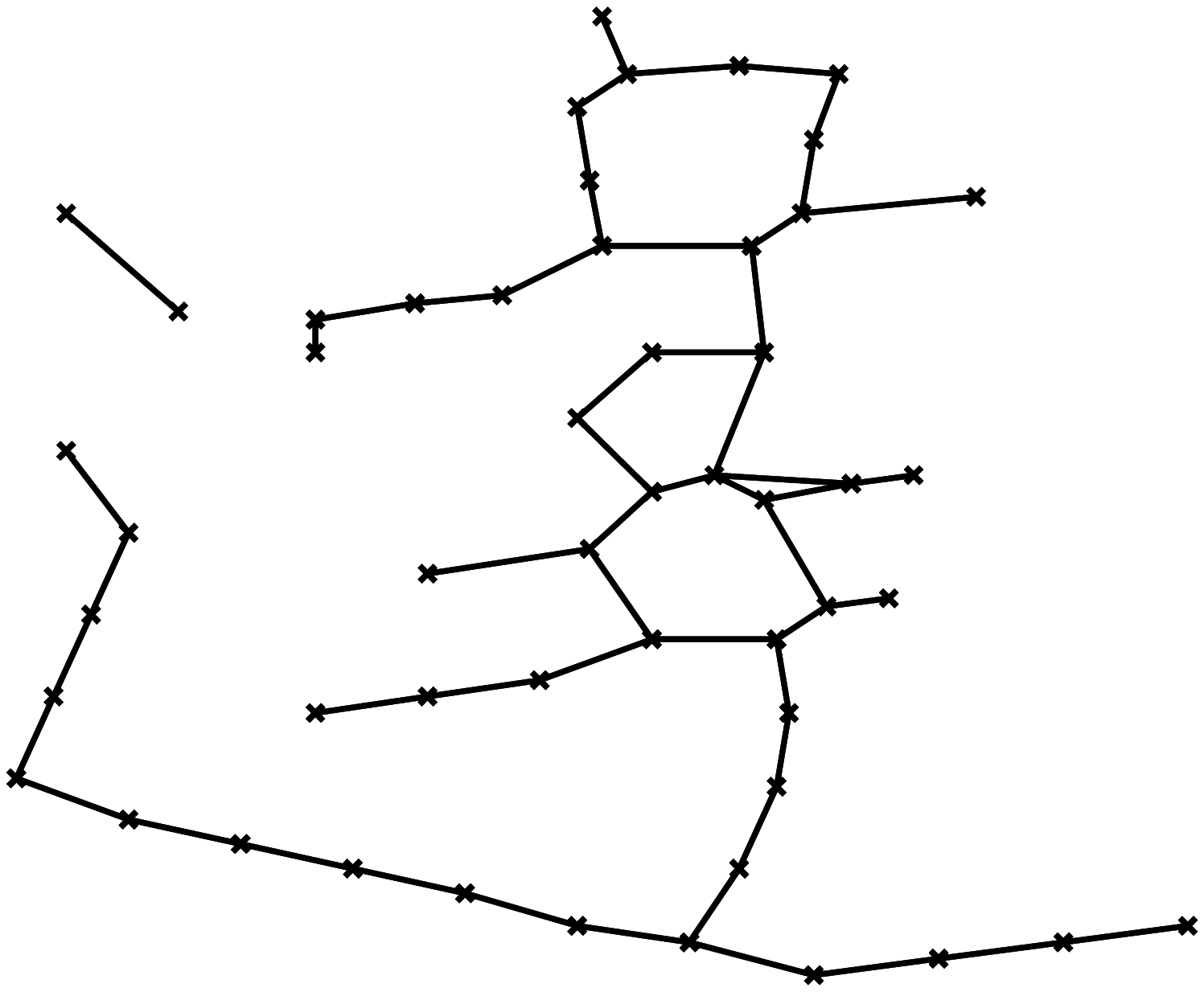}&
\includegraphics[width=2cm]{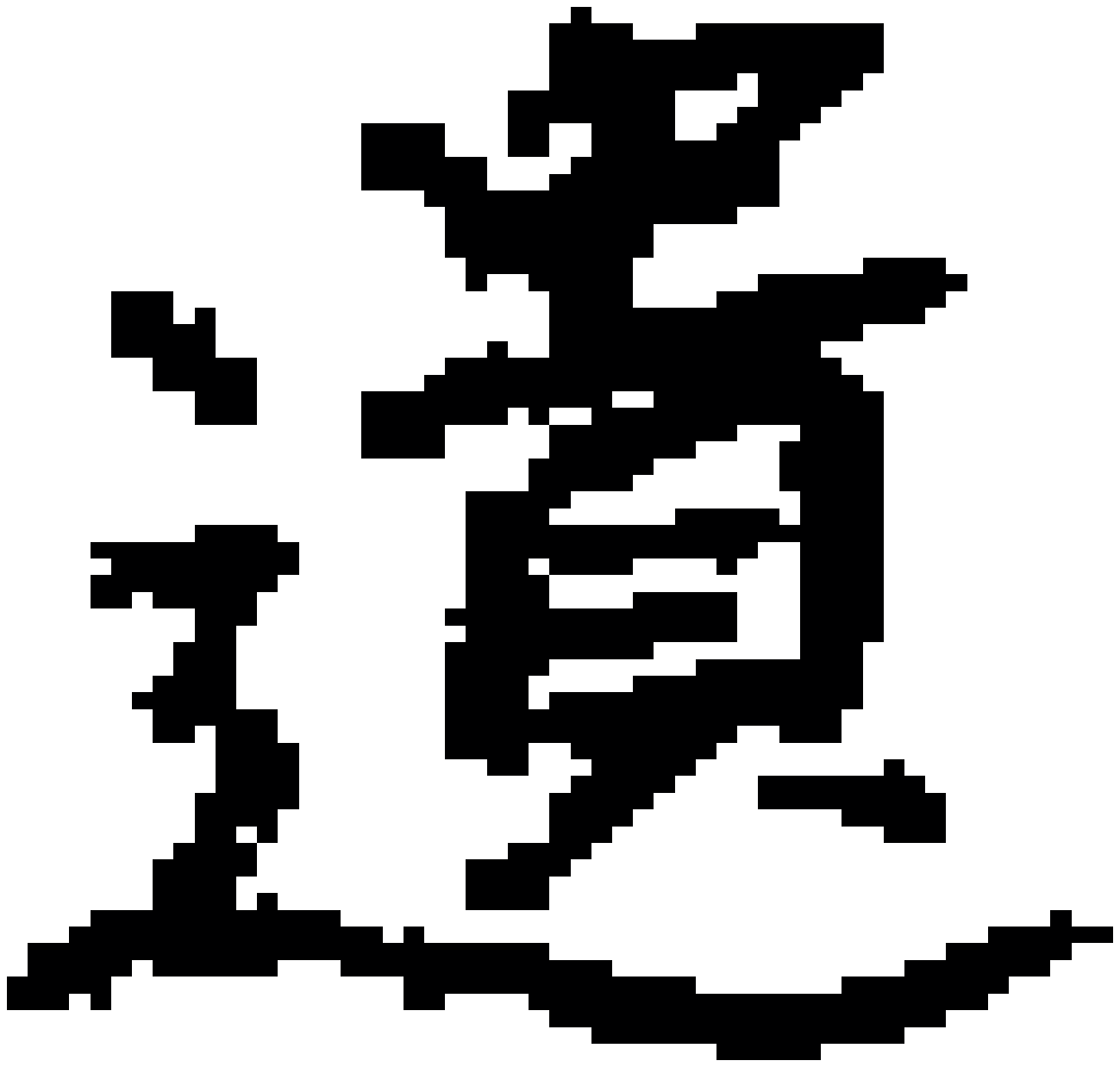}
\includegraphics[width=2cm]{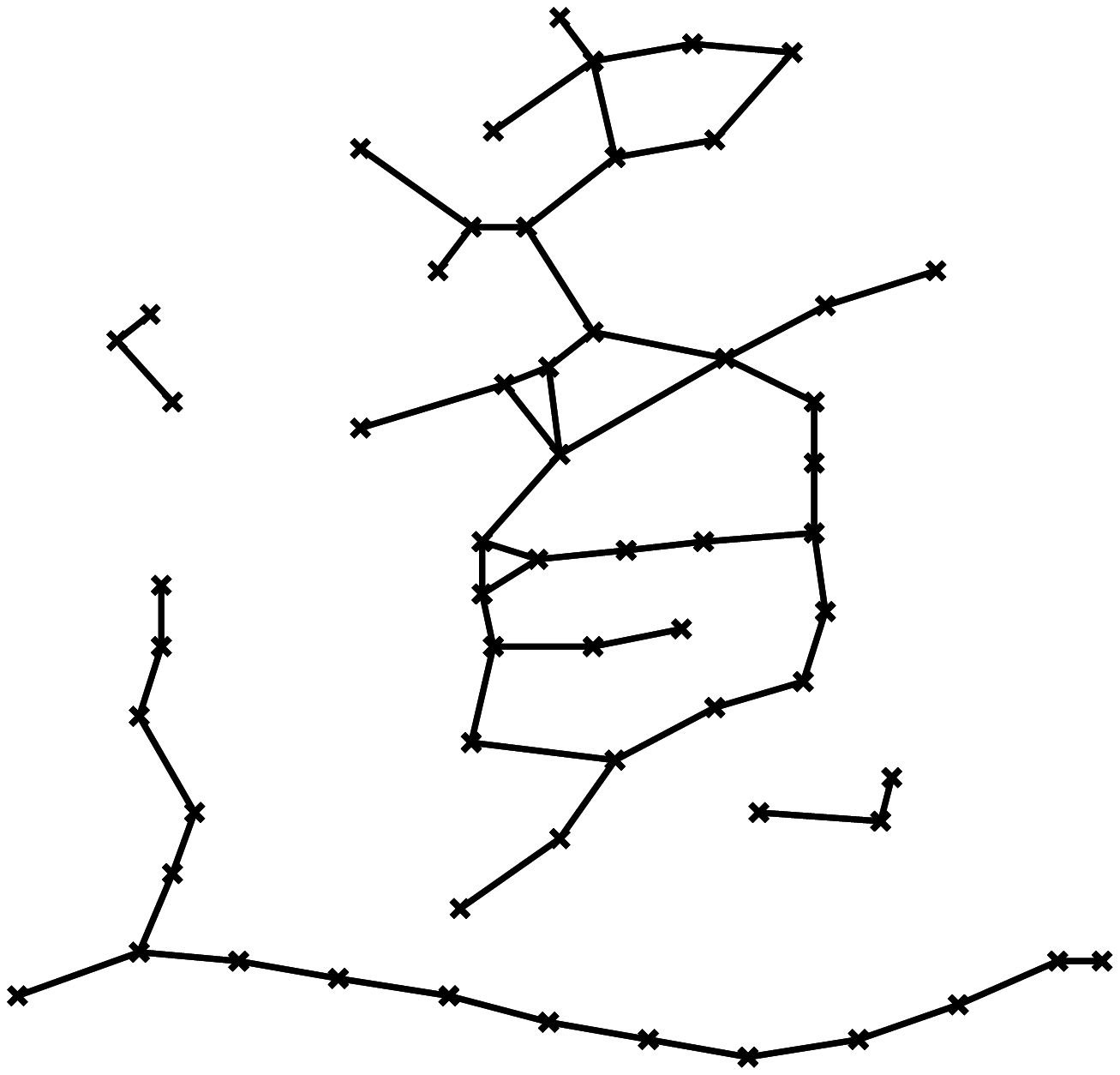}&
\includegraphics[width=2cm]{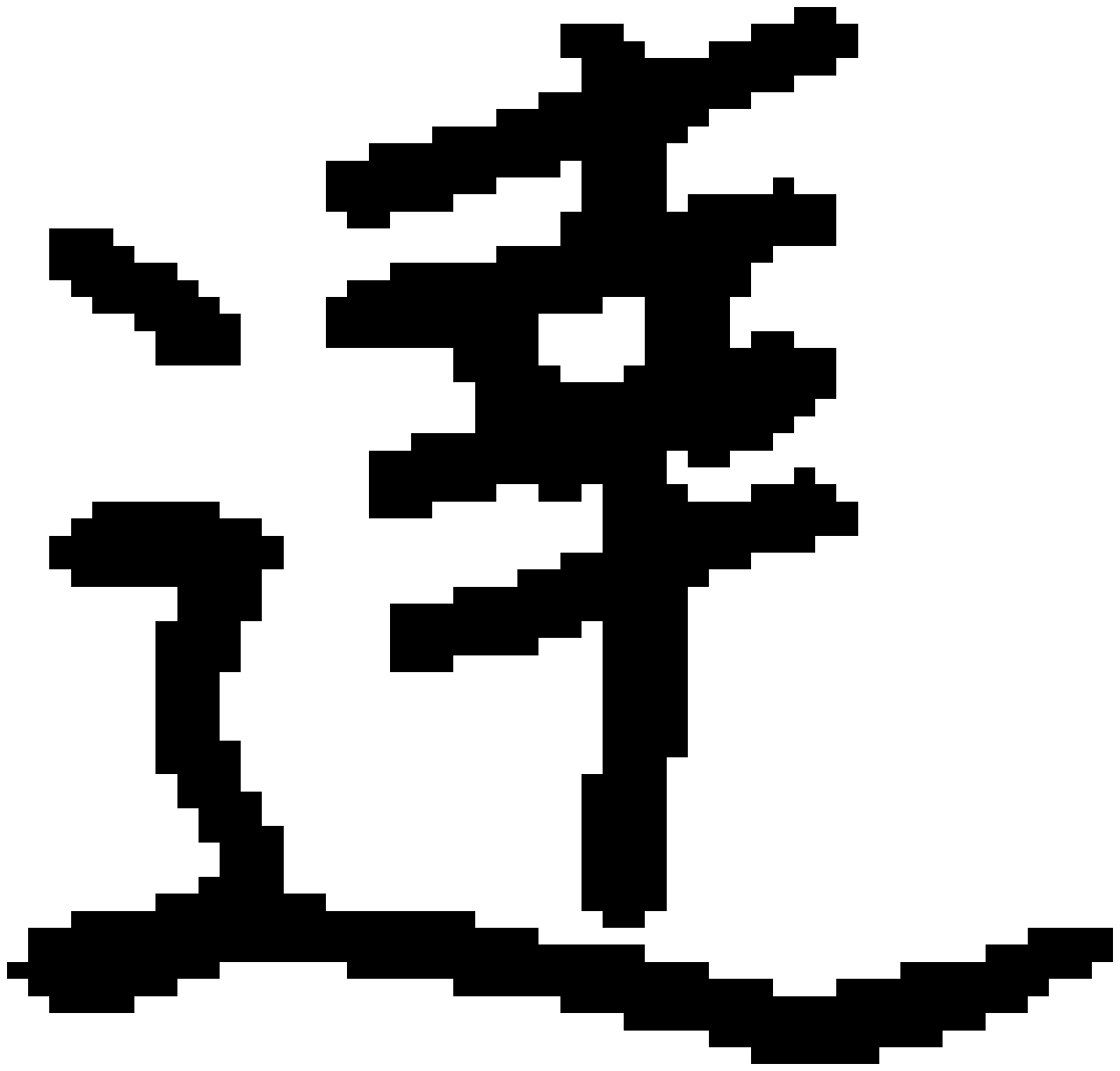}
\includegraphics[width=2cm]{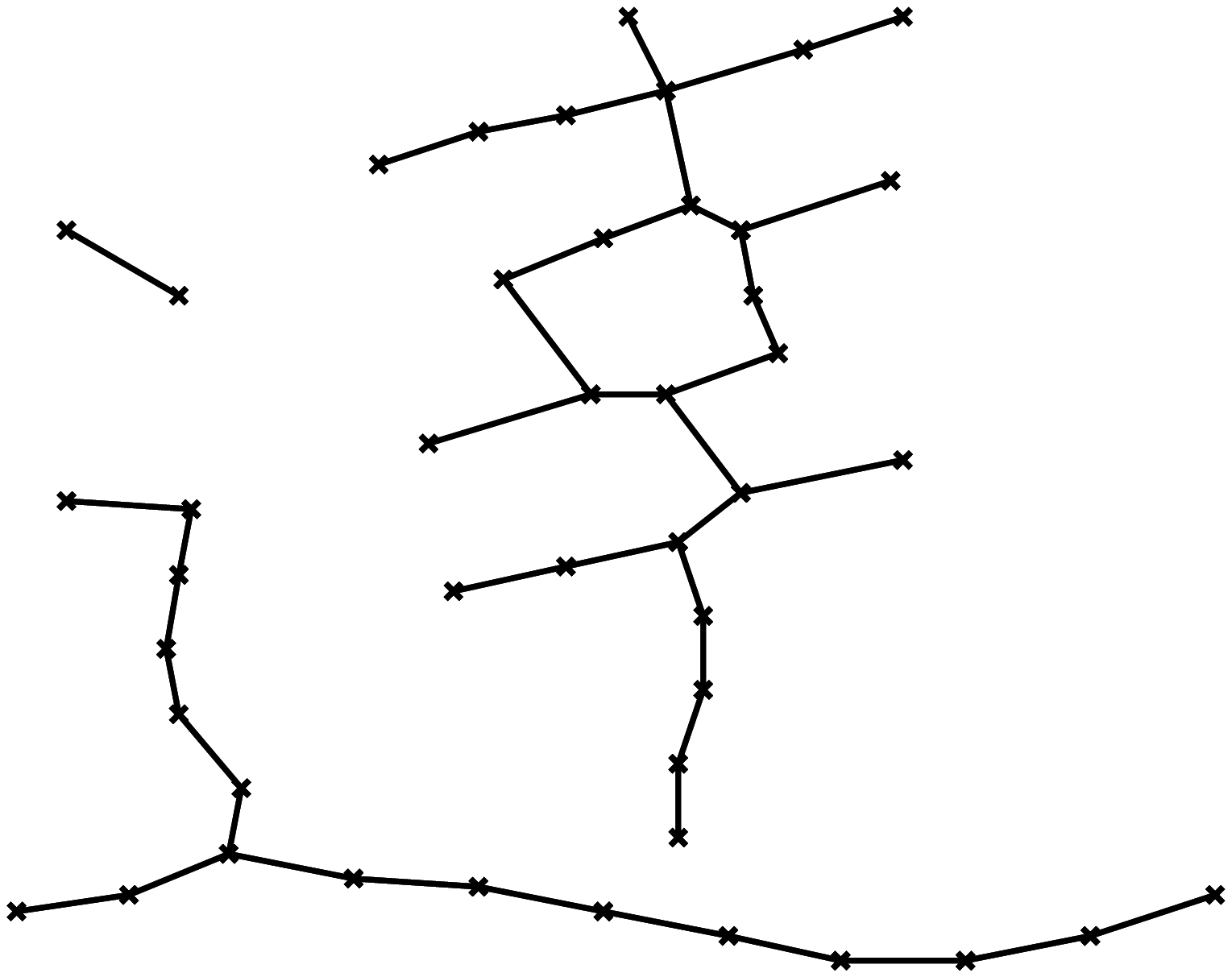}\\
\includegraphics[width=2cm]{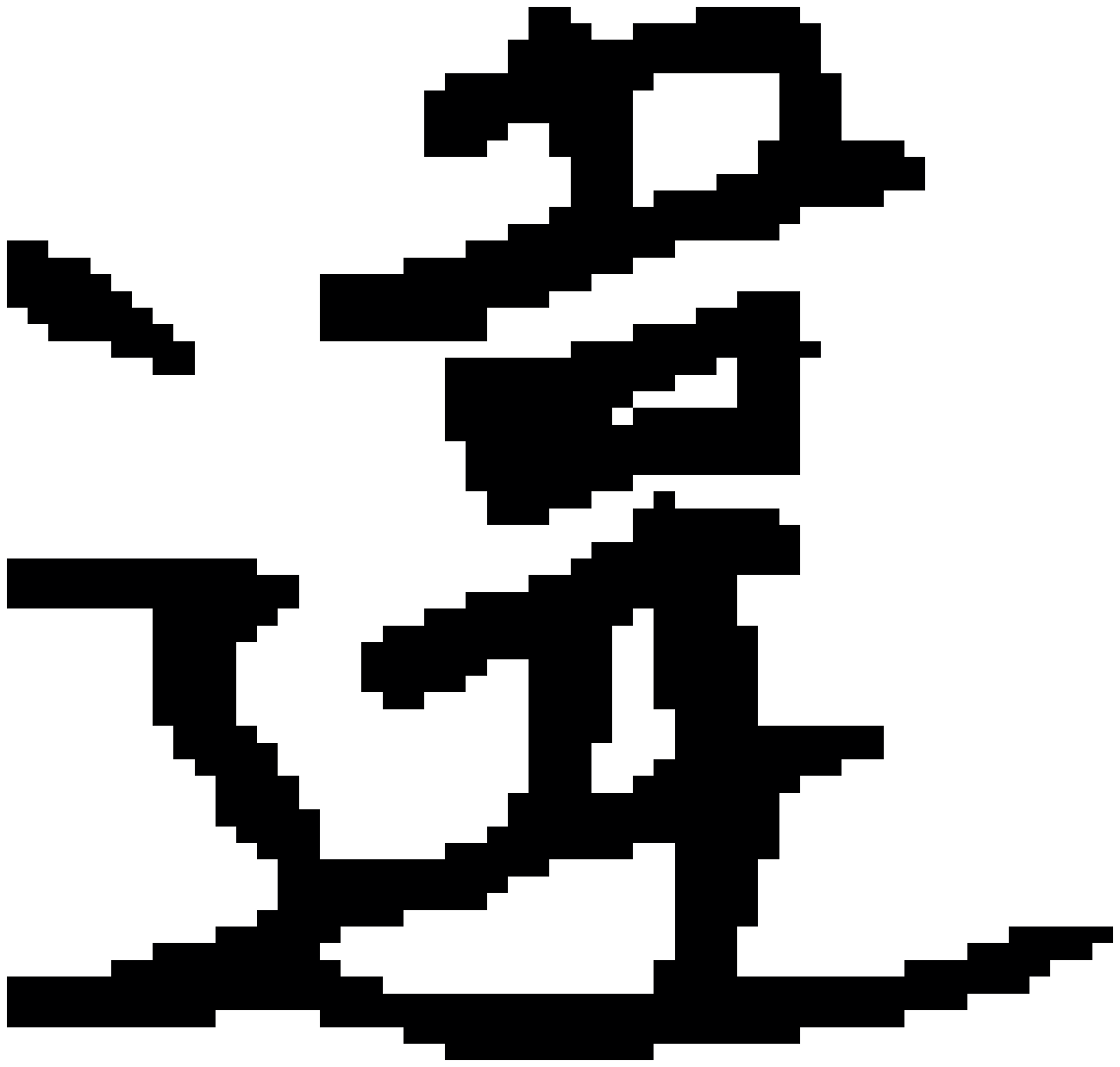}
\includegraphics[width=2cm]{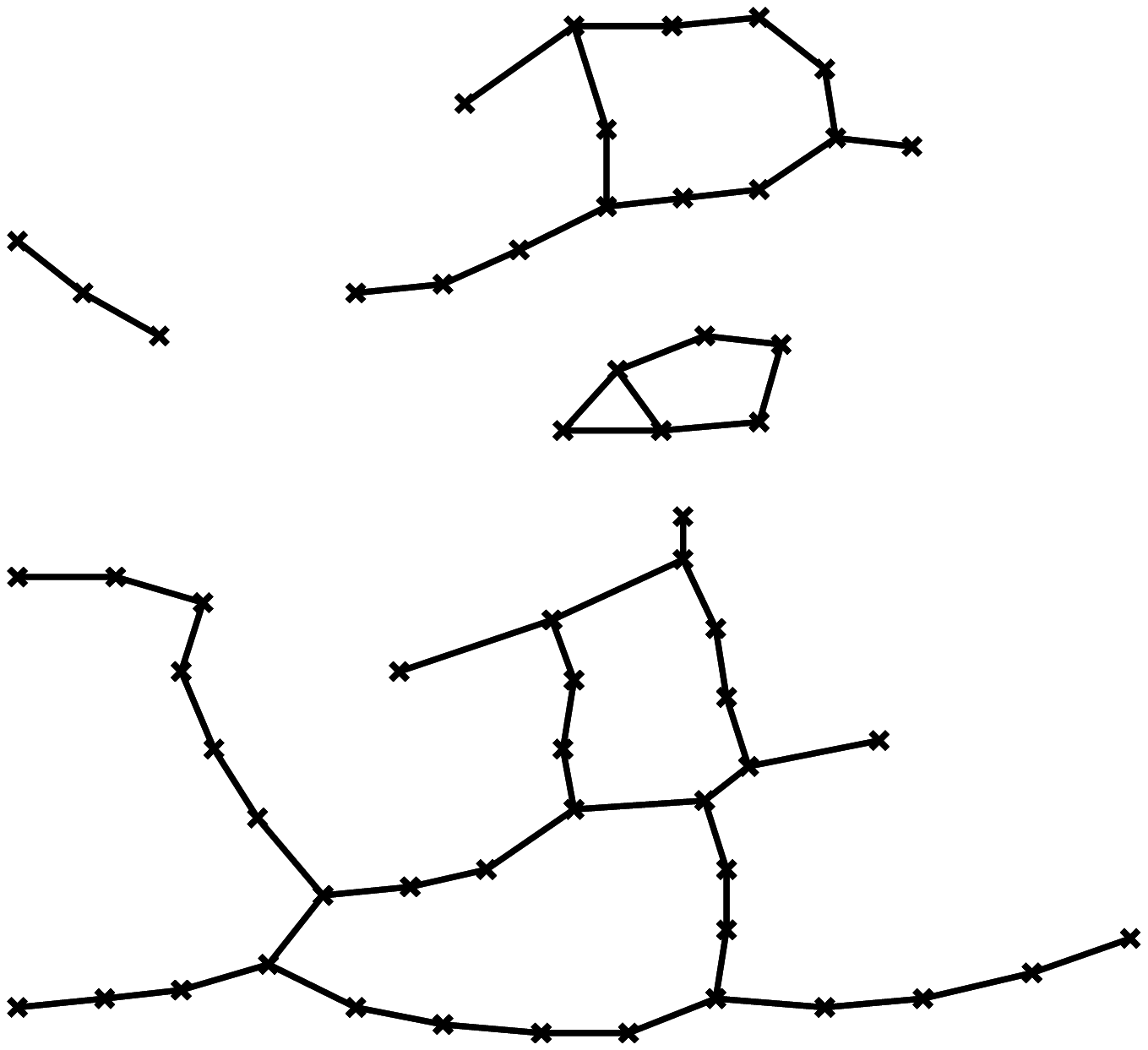}&
\includegraphics[width=2cm]{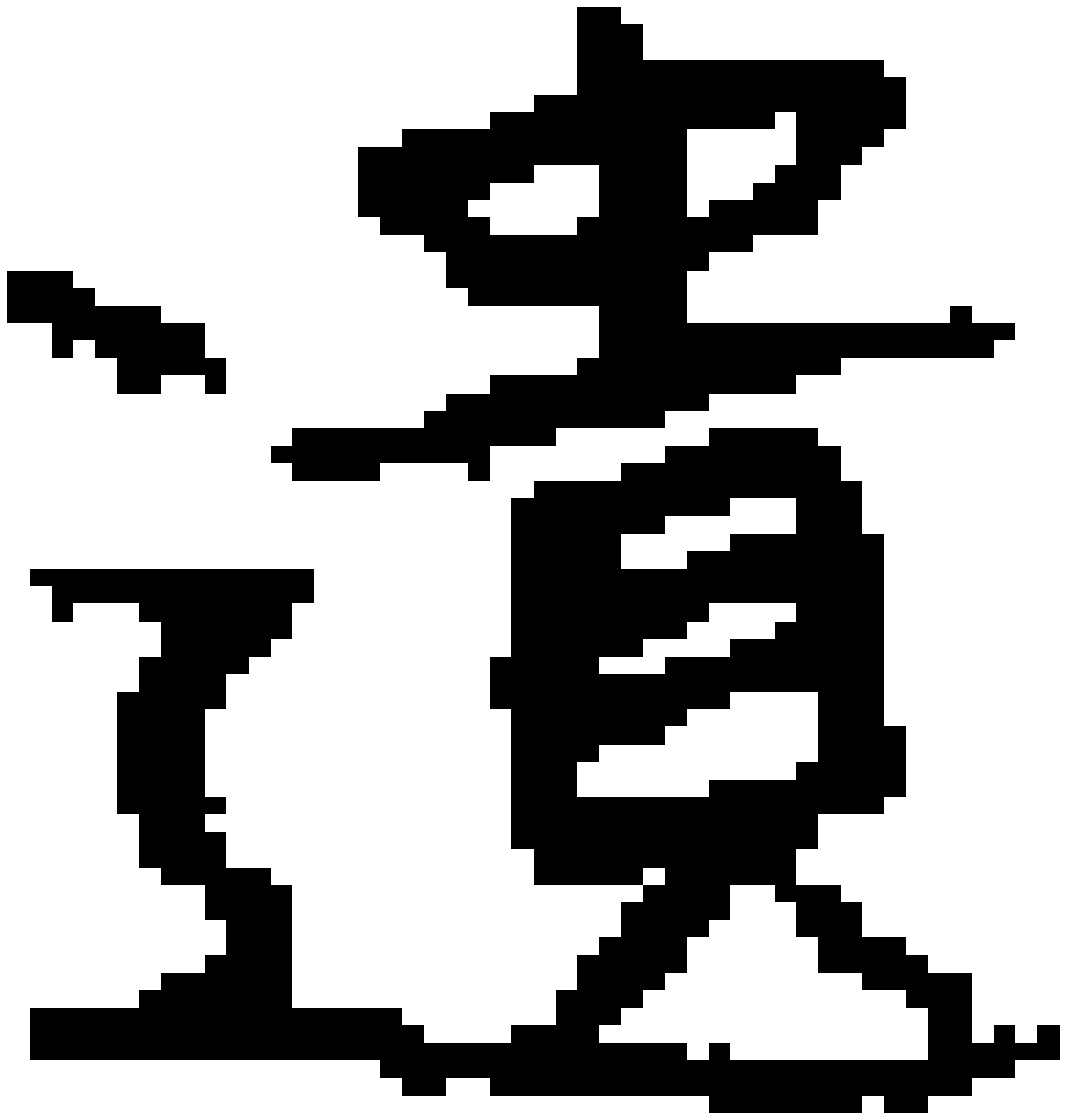}
\includegraphics[width=2cm]{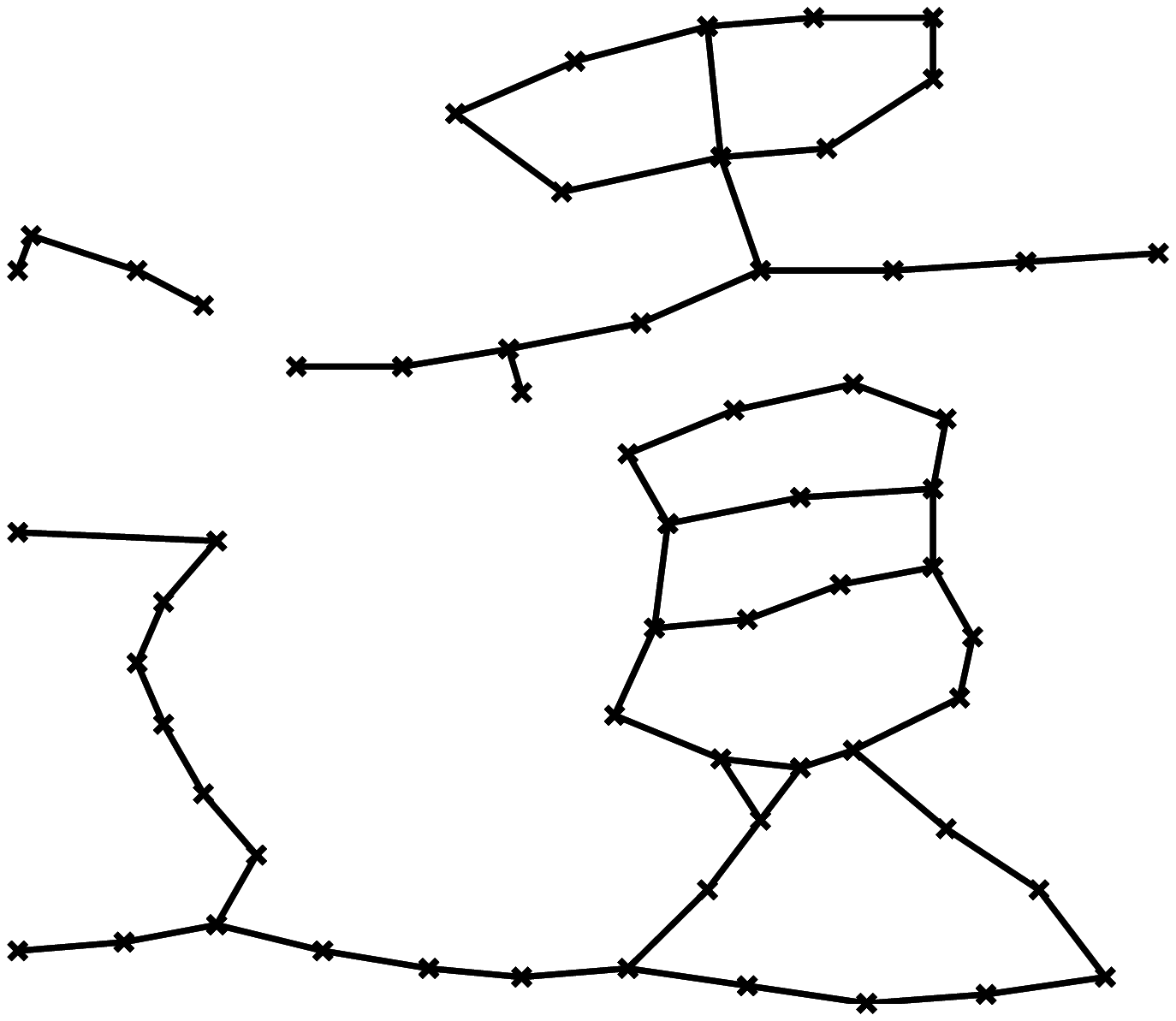}&
\includegraphics[width=2cm]{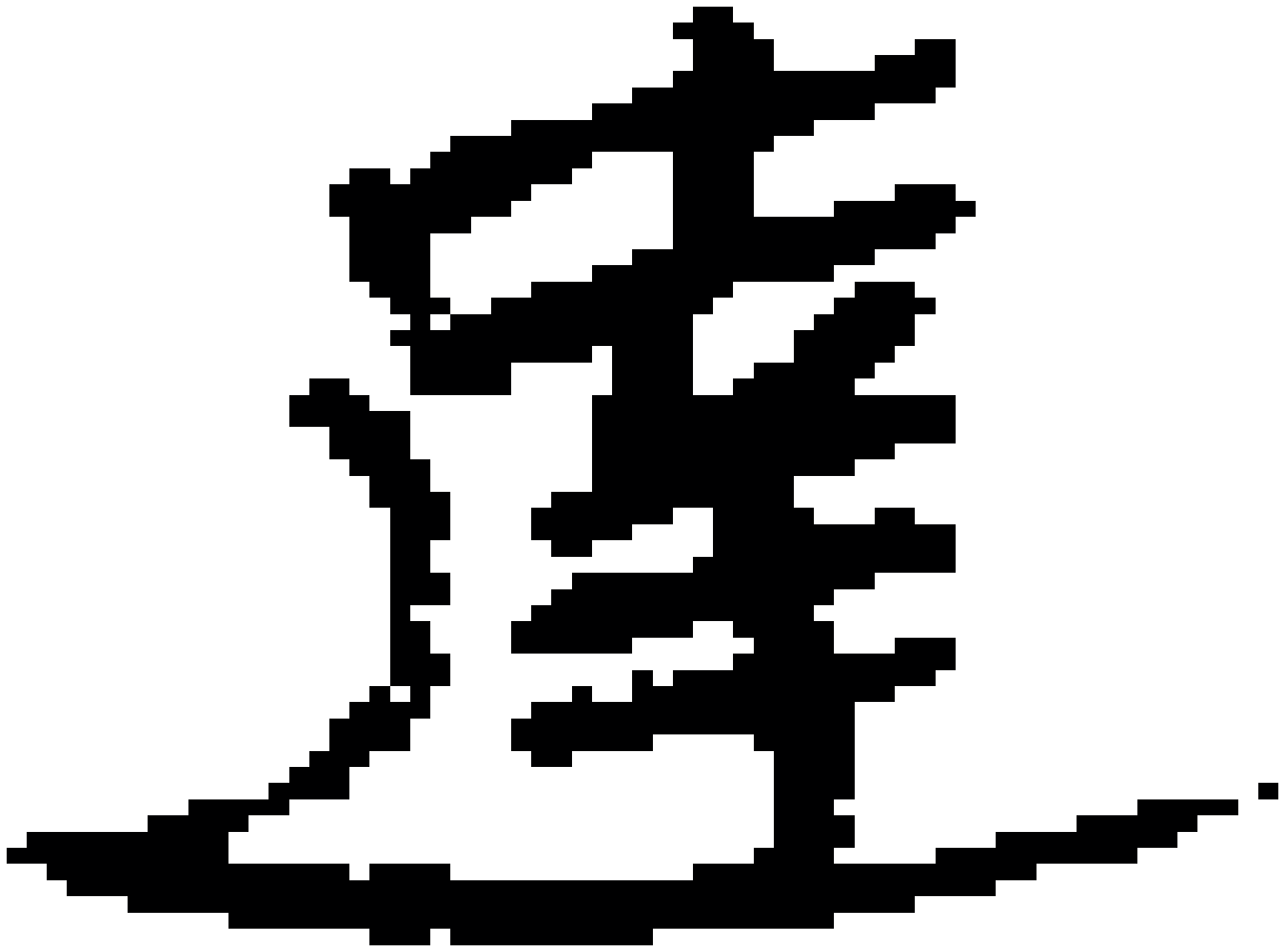}
\includegraphics[width=2cm]{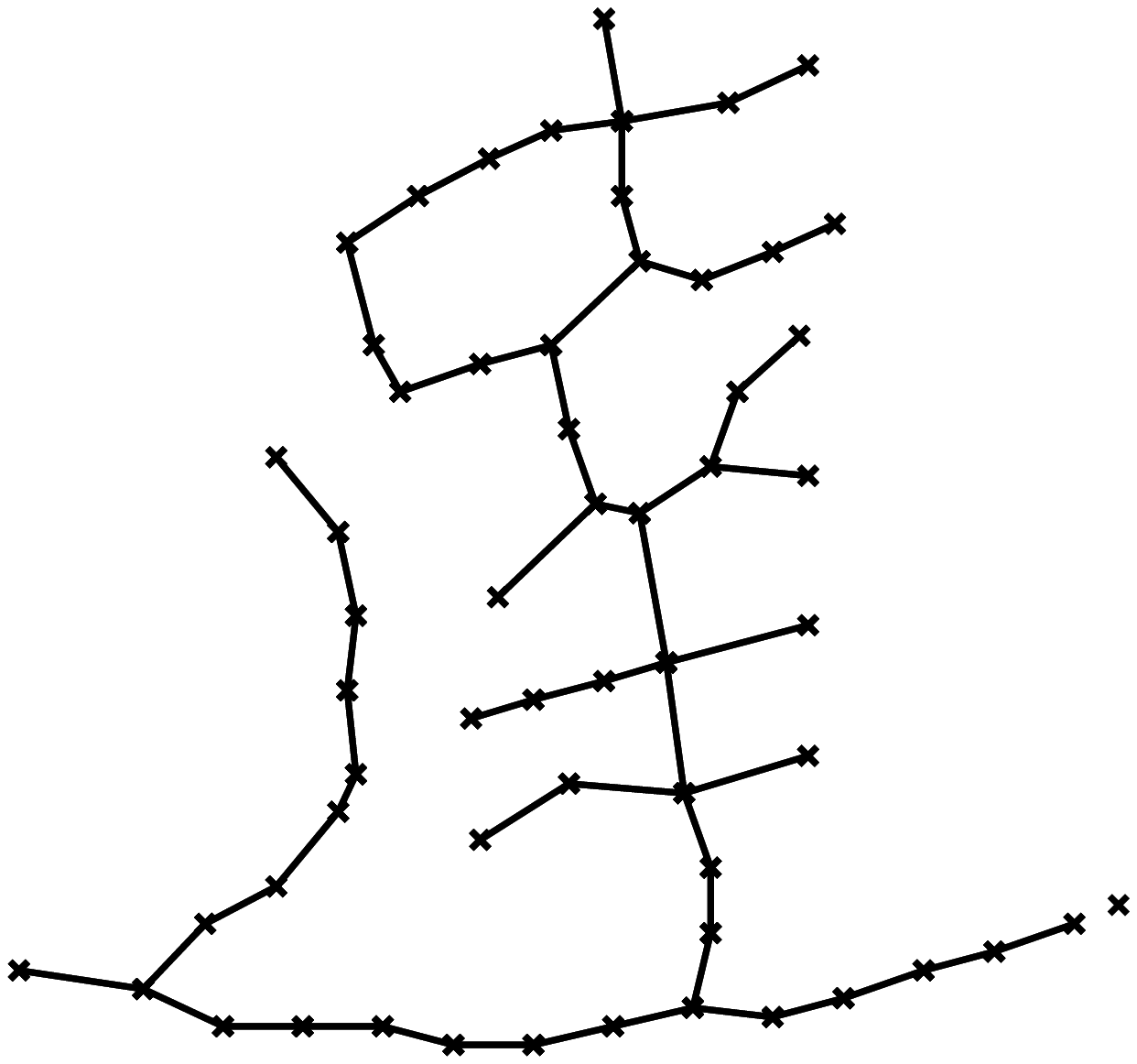}\\
\hline
\end{tabular}\\
\caption{Chinese characters from the ETL9B dataset.}
\label{fig:chinese_characters_examples}
\end{figure*}
\end{center}

\begin{figure*}[htbp]
 \centering
 \subfigure[]{\includegraphics[width=7cm]{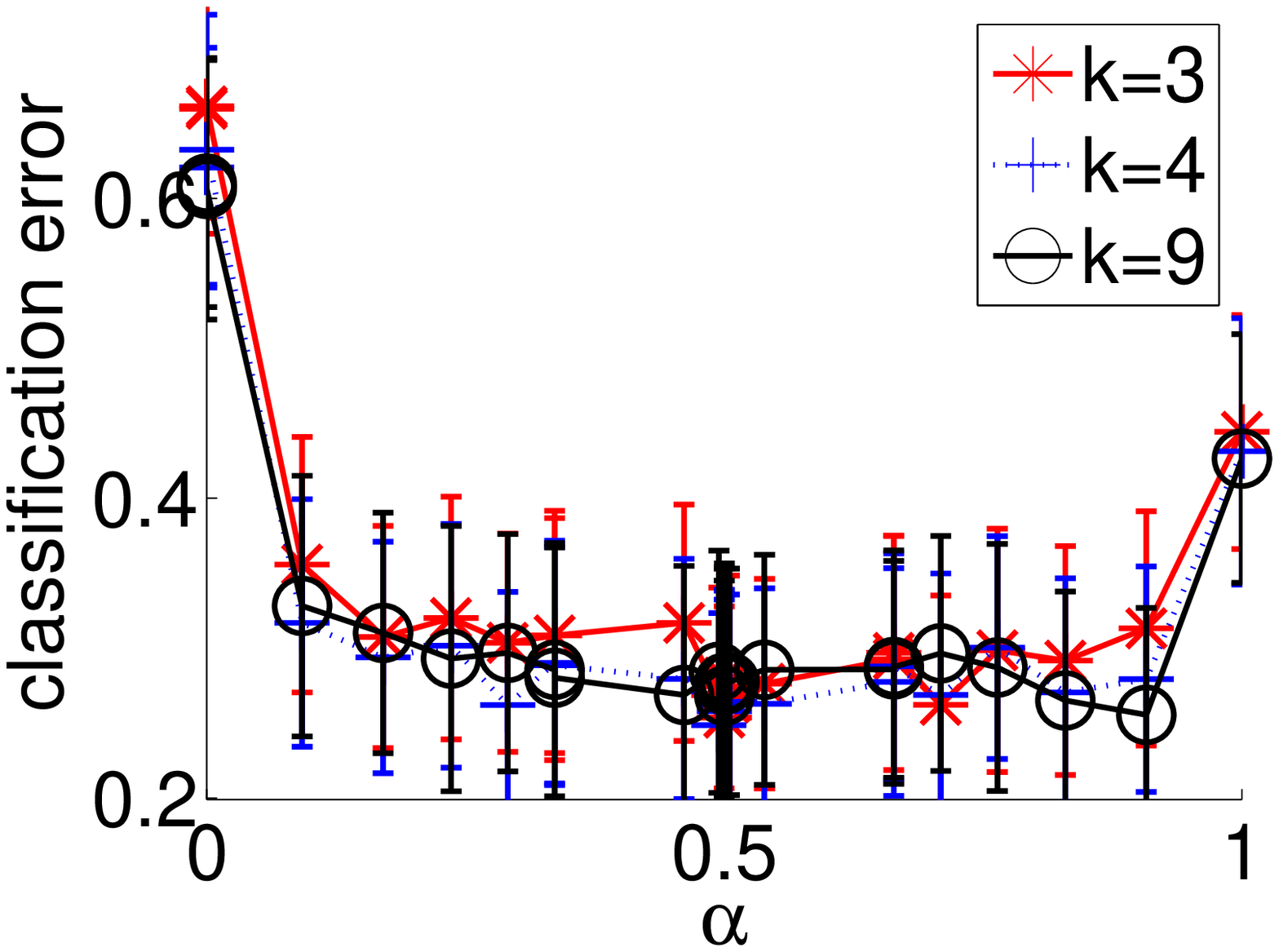}}
 \subfigure[]{\includegraphics[width=7cm]{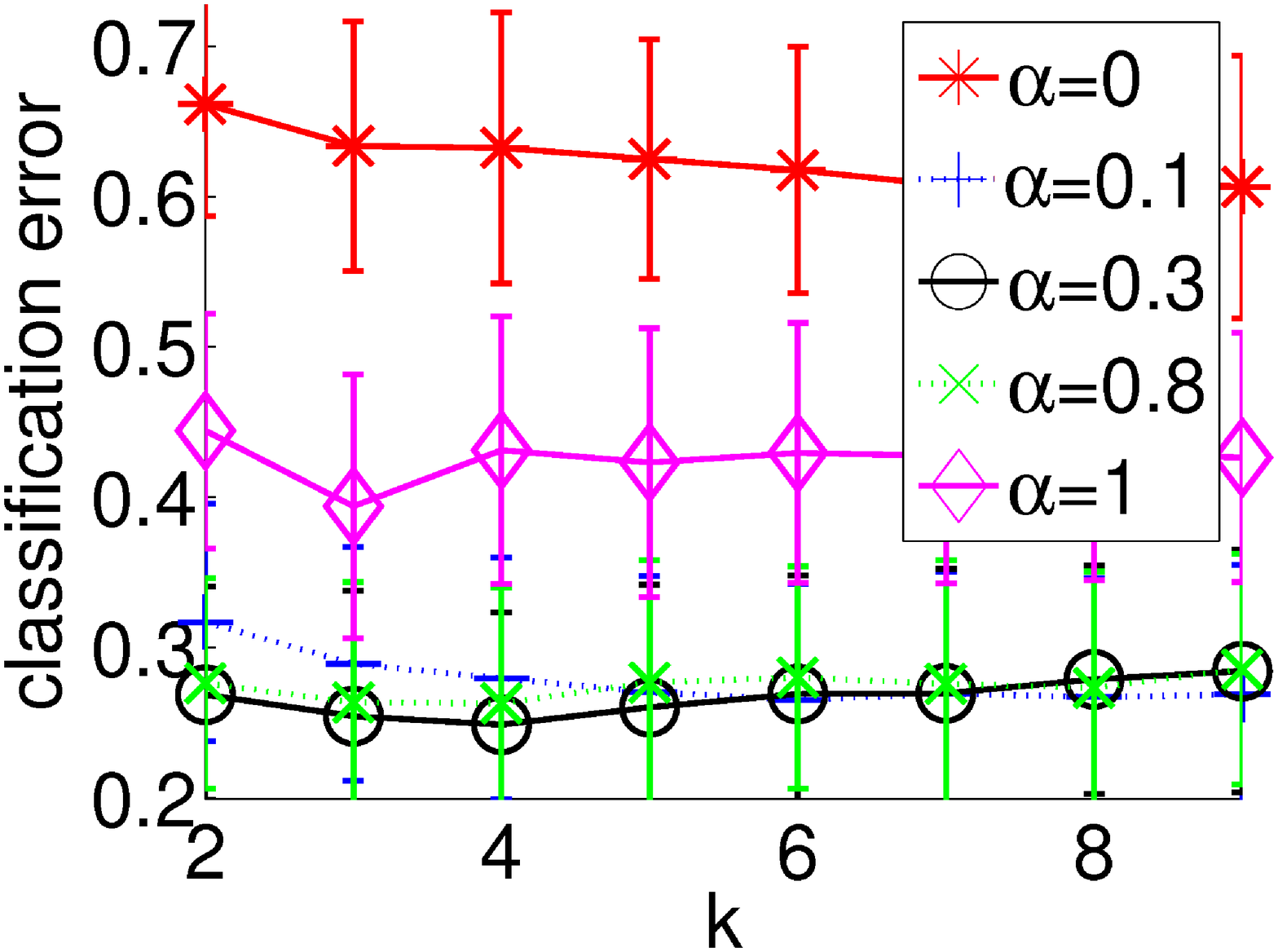}}\\
 \caption{(a) Classification error as a function of $\alpha $. (b) Classification error as a function of $k$. Classification error is estimated as cross validation error (five folds, 50 repetition), the range of the error bars is the standard deviation of test error over one fold (not averaged over folds and repetition) }
 \label{fig:accuracy_graphm}
\end{figure*}

Another example that we consider in this paper is the problem of chinese character recognition from the ETL9B dataset \cite{ETL9B}. The main idea is to use a score of optimal matching as a similarity measure between two images of characters. This similarity measure can be used then in machine learning algorithms, K-nearest neighbors (KNN) for instance, for character classification. Here we compare the performance of four methods: linear support vector machine (SVM), SVM with gaussian kernel, KNN based on score of shape context matching and KNN based on scores from graph matching which combines structural and shape context information. As a score, we use just the value of the objective function (\ref{eq:F_lambda_alpha}) at the (locally) optimal point. We have selected three chinese characters known to be difficult to distinguish by automatic methods. Examples of these characters as well as examples of extracted graphs (obtained by thinning and uniformly subsampling the images) are presented in Figure \ref{fig:chinese_characters_examples}. For SVM based algorithms, we use directly the values of image pixels (so each image is represented by a binary vector), in graph matching algorithm we use binary adjacency matrices of extracted graphs and shape context matrices (see \cite{belongie_shape_matching_shape_context}).
Our data set consist of 50 exemples (images) of each class. Each image is represented by $63\times64$ binary matrix. To compare different methods we use the cross validation error (five folds). The dependency of classification error from two algorithm parameters ($\alpha$ --- coefficient of linear combination (\ref{eq:F_lambda_alpha}) and $k$ --- number of nearest neighbors used in KNN)) is shown in Figure \ref{fig:accuracy_graphm}.

 Two extreme choices $\alpha = 1$ and $\alpha =0$ correspond respectively to pure shape context matching, i.e., when only node labels information is used, and pure unlabeled graph matching. It is worth observing here that KNN based just on the score of unlabeled graph matching does not work very well, the classification error being about 60\%. An explanation of this phenomenon is the fact that learning patterns have very unstable graph structure within one class. The pure shape context method has a classification error of about 39\%. The combination of shape context and graph structure informations allows to decrease the classification error down to 25\%. Beside the PATH algorithm, we tested also the QCV algorithm and the Umeyama algorithm,the Umeyama algorithm almost does not decrease the classification error. The QCV algorithm works better than then Umeyama algorithm, but still worse than the PATH algorithm.  Complete results can be found in Table \ref{tab:res_ch_chars}.

\begin{table}[htbp]
\caption{Classification of chinese characters. ($CV$, $STD$)---mean and standard deviation of test error over cross-validation runs (five folds, 50 repetitions)}
\label{tab:res_ch_chars}
\centering
\begin{tabular}{|l|c|c|c|}
\hline
Method&$CV$&$STD$\\
\hline
\hline
Linear SVM&0.377&$\pm$ 0.090 \\
SVM with gaussian kernel&0.359&$\pm$ 0.076 \\
KNN (PATH) ($\alpha$=1): shape context&0.399&$\pm$ 0.081 \\
KNN (PATH) ($\alpha$=0.4) &0.248&$\pm$ 0.075 \\
KNN (PATH) ($\alpha$=0): pure graph matching &0.607&$\pm$ 0.072 \\
KNN (U) ($\alpha$=0.9): $\alpha$ best choice &0.382&$\pm$ 0.077 \\
KNN (QCV) ($\alpha$=0.3): $\alpha$ best choice &0.295&$\pm$ 0.061 \\
\hline
\end{tabular}
\end{table}

\section{Conclusion}
We have presented the PATH algorithm, a new technique for graph matching based on convex-concave relaxations of the initial integer programming problem. PATH allows to integrate the alignment  of graph structural elements with the matching of vertices with similar labels. Its results are competitive with state-of-the-art methods in several graph matching and QAP benchmark experiments. Moreover, PATH has a theoretical and empirical complexity competitive with the fastest available graph matching algorithms.

Two points can be mentioned as interesting directions for further research.
First, the quality of the convex-concave approximation is defined by the choice of convex and concave relaxation functions. Better performances may be achieved by more appropriate choices of these functions. Second, another interesting point concerns the construction of a good concave relaxation for the problem of directed graph matching, i.e., for asymmetric adjacency matrix. Such generalizations would be interesting also as possible polynomial-time approximate solutions for the general QAP problem.

\appendix
\section{A toy example}
\label{app:toy}
The PATH algorithm does not generally find the global optimum of the NP-complete optimization problem. In this appendix we illustrate with two examples how the set of local optima tracked by PATH may or may not lead to the global optimum.

More precisely, we consider two simple graphs with the following adjacency matrices:
\begin{center}
$
\mathbf{G} =
\begin{bmatrix}
0 & 1 & 1 \\
1 & 0 & 0 \\
1 & 0 & 0 \\
\end{bmatrix}
$
and $\mathbf{H} =
\begin{bmatrix}
0 & 1 & 0 \\
1 & 0 & 0 \\
0 & 0 & 0 \\
\end{bmatrix}$.
\end{center}

\begin{figure*}[htb]
\centering
\subfigure[]{\includegraphics[width=7cm]{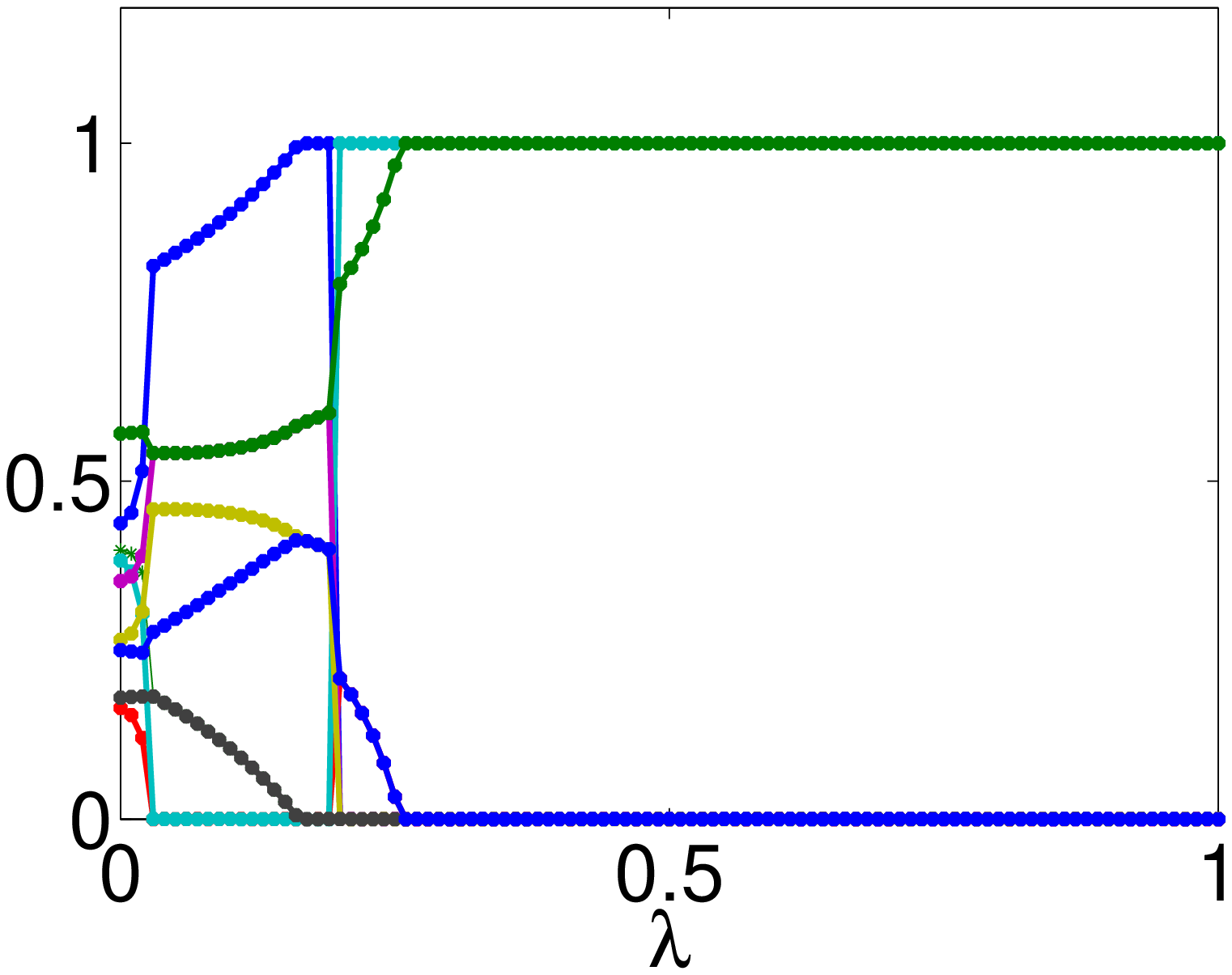}}
\subfigure[]{\includegraphics[width=7cm]{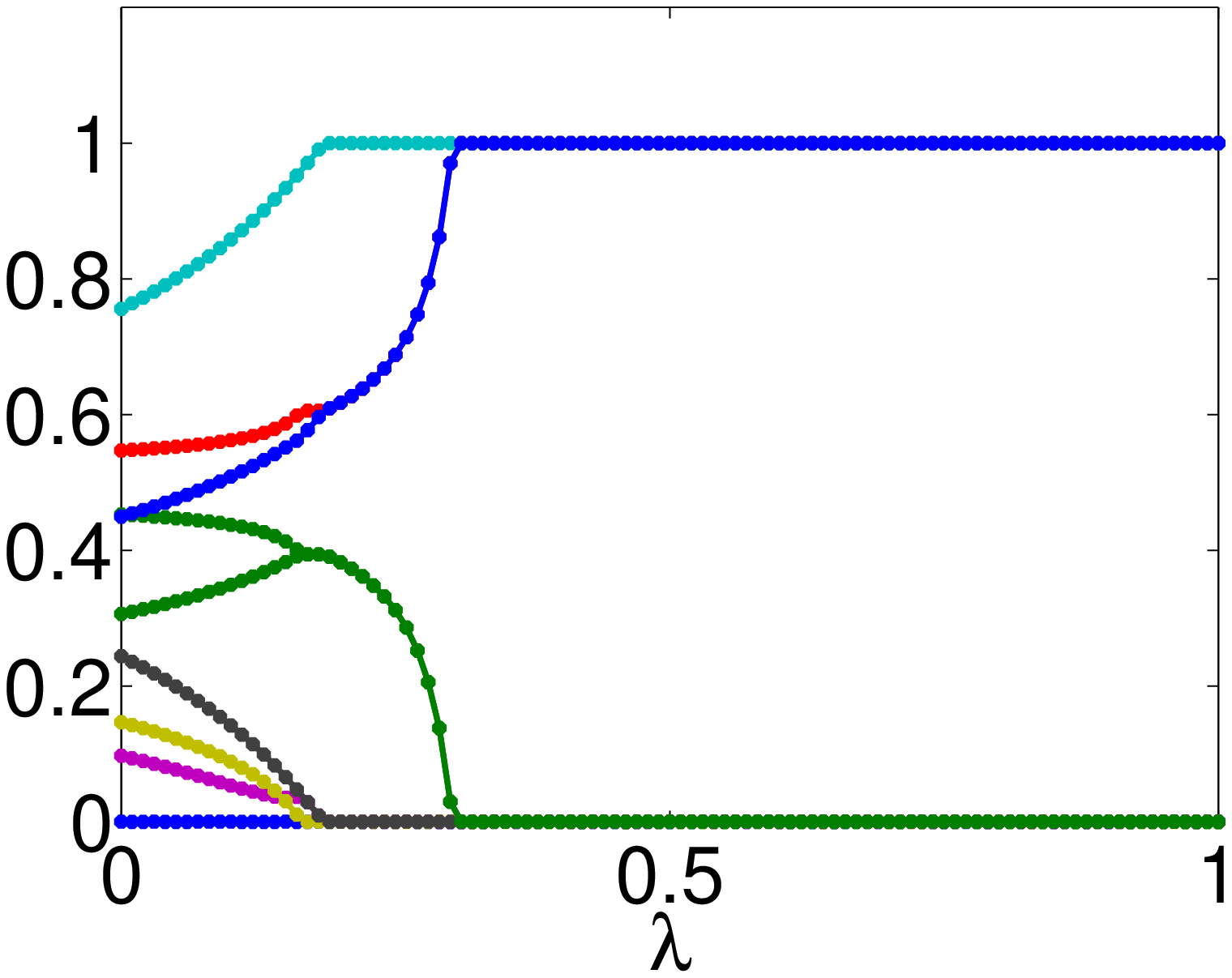}}
\caption{Nine coordinates of global minimum of $F^\alpha_\lambda$ as a function of $\lambda$}
\label{fig:ex_3}
\end{figure*}

Let $C$ denote the cost matrix of vertex association
$$\mathbf{C} =
\begin{bmatrix}
0.1691&0.0364&1.0509\\
0.6288&0.5879&0.8231\\
0.8826&0.5483&0.6100\\
\end{bmatrix}.$$
Let us suppose that we have fixed the tradeoff $\alpha=0.5$, and that our objective is then to find the global minimum of the following function:
\begin{equation}
F_0(P)=0.5||GP-PH||_F^2+0.5\trace(C'P),\qquad P\in{\mathcal P}.
\label{eq:toy_F}
\end{equation}
As explained earlier, the main idea underlying the PATH algorithm is to try to follow the path of global minima of $F_\lambda^\alpha(P)$ (\ref{eq:F_lambda_alpha}). This may be possible if all global minima $P^*_{\lambda}$ form a continuous path, which is not true in general. In the case of small graphs we can find the exact global minimum of $F_\lambda^\alpha (P)$ for all $\lambda$. The trace of global minima as functions of $\lambda$ is presented in Figure \ref{fig:ex_3}(a) (i.e., we plot the values of the nine parameters of the doubly stochastic matrix, which are, as expected, all equal to zero or one when $\lambda=1$). When $\lambda$ is near $0.2$ there is a jump of global minimum from one face to another. However if we change the linear term $C$ to
$$\mathbf{C'} =
\begin{bmatrix}
0.4376&0.3827&0.1798\\
0.3979&0.3520&0.2500\\
0.1645&0.2653&0.5702\\
\end{bmatrix},$$
then the trace becomes smooth (see Figure \ref{fig:ex_3}(b)) and the PATH algorithm then finds the globally optimum point. Characterizing cases where the path is indeed smooth is the subject of ongoing research.
\section{Kronecker product}
\label{sec:kron}
The Kronecker product of two matrices $A\otimes B$ is defined as follows:
\begin{equation*}
A\otimes B=
\begin{bmatrix} 
B a_{11} & \cdots & B a_{1n} \\ 
\vdots & \ddots & \vdots \\ 
B a_{m1} & \cdots & B a_{mn}  
\end{bmatrix}. 
\end{equation*} 

Two important properties of Kronecker product that we use in this paper are: 
\begin{equation*}
\begin{split}
&(A^T\otimes B) \mbox{vec}(X) = \mbox{vec}(BXA)\mbox{, }\\
&\mbox{ and } \trace(X^TAXB^T)=\mbox{vec}(X)^T (B\otimes A) \mbox{vec}(X)\,.
\end{split}
\end{equation*}


\end{document}